\documentclass{article}

\PassOptionsToPackage{authoryear, round}{natbib}

 \usepackage[main, final]{neurips_2026}

\usepackage[utf8]{inputenc} 
\usepackage[T1]{fontenc}    
\usepackage{url}            
\usepackage{booktabs}       
\usepackage{amsfonts}       
\usepackage{nicefrac}       
\usepackage{microtype}      
\usepackage{xcolor}         

\usepackage{multirow}
\usepackage{tabularx}
\usepackage{makecell}
\usepackage{array}

\newcolumntype{Y}{>{\centering\arraybackslash}X}
\definecolor{royalblue}{rgb}{0.254, 0.411, 0.883}
 \definecolor{myblue}{rgb}{0.12,0.35,0.62}\usepackage[pagebackref,breaklinks,colorlinks,allcolors=myblue]{hyperref}
\newcommand{\method}{\textsc{EPIC}}
\newcommand{\methodfull}{Efficient Predicate-Guided Inference-Time Control}

\newcommand{\best}[1]{\textbf{#1}}
\newcommand{\second}[1]{\underline{#1}}

\newif\ifshowcomments
\showcommentstrue

\usepackage{amsmath} 
\usepackage[pdftex]{graphicx}
\usepackage{subcaption}
\usepackage{placeins}

\usepackage{tcolorbox}

\title{EPIC: \textcolor{royalblue}{E}fficient \textcolor{royalblue}{P}redicate-Guided \textcolor{royalblue}{I}nference-Time \textcolor{royalblue}{C}ontrol for Compositional Text-to-Image Generation}

\author{%
  Sunung Mun\textsuperscript{1}
  \quad
  Sunghyun Cho\textsuperscript{1,2}
  \quad
  Jungseul Ok\textsuperscript{1,2}\thanks{
    Correspondence to
    \protect\href{mailto:jungseul@postech.ac.kr}
    {\texttt{jungseul@postech.ac.kr}}.
  } \\
  \textsuperscript{1}Graduate School of Artificial Intelligence, POSTECH \\
  \textsuperscript{2}Department of Computer Science \& Engineering, POSTECH \\
  \texttt{\{mtablo,s.cho,jungseul\}@postech.ac.kr} \\
    {\small Project page:
    \url{https://mtablo.github.io/EPIC_page/}}
}

\begin{document}

\maketitle

\begin{abstract}
Recent text-to-image (T2I) generators can synthesize realistic images,
but still struggle with compositional prompts involving multiple
objects, counts, attributes, and relations. We introduce \method{}
(\methodfull{}), a training-free inference-time refinement framework
for compositional T2I generation. \method{} casts refinement as
predicate-guided search: it parses the original prompt once into a
visual program of object variables and typed predicates,
then verifies each candidate image against this program
using visual evidence extracted from that image.
A candidate is judged to satisfy the prompt when all predicates pass;
otherwise, unmet predicates guide the next step,
routing localized failures to targeted editing and
layout-level or persistent failures to resampling.
On GenEval2, \method{} improves the prompt-level Soft-TIFA GM score from
$34.16\%$ for single-pass generation to $71.46\%$.
With the same generator, editor, and image-model execution budget,
\method{} outperforms the strongest baseline by $19.23$ points.
It also uses $31\%$ fewer image-model executions, $72\%$ fewer
multimodal large language model (MLLM) calls, and $81\%$ fewer MLLM
tokens per prompt.
\end{abstract}

\section{Introduction}
\label{sec:intro}

Recent text-to-image (T2I) generators can synthesize realistic images
from natural language prompts~\citep{
ho2020ddpm,song2020ddim,lipman2022flow,liu2022rectflow,
song2020score,dhariwal2021diffusion,esser2024scaling,nichol2021glide}.
However, they still struggle with compositional prompts that specify
multiple objects, counts, attributes, and spatial
relations~\citep{ghosh2023geneval,kamath2025geneval2,
huang2025t2icompp}. For example, given a prompt asking for three red
cups to the left of four blue bowls, a generator may produce only one cup
or place the cups on the wrong side.
Such an image may look plausible, but it does not satisfy all the
requirements in the prompt.
Joint satisfaction becomes harder as
requirements accumulate.

Improving the generator itself can reduce these failures,
for example through larger models, more training data,
better captions, or preference-based
post-training~\citep{xu2023imagereward,wu2023hpsv2,ma2025hpsv3}.
However, such training-time improvements can be costly
and may be unavailable for closed models.
Inference-time scaling offers another approach: it keeps the
image model fixed and uses additional computation at test time to
improve the generated result. 
An efficient method should therefore improve prompt
satisfaction per unit of inference-time computation,
rather than simply relying on a larger budget.
Thus, the allocation of this computation is an important design choice.

Inference-time computation for T2I generation can be spent in several
ways: increasing denoising steps, searching over initial noises or
denoising trajectories, generating and ranking multiple candidates,
rewriting prompts, or editing earlier images over several
rounds~\citep{ma2025itsdiffusion,xie2025sana,chen2025t2icopilot,
khan2025ttprompt,jiang2026raise,li2025reflectdit,zhuo2025reflectflow,yang2024rpg}.
Many methods use a verifier to guide this computation,
for example by computing an image--text alignment score for each candidate~\citep{radford2021clip,tschannen2025siglip2,li2024vqascore}.
Such scores provide the feedback needed for
selection-based methods, such as best-of-$N$ sampling,
where the system ranks candidates and keeps the highest-scoring one.
However, a scalar score provides limited feedback on how an image
fails to match the prompt. A system that refines or edits an image
needs more specific feedback about which requirements are unmet.

\begin{figure}[t]
    \centering
    \includegraphics[width=\linewidth]{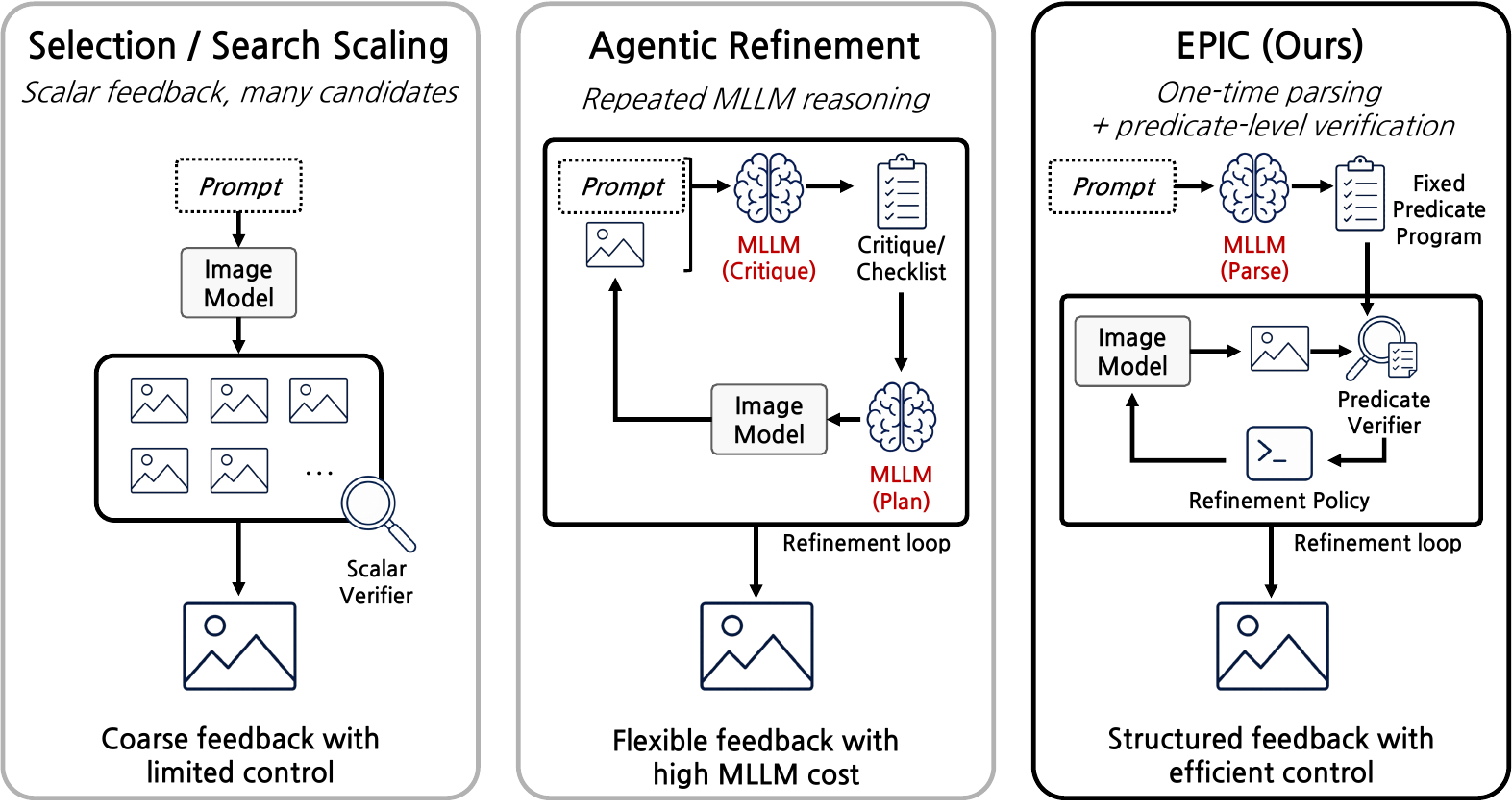}
    \caption{
        Comparison of inference-time scaling strategies for compositional
        text-to-image generation. Selection/search scaling uses scalar
        feedback to choose among many candidates, while agentic refinement
        obtains flexible feedback through repeated MLLM critique and
        planning. \method{} instead parses the prompt once into a fixed
        predicate program and uses predicate-level verification to provide
        structured feedback for efficient refinement control.
    }
    \label{fig:teaser}
\end{figure}

Knowing which requirement failed matters because different failures
may require different actions. A count or attribute error may be
addressed with a targeted edit, such as adding missing cups or
recoloring an object, whereas an incorrect spatial relation may
require a new layout. When repeated edits fail to fix an error,
generating a new image can work better than editing again.

One way to identify such failures is to use multimodal large
language models (MLLMs) to compare the generated image with the
prompt and provide feedback on unmet requirements.
Recent agentic T2I methods use this feedback to guide
refinement~\citep{chen2025t2icopilot,jiang2026raise,yang2024rpg}.
These methods often represent requirements and verification
feedback as natural-language reports or checklists.
RAISE~\citep{jiang2026raise}, for example, revises its
requirement checklist using the original prompt, the current
candidate, and previous verification feedback.
While this allows the checklist to adapt to intermediate
results, an incorrect judgment may affect the requirements
used in subsequent rounds, potentially causing refinement
to drift from the original prompt.
Repeated requirement analysis also incurs additional MLLM
calls, even though the original prompt remains unchanged.

Our key idea is to use prompt decomposition not only for
evaluation, but also as a fixed representation for controlling
refinement. We introduce \method{} (\methodfull{}), a
training-free framework that parses the original prompt once
into a visual program of object variables and typed predicates.
Each predicate represents a checkable requirement, such as
an object count, attribute, or spatial relation.
The program remains fixed throughout refinement, so an
incorrect judgment on one candidate cannot change the
requirements used to evaluate subsequent candidates.
For each generated or edited image, modular visual tools
evaluate the predicates and assign states of satisfied,
uncertain, or violated.
These states determine whether to accept a candidate
or refine it further.
When refinement is needed, the unmet predicates and
previous attempts guide the choice between targeted
editing and resampling.
Figure~\ref{fig:teaser} contrasts this approach with
selection-based scaling and agentic refinement.

We evaluate \method{} on GenEval2~\citep{kamath2025geneval2} using
the same generator across methods and the same editor where
applicable. All inference-time methods share the same maximum
budget of 32 image-model executions. \method{} improves the prompt-level
Soft-TIFA GM score from $34.16\%$ for single-pass generation to
$71.46\%$. The gain is particularly pronounced on prompts with many
visual requirements. Compared with RAISE, \method{} improves the
score from $52.23\%$ to $71.46\%$ while reducing average image-model
executions from $19.8$ to $13.6$ per prompt. MLLM calls also decrease
from $9.12$ to $2.53$, and MLLM token use falls from $49.4$k to
$9.2$k per prompt.

Our contributions are summarized as follows:
\begin{itemize}
\item We formulate compositional T2I refinement as
predicate-guided search, using a fixed visual program
to represent the original prompt requirements.

\item We introduce \method{}, a training-free framework
that verifies candidate images with predicate-specific
visual tools and uses the resulting states to guide
targeted editing or resampling.

\item We show on GenEval2 that \method{} improves the prompt-level
Soft-TIFA GM score over single-pass generation and the compared
inference-time baselines, while using fewer image-model executions,
MLLM calls, and MLLM tokens than RAISE.
\end{itemize}

\section{Related Work}
\label{sec:related}

\subsection{Compositional T2I evaluation via atomic checks}
Compositional T2I alignment requires an image to satisfy multiple
visual conditions jointly, including object presence, counts,
attributes, and relations. Benchmarks such as
GenEval~\citep{ghosh2023geneval}, GenEval2~\citep{kamath2025geneval2}, and
T2I-CompBench~\citep{huang2023t2icomp,huang2025t2icompp} evaluate these conditions and show that current
generators still struggle as prompt compositionality
increases~\citep{gokhale2022benchmarking,cho2023dall}. This has motivated evaluation methods that
decompose prompt following into finer-grained checks, including
object-centric evaluation~\citep{ghosh2023geneval,kamath2025geneval2},
visual-question decomposition~\citep{hu2023tifa}, and
programmatic or graph-based representations~\citep{gupta2023visprog,
cho2023visual,cho2023davidsonian}.
\method{} builds on this atomic view, but uses a fixed visual program to guide both verification
and refinement, rather than evaluation alone.

\subsection{Inference-time scaling and verifier-guided search}
Inference-time scaling improves generation by allocating
additional computation at test time.
Prior work has explored several scaling axes beyond
increasing denoising steps, including search over
initial noise samples or denoising
trajectories~\citep{ma2025itsdiffusion,zhang2025inference},
reward-guided sampling and candidate
selection~\citep{xie2025sana,singhal2025fksteering},
and prompt rewriting~\citep{khan2025ttprompt}.
Other approaches train generators to support iterative
reflection-based refinement~\citep{
li2025reflectdit,zhuo2025reflectflow}.

Search and selection methods commonly use scalar or
holistic image--text feedback from CLIP-like similarity
models~\citep{radford2021clip,tschannen2025siglip2},
VQA-based scorers~\citep{li2024vqascore}, or
VLM-based evaluators~\citep{liu2025nvila}.
Such scores provide useful feedback for candidate
selection and search, but offer limited information
about which requirements of a compositional prompt
are unmet or how they should be corrected.
\method{} instead represents verification as a
vector of typed predicate states, which guides
both candidate evaluation and the choice between
targeted editing and resampling.

\subsection{Agentic refinement and requirement-driven T2I systems}
Agentic T2I systems use LLMs or MLLMs for prompt
interpretation and planning~\citep{
lian2023llmdiffusion,feng2023layoutgpt,yang2024rpg},
as well as image evaluation and iterative
refinement~\citep{yang2026m3,chen2025t2icopilot,
wang2024genartist,jiang2026raise}.
These methods often use natural-language reports
or checklists to represent requirements and provide
refinement feedback.

T2I-Copilot~\citep{chen2025t2icopilot} focuses on
interactive generation through prompt interpretation,
ambiguity resolution, model selection, and MLLM-based
quality evaluation.
Its analysis report and quality scores provide a
flexible interface for regeneration or user interaction.

RAISE~\citep{jiang2026raise}, the strongest refinement
baseline in our comparison, combines noise resampling,
prompt rewriting, and instructional editing.
It evolves a candidate population using MLLM analysis,
tool-grounded verification, and a requirement checklist
revised across refinement rounds.
In contrast, \method{} uses a fixed, typed predicate
program to verify each candidate and guide refinement
actions without repeatedly re-analyzing the original
requirements.

\section{Method}
\label{sec:method}

\subsection{\methodfull{} (\method{}) Overview}
\label{sec:method_overview}

\begin{figure}[t]
    \centering
    \includegraphics[width=\linewidth]{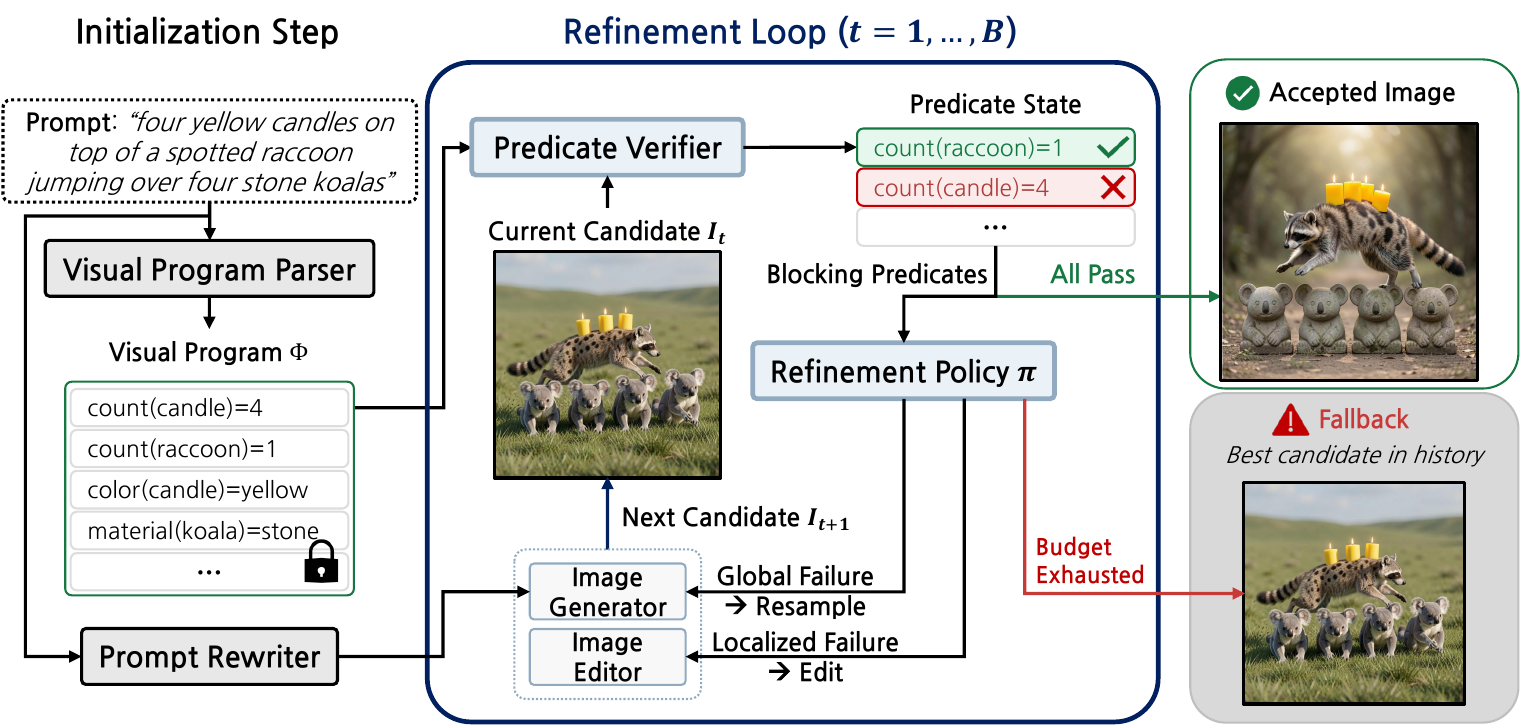}
    \caption{
        Method overview.
        \method{} compiles the input prompt once into a fixed visual program
        $\Phi$ of typed predicates and prepares prompt rewrites intended
        to preserve meaning for candidate generation.
        Starting from an initial candidate generated from a rewrite, each
        round verifies the current candidate, forms a blocking set from
        unsatisfied predicates, and uses this state to choose acceptance,
        resampling from a rewrite, or targeted editing.
        Acceptance follows Eq.~\eqref{eq:acc_gate}, including the limited
        auditor override.
        If the budget is exhausted before acceptance, \method{} returns the
        best partial candidate from the verified history as a fallback.
    }
    \label{fig:method}
\end{figure}

We propose \method{}, a training-free inference-time refinement
framework for compositional text-to-image generation.
As illustrated in Figure~\ref{fig:method}, \method{} first
compiles the original prompt $p$ into a fixed visual program
of typed predicates and prepares prompt rewrites for
generating alternative candidates.
Every generated or edited candidate is verified against
the same visual program.

Given an image-model execution budget $B$, where each
execution is one call to the generator or editor, \method{}
refines the image over a sequence of rounds.
Each round produces one candidate by generation or editing
and verifies it against the fixed visual program.
If the candidate is accepted, inference terminates.
Otherwise, the currently blocking predicates determine
the next action: localized failures can trigger targeted
editing, while layout-level or persistent failures trigger
resampling.
Here, resampling means generating a fresh image from a
precomputed prompt rewrite.
Thus, refinement decisions are based on verification
results for individual predicates rather than repeated
holistic MLLM critiques.
Appendices~\ref{appendix:method_details}
and~\ref{appendix:implementation_details} provide the
complete program schema, verifier rules, routing policy,
and implementation details.

\subsection{Visual-program construction}
\label{sec:method_compile}

\method{} compiles the source prompt into a visual program
\begin{equation}
\label{eq:visual_program}
    \Phi
    =
    \mathrm{Compile}(p)
    =
    (\mathcal{O}, \mathcal{P}),
    \qquad
    \mathcal{P} = \{\phi_k\}_{k=1}^{K},
\end{equation}
where $\mathcal{O}$ is the set of prompt-described object
entities and $\mathcal{P}$ is the set of typed visual
predicates over those entities.
Predicate types cover the compositional conditions that
the framework can verify, including object existence and
cardinality, attributes, spatial relations, global scene
requirements, exclusions, and visible text.
Attribute predicates cover object-level properties such
as appearance, pose, state, and action-related requirements.
Object existence is represented as a count-style requirement,
such as requiring at least one instance of an object category.
Each predicate specifies typed arguments and an expected
condition, including ordered subject--reference pairs for
spatial relations.

Compilation consists of MLLM-based parsing, deterministic
normalization and validation, and conditional MLLM review.
During normalization and validation, \method{} applies rule-based
repairs and checks for issues such as undeclared object references
and unsupported predicate vocabulary.
If the resulting report records a fix or warning, an MLLM reviewer
is invoked once to validate or repair the normalized program.

Separately, \method{} prepares a set of prompt rewrites,
\[
    \mathcal{R}
    =
    \mathrm{Rewriter}(p)
    =
    \{r_j\}_{j=1}^{M}.
\]
These rewrites are intended to preserve the original
requirements.
They are generated once and used as the prompt pool for
initial generation and later resampling.

\subsection{Predicate verification}
\label{sec:method_verify}

At refinement round $t$, \method{} has a candidate image
$I_t$, produced either by generation or by editing $I_{t-1}$.
The verifier evaluates $I_t$ against the fixed visual
program $\Phi$ by assigning each predicate a discrete state:
\begin{equation}
\label{eq:predicate_state}
    s_{t,k} = V_{\phi_k}(I_t),
    \qquad
    S_t = (s_{t,1}, \ldots, s_{t,K}),
    \qquad
    s_{t,k} \in
    \{\mathrm{satisfied}, \mathrm{uncertain}, \mathrm{violated}\}.
\end{equation}
Here, $V_{\phi_k}$ is the predicate-specific verifier for
$\phi_k$, and $S_t$ is the full predicate satisfaction
vector for $I_t$.
The specification $\Phi$ stays fixed, while $S_t$ is
recomputed for every candidate.
The $\mathrm{uncertain}$ state is an abstention: the
available evidence is partial, ambiguous, or unreliable
for deciding that the predicate is either satisfied
or violated.

Each predicate is verified separately, although multiple
predicates may share the same underlying visual evidence.
Object existence and count predicates are evaluated
using detected object instances.
Spatial relations use detected boxes and masks of the
subject and reference objects, together with depth estimates
for depth-sensitive relations.
Attribute predicates use region--text scores between
attribute--object queries and context-preserving crops
of detected object regions.
For action-related attributes, region--text scores are used
initially; once all non-action predicates are satisfied,
a crop-level MLLM verifier determines their states.
Global-scene predicates are evaluated using detected scene regions,
with image--background similarity scores as a fallback.
Exclusion predicates query forbidden categories and require
that no valid detection remains.
Visible-text predicates use a text-in-image verifier.
Thus, most predicate families are checked by modular
perception tools rather than holistic MLLM critique.

For each predicate, the verifier returns a discrete state
and a verification score when available.
Let $Q_t = (q_{t,1}, \ldots, q_{t,K})$ denote these scores,
which are used for action selection, retry decisions,
and fallback ranking.
A candidate satisfies the fixed program only when all
predicates are satisfied.
When this program-level acceptance criterion fails,
\method{} forms the set of predicates that still
block acceptance:
\begin{equation}
\label{eq:program_gate}
\begin{aligned}
    A_{\mathrm{prog}}(I_t, \Phi)
    &=
    \prod_{\phi_k \in \mathcal{P}}
    \mathbf{1}[s_{t,k}=\mathrm{satisfied}], \\
    \mathcal{F}_t
    &=
    \{\phi_k \in \mathcal{P}
    \mid s_{t,k}\neq \mathrm{satisfied}\}.
\end{aligned}
\end{equation}
This gate is conservative: any predicate not verified
as satisfied blocks acceptance.
To handle persistent verification failures, \method{}
allows a limited auditor override for eligible attribute,
relation, or scene predicates.
Let $o_t^{\mathrm{audit}}\in\{0,1\}$ denote this decision.
The override is active only when all remaining blocking
predicates have persisted, belong to eligible families,
and an MLLM auditor judges them as satisfied; otherwise,
$o_t^{\mathrm{audit}}=0$.
The override can apply to uncertain or violated states,
but counts, exclusions, visible text, and directional
or depth-sensitive relations remain governed by their
standard predicate checks.
The resulting acceptance gate is
\begin{equation}
\label{eq:acc_gate}
    A_{\mathrm{acc}}(I_t, \Phi)
    =
    A_{\mathrm{prog}}(I_t, \Phi)
    \vee
    o_t^{\mathrm{audit}}.
\end{equation}

\subsection{Predicate-driven refinement}
\label{sec:method_refine}

If the final acceptance gate in Eq.~\eqref{eq:acc_gate}
fails and the execution budget remains, \method{} uses
the blocking predicates as action targets.
A refinement policy $\pi$ chooses the next action from
the blocking set $\mathcal{F}_t$, the predicate states $S_t$,
the available verification scores $Q_t$, and the history
$H_t$ of previous refinement attempts.
The next candidate is obtained by either editing
the current image or resampling:
\begin{equation}
\label{eq:refine_step}
\begin{aligned}
    a_t
    &=
    \pi(\mathcal{F}_t, S_t, Q_t, H_t),
    \qquad
    a_t \in
    \{\mathtt{edit}, \mathtt{resample}\}, \\
    I_{t+1}
    &=
    \begin{cases}
        E(I_t, u_t), & a_t = \mathtt{edit}, \\
        G(r_t),      & a_t = \mathtt{resample},
    \end{cases}
    \qquad
    r_t \in \mathcal{R}.
\end{aligned}
\end{equation}
Here, $E$ is the editor, $G$ is the generator, $u_t$
is a targeted edit instruction, and $r_t$ is a rewrite
selected from the precomputed rewrite pool.
The history $H_t$ records which predicates have been
targeted, which actions were taken, and how the
corresponding predicate states or scores changed.

The policy first selects a blocking predicate based on its
state, type, and verification score.
It then determines whether the error can be addressed by
editing the current image.
For exact count errors, the policy considers adding or
removing objects only when the required object categories
are already detected; otherwise, it resamples.
Exclusion failures can trigger edits to remove forbidden
objects.
Errors in attributes or actions can trigger edits once
the relevant objects and their prerequisites are established.
Spatial relation errors generally lead to resampling
because they often require changes to the image layout.
Edits to the overall scene are considered only after
requirements on objects, counts, attributes, spatial
relations, and exclusions are satisfied.
The policy permits limited retries when verification
improves relative to a previous unsuccessful edit;
otherwise, it resamples.
These decisions use the failed requirements and previous
refinement outcomes to determine how the remaining
image-model executions are allocated.

\subsection{Acceptance and fallback}
\label{sec:method_select}

Refinement returns the first candidate that satisfies
the acceptance gate in Eq.~\eqref{eq:acc_gate}.
If the budget is exhausted before any candidate satisfies
the acceptance gate, \method{} returns a best-partial
fallback from the verified candidate history.
Candidates are ranked lexicographically by satisfied
predicate counts across prioritized predicate families,
followed by total predicate satisfaction, verification
scores for remaining unsatisfied predicates, and
generation order.
Such outputs are recorded as budget-exhausted fallbacks
rather than internally accepted candidates.

\section{Experiments}
\label{sec:experiments}

\subsection{Experimental setup}
\label{sec:exp_setup}

\paragraph{Benchmark and metrics.}
We evaluate compositional prompt alignment on
GenEval2~\citep{kamath2025geneval2}, which contains 800 prompts
with varying atomicity, the number of primitive visual requirements.
We report prompt-level and atom-level accuracy estimates using
Soft-TIFA GM and AM, respectively.
For each prompt, these metrics aggregate correct-answer probabilities
across its requirements using the geometric or arithmetic mean,
respectively, and then average the results across prompts.
We use prompt-level accuracy as the primary metric because
compositional alignment requires the joint satisfaction
of the prompt's visual requirements.

\paragraph{Models, baselines, and cost.}
Unless otherwise stated, all methods use
FLUX.2-Klein-9B~\citep{flux2klein9b} as the generator and,
where applicable, as the image-conditioned editor.
Agentic methods use Qwen3-VL-32B~\citep{bai2025qwen3}
with MLLM decoding temperature $T=0$.
We compare single-pass generation,
BoN+NVILA~\citep{liu2025nvila},
T2I-Copilot~\citep{chen2025t2icopilot},
and RAISE~\citep{jiang2026raise}.
For the main comparison, inference-time methods share
a maximum budget of $B=32$ image-model executions,
each corresponding to one generator or editor call.
We additionally vary this budget to examine the
accuracy--compute trade-off.
We report average image-model executions (\emph{Exec.}),
MLLM calls (\emph{Call}), and MLLM input/output tokens
in thousands (\emph{kTok}) per prompt.

\subsection{Main results on GenEval2}
\label{sec:geneval2_main}

\begin{figure}[t]
\centering
\begin{subfigure}[t]{0.49\linewidth}
    \centering
    \includegraphics[width=\linewidth]{
        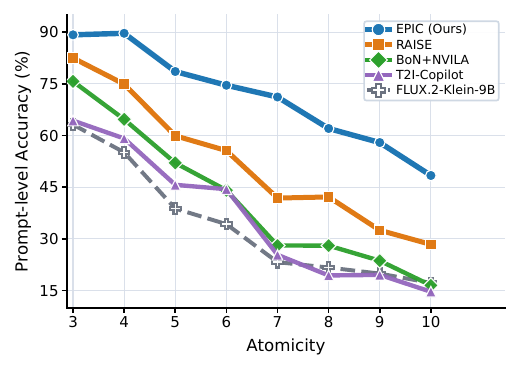
    }
    \vspace{-0.6cm}
    \caption{
    Prompt-level accuracy by atomicity
    }
    \label{fig:atomicity_plot}
\end{subfigure}
\hfill
\begin{subfigure}[t]{0.49\linewidth}
    \centering
    \includegraphics[width=\linewidth]{
        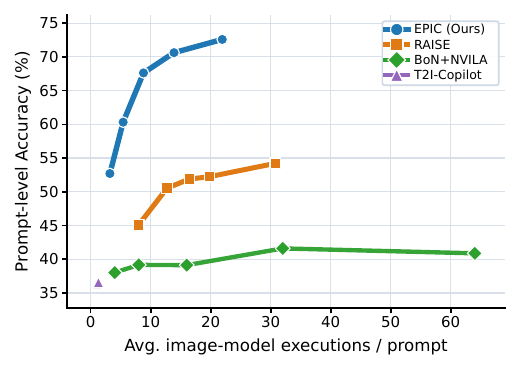
    }
    \vspace{-0.6cm}
    \caption{
    Prompt-level accuracy vs.\ realized executions
    }
    \label{fig:pareto_plot}
\end{subfigure}
\caption{
GenEval2 results at MLLM decoding temperature $T=0$.
Left: prompt-level accuracy by atomicity. Right: prompt-level
accuracy versus realized image-model executions under budget sweeps (Appendix~\ref{appendix:geneval2_budget_behavior}).
}
\label{fig:main_results_plots}
\end{figure}

Figure~\ref{fig:main_results_plots} shows prompt-level accuracy
across prompt complexity and inference-time compute.
Accuracy generally decreases as atomicity increases,
but \method{} maintains a substantial advantage on prompts
with many requirements.
Across the budget sweep, \method{} also achieves
higher accuracy than RAISE with fewer realized
image-model executions.

\begin{table}[t]
\centering
\caption{GenEval2 prompt-level and atom-level accuracy (\%)
and inference cost at $T=0$.
$\Delta P$ is the gain in prompt-level accuracy over single-pass
FLUX.2-Klein-9B, in percentage points. \emph{Exec.} counts image-model executions;
\emph{Call} and \emph{kTok} count Qwen3-VL-32B calls and
input/output tokens in thousands. Best and second-best inference-time
methods are shown in bold and underlined, respectively.}
\label{tab:geneval2_main}
\scriptsize
\setlength{\tabcolsep}{2.9pt}
\renewcommand{\arraystretch}{1.10}
\begin{tabular*}{\linewidth}{@{\extracolsep{\fill}}lccccccccc@{}}
\toprule
\multirow{2}{*}{Method}
& \multicolumn{3}{c}{Accuracy (\%)}
& \multicolumn{3}{c}{Cost / prompt}
& \multicolumn{3}{c}{Gain / cost} \\
\cmidrule(lr){2-4}
\cmidrule(lr){5-7}
\cmidrule(lr){8-10}
& GM $\uparrow$
& $\Delta P$ $\uparrow$
& AM $\uparrow$
& Exec. $\downarrow$
& Call $\downarrow$
& kTok $\downarrow$
& $\Delta P$/Exec. $\uparrow$
& $\Delta P$/Call $\uparrow$
& $\Delta P$/kTok $\uparrow$ \\
\midrule
FLUX.2-Klein-9B
& 34.16 & -- & 79.06
& 1.0 & -- & -- & -- & -- & -- \\

\midrule
\multicolumn{10}{l}{\emph{Inference-time scaling with FLUX.2-Klein-9B}} \\

BoN+NVILA
& 41.60 & +7.44 & 82.38
& 32.0 & -- & -- & 0.23 & -- & -- \\

T2I-Copilot
& 36.56 & +2.40 & 81.19
& \best{1.24} & \second{4.48} & \second{15.2}
& \second{1.94} & 0.54 & 0.16 \\

RAISE
& \second{52.23} & \second{+18.07} & \second{86.28}
& 19.8 & 9.12 & 49.4
& 0.91 & \second{1.98} & \second{0.37} \\

\method{}
& \best{71.46} & \best{+37.30} & \best{90.25}
& \second{13.6} & \best{2.53} & \best{9.2}
& \best{2.74} & \best{14.74} & \best{4.05} \\
\bottomrule
\end{tabular*}
\end{table}

\begin{figure}[t]
\centering
\scriptsize
\setlength{\tabcolsep}{1.2pt}
\renewcommand{\arraystretch}{1.04}
\begin{tabular}{ccccc}
\textbf{FLUX.2-Klein-9B}
& \textbf{BoN+NVILA}
& \textbf{T2I-Copilot}
& \textbf{RAISE}
& \textbf{\method{} (ours)} \\

\includegraphics[width=0.185\linewidth]{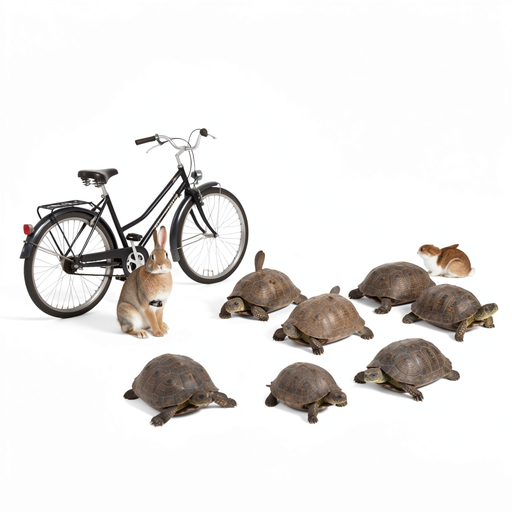}
& \includegraphics[width=0.185\linewidth]{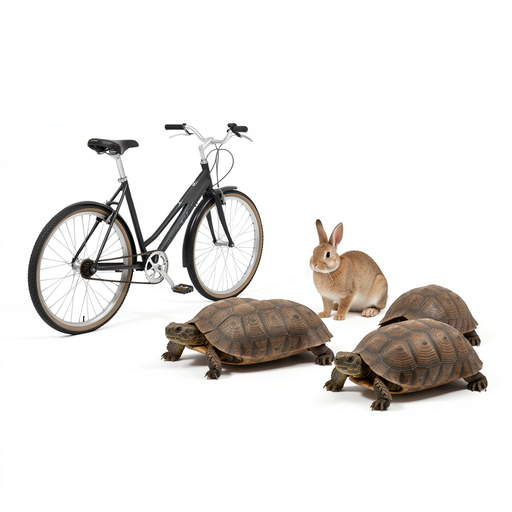}
& \includegraphics[width=0.185\linewidth]{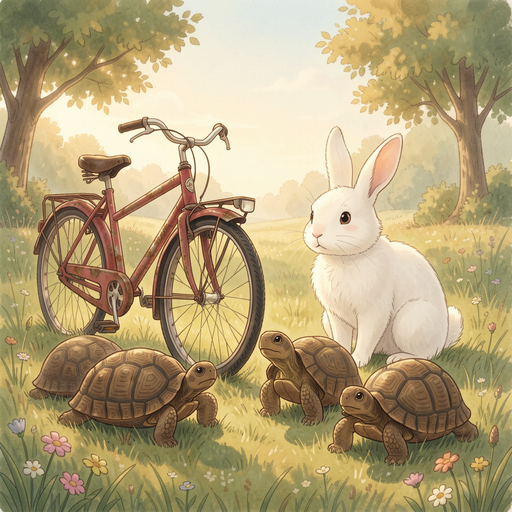}
& \includegraphics[width=0.185\linewidth]{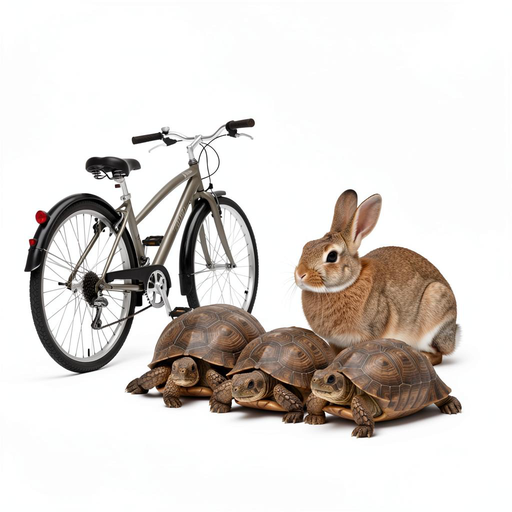}
& \includegraphics[width=0.185\linewidth]{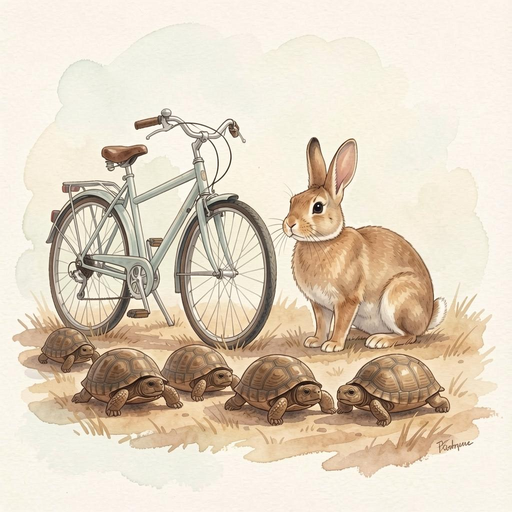} \\
\multicolumn{5}{p{0.96\linewidth}}{\centering
\textbf{Atomicity 6.} \textit{Prompt:} a bicycle to the left of a rabbit, and five brown turtles
} \\[0.55em]

\includegraphics[width=0.185\linewidth]{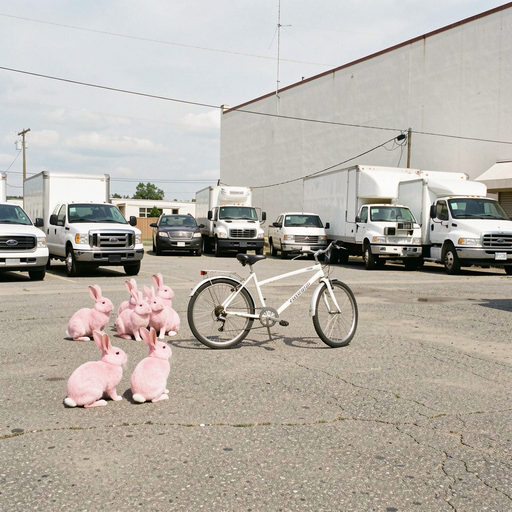}
& \includegraphics[width=0.185\linewidth]{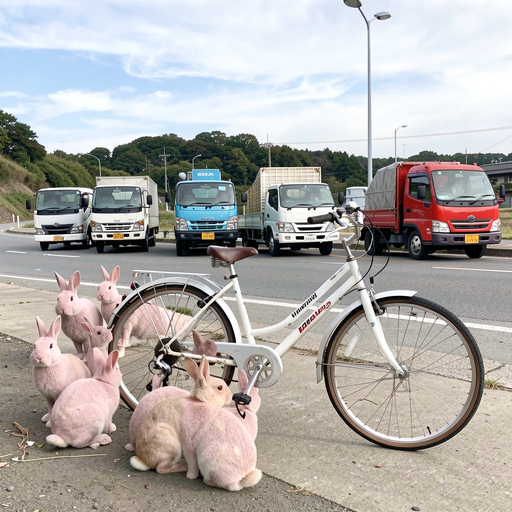}
& \includegraphics[width=0.185\linewidth]{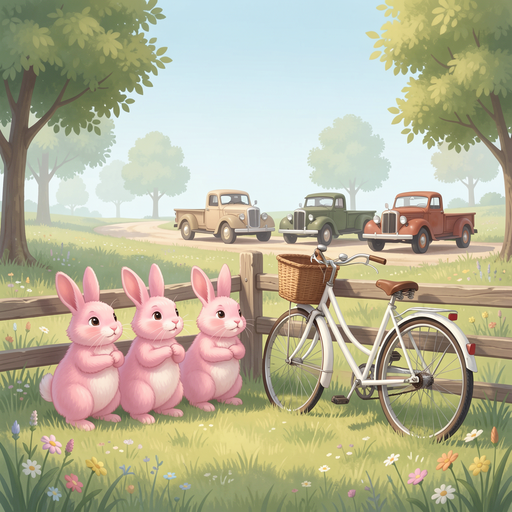}
& \includegraphics[width=0.185\linewidth]{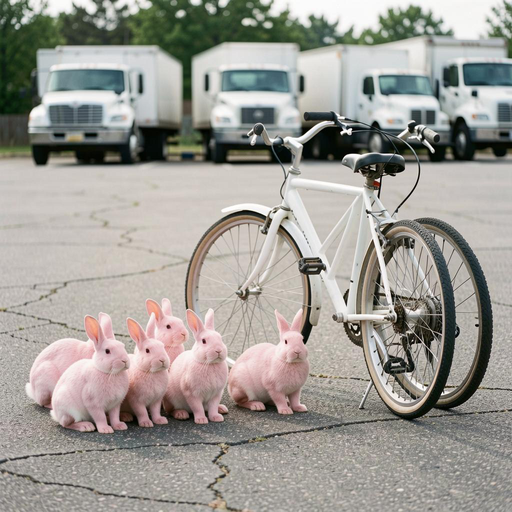}
& \includegraphics[width=0.185\linewidth]{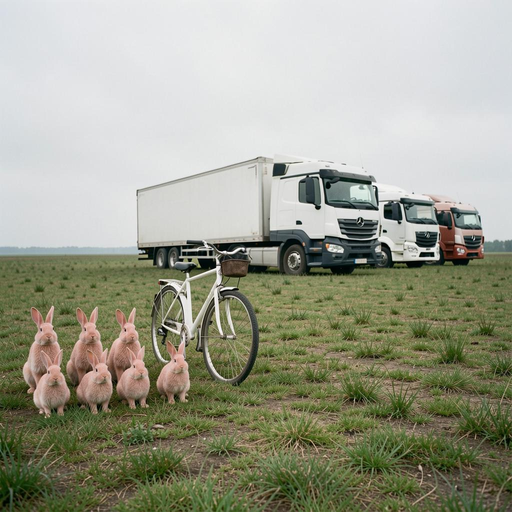} \\
\multicolumn{5}{p{0.96\linewidth}}{\centering
\textbf{Atomicity 8.} \textit{Prompt:} seven pink rabbits to the left of a white bicycle, and three trucks
} \\[0.55em]

\includegraphics[width=0.185\linewidth]{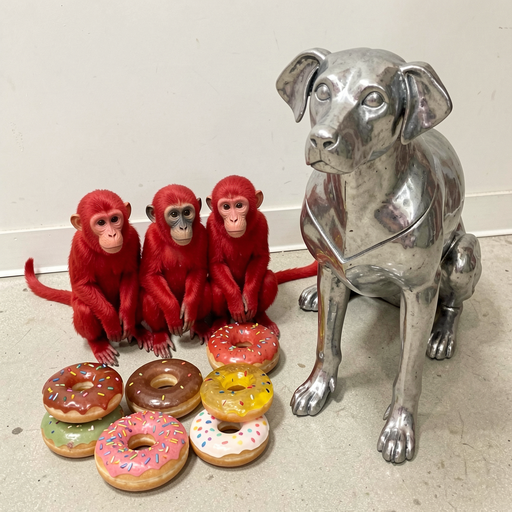}
& \includegraphics[width=0.185\linewidth]{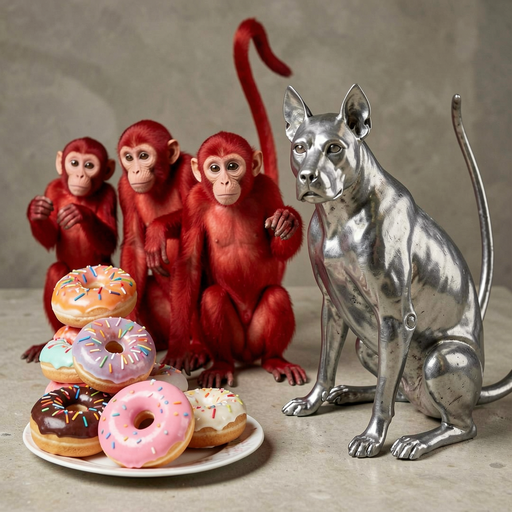}
& \includegraphics[width=0.185\linewidth]{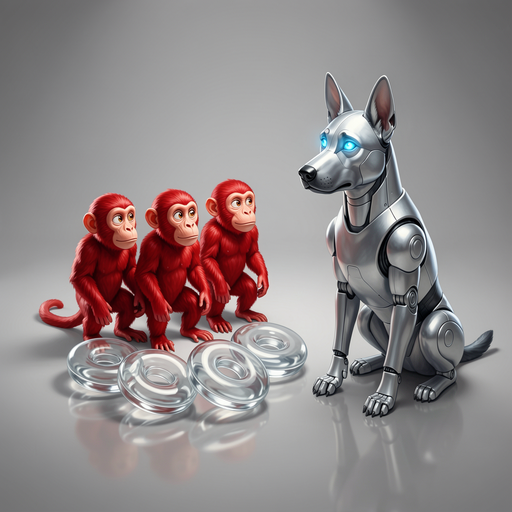}
& \includegraphics[width=0.185\linewidth]{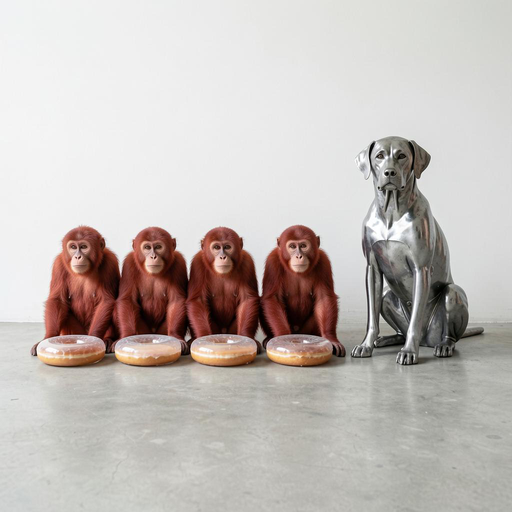}
& \includegraphics[width=0.185\linewidth]{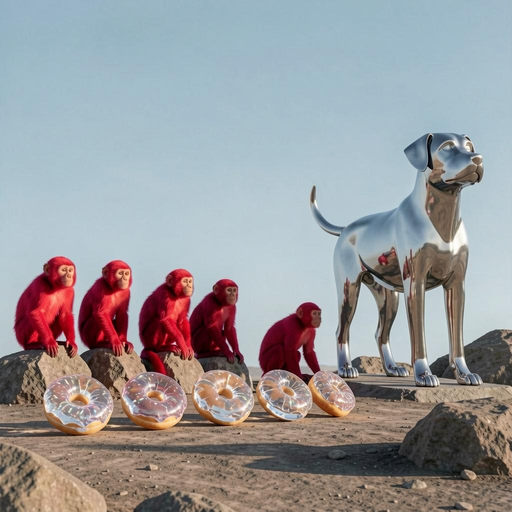} \\
\multicolumn{5}{p{0.96\linewidth}}{\centering
\textbf{Atomicity 9.} \textit{Prompt:} five red monkeys, and five glass donuts to the left of a metal dog
} \\[0.55em]

\includegraphics[width=0.185\linewidth]{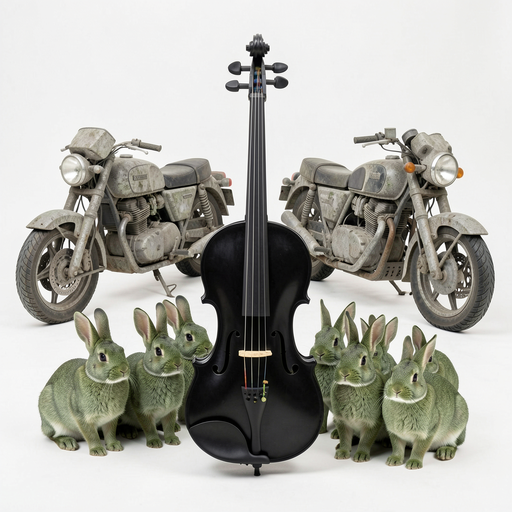}
& \includegraphics[width=0.185\linewidth]{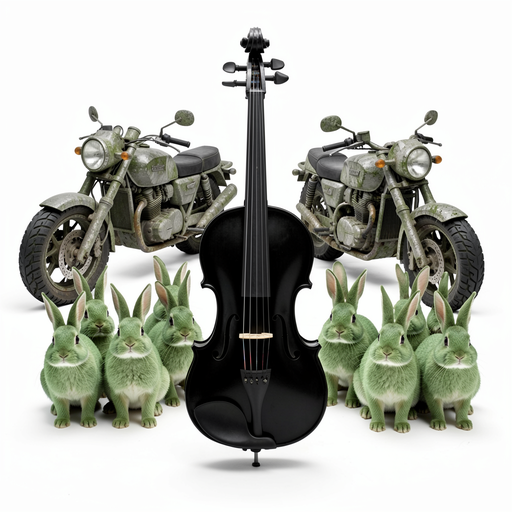}
& \includegraphics[width=0.185\linewidth]{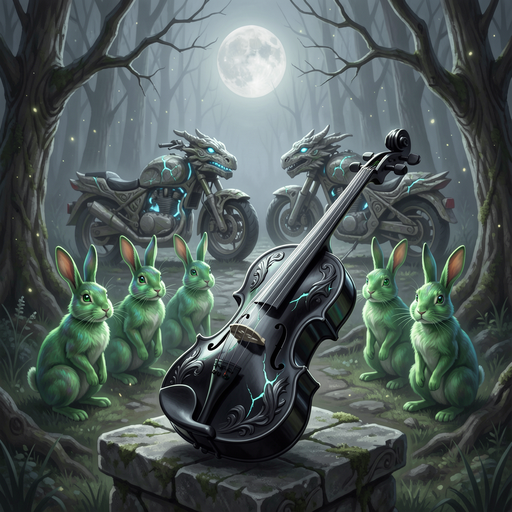}
& \includegraphics[width=0.185\linewidth]{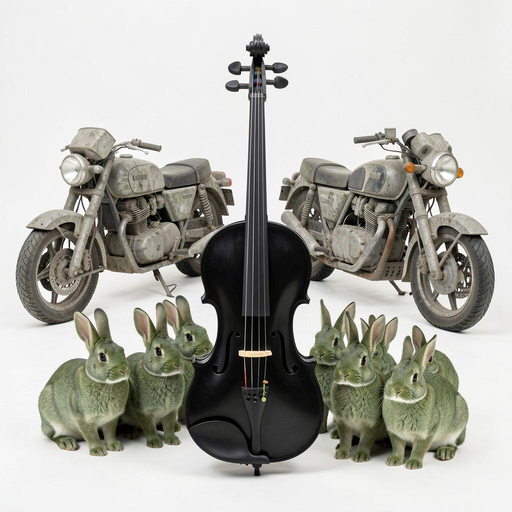}
& \includegraphics[width=0.185\linewidth]{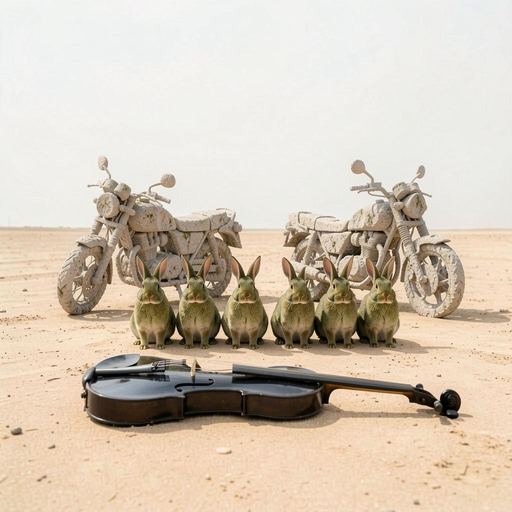} \\
\multicolumn{5}{p{0.96\linewidth}}{\centering
\textbf{Atomicity 10.} \textit{Prompt:} two stone motorcycles behind six green rabbits behind a black violin
} \\
\end{tabular}
\caption{
Qualitative GenEval2 examples across methods, ordered by prompt
atomicity. Each row uses the same prompt across five generation or
refinement methods. The examples illustrate differences in
satisfying object counts, attributes, and spatial relations
across methods.
}
\label{fig:qualitative_examples}
\end{figure}

Table~\ref{tab:geneval2_main} reports aggregate results.
Single-pass FLUX.2-Klein-9B achieves $34.16\%$
prompt-level accuracy despite $79.06\%$ atom-level accuracy,
reflecting the difficulty of satisfying multiple
requirements jointly.
\method{} raises these scores to $71.46\%$ and $90.25\%$,
respectively, improving prompt-level accuracy by
$37.30$ percentage points over single-pass generation.
Compared with RAISE ($52.23\%$ prompt-level accuracy),
\method{} also uses fewer image-model executions,
MLLM calls, and MLLM tokens.
Qualitative examples in
Figure~\ref{fig:qualitative_examples} illustrate improved
satisfaction of count, attribute, and relation requirements
within the same image.

\subsection{Ablation study}
\label{sec:ablation}

\begin{table}[t]
\centering
\caption{GenEval2 ablations at MLLM decoding temperature $T=0$.
All variants follow the setup in Sec.~\ref{sec:exp_setup} except
for the ablated component. $\Delta P$ is the gain in prompt-level
accuracy over single-pass FLUX.2-Klein-9B, in percentage points.}
\label{tab:geneval2_ablation}
\small
\setlength{\tabcolsep}{5.2pt}
\renewcommand{\arraystretch}{1.10}
\begin{tabular*}{\linewidth}{@{\extracolsep{\fill}}lcccccc@{}}
\toprule
\multirow{2}{*}{Method}
& \multicolumn{3}{c}{Accuracy (\%)}
& \multicolumn{3}{c}{Cost / prompt} \\
\cmidrule(lr){2-4}
\cmidrule(lr){5-7}
& GM $\uparrow$
& $\Delta P$ $\uparrow$
& AM $\uparrow$
& Exec. $\downarrow$
& Call $\downarrow$
& kTok $\downarrow$ \\
\midrule
FLUX.2-Klein-9B
& 34.16 & -- & 79.06
& 1.0 & -- & -- \\

\midrule
\method{}
& \best{71.46} & \best{+37.30} & \best{90.25}
& 13.6 & 2.53 & 9.2 \\

\makecell[l]{\quad w/o modular verifier}
& 63.73 & +29.57 & 87.74
& 5.75 & 7.94 & 21.0 \\

\makecell[l]{\quad Random policy}
& 65.89 & +31.73 & 89.16
& 16.5 & 3.76 & 12.2 \\

\quad w/o resampling
& 63.98 & +29.82 & 87.35
& 17.0 & 4.12 & 13.4 \\

\quad w/o editing
& 64.03 & +29.87 & 89.15
& 18.9 & 2.23 & 8.4 \\

\quad w/o rewrites
& 59.71 & +25.55 & 86.66
& 17.8 & 1.47 & 7.2 \\
\bottomrule
\end{tabular*}
\end{table}

Table~\ref{tab:geneval2_ablation} isolates the main components
under the same GenEval2, FLUX.2-Klein-9B, and $T=0$ setting.
Removing any component reduces prompt-level accuracy.
The largest drops occur when prompt rewrites are removed
($71.46\% \rightarrow 59.71\%$) or the modular verifier
is replaced with an MLLM verifier
($71.46\% \rightarrow 63.73\%$).
These results indicate that both prompt variation and
predicate-specific verification contribute to performance.
Replacing the structured refinement policy with a random policy
also reduces accuracy, as does removing resampling or editing,
supporting the use of
predicate states to select subsequent actions.
Appendices~\ref{appendix:geneval2_results}
and~\ref{appendix:additional_benchmarks} provide full cost and
ablation results, additional GenEval2 settings,
and other benchmark evaluations.

\subsection{Generalization and seed stability}
\label{sec:generalization}
\begin{table}[t]
\centering
\caption{Generalization and generation-seed stability.
(a) Mean per-prompt testpoint accuracy on UniGenBench++.
(b) GenEval2 prompt-level accuracy (Soft-TIFA GM), reported as
mean $\pm$ sample SD across five paired generation seeds.
All accuracy values are in percent.}
\label{tab:rebuttal_summary}
\small
\begin{subtable}[t]{0.46\linewidth}
\centering
\caption{UniGenBench++ (600 prompts).}
\label{tab:unigen_main}
\begin{tabular*}{\linewidth}{@{\extracolsep{\fill}}lr@{}}
\toprule
Method & Mean per-prompt acc. \\
\midrule
Direct & 81.07 \\
BoN+NVILA & 81.71 \\
T2I-Copilot & 85.80 \\
RAISE & 85.15 \\
\method{} & 83.34 \\
\bottomrule
\end{tabular*}
\end{subtable}
\hfill
\begin{subtable}[t]{0.50\linewidth}
\centering
\caption{Five GenEval2 seeds (240 prompts).}
\label{tab:seed_stability_main}
\begin{tabular*}{\linewidth}{@{\extracolsep{\fill}}lr@{}}
\toprule
Method & Prompt accuracy \\
\midrule
Direct & $35.69\pm1.89$ \\
T2I-Copilot & $34.41\pm0.73$ \\
RAISE & $49.39\pm2.64$ \\
\method{} & $65.88\pm2.32$ \\
\bottomrule
\end{tabular*}
\end{subtable}
\end{table}

On 600 natural-language prompts from
UniGenBench++~\citep{wang2025unigenbench}, \method{} achieves
$83.34\%$ mean per-prompt testpoint accuracy
(Table~\ref{tab:unigen_main}), above direct generation
and BoN+NVILA but below T2I-Copilot and RAISE.
None of the paired differences involving \method{} is
statistically significant after Holm correction.

On the 200 DrawBench~\citep{saharia2022drawbench} prompts,
\method{} improves full-prompt
VQAScore~\citep{li2024vqascore} from $0.856$ for direct
generation to $0.896$, close to RAISE's $0.898$.

Across five paired generation seeds on a fixed,
stratified subset of 240 GenEval2 prompts, \method{}
achieves $(65.88\pm2.32)\%$ prompt-level accuracy
(sample SD), compared with $(49.39\pm2.64)\%$ for RAISE,
and scores higher in all five seeds
(Table~\ref{tab:seed_stability_main}).

\begin{table}[t]
\centering
\caption{GenEval2 accuracy (\%) and inference cost with
Qwen-Image generation and editing backbones on 240 prompts.
Costs are realized averages per prompt; Exec. denotes generator or editor executions.
MLLM calls count Qwen3-VL-32B calls; NVILA reranking costs are not included.}
\label{tab:qwen_image_transfer}
\small
\setlength{\tabcolsep}{4pt}
\begin{tabular*}{\linewidth}{@{\extracolsep{\fill}}lrrrr@{}}
\toprule
Method & Prompt acc. & Atom acc. & Exec. & MLLM calls \\
\midrule
Direct Qwen-Image & 31.29 & 77.30 & 1.0 & -- \\
BoN+NVILA & 38.31 & 81.66 & 32.0 & -- \\
T2I-Copilot & 45.82 & 83.80 & 1.11 & 4.2 \\
RAISE & 54.12 & 86.85 & 20.1 & 9.3 \\
\method{} & \textbf{69.39} & \textbf{89.57} & 13.4 & 2.5 \\
\bottomrule
\end{tabular*}
\end{table}

Using Qwen-Image generation and editing
backbones~\citep{qwenimage2512,qwenimageedit2511}
in a 240-prompt GenEval2 evaluation
(Table~\ref{tab:qwen_image_transfer}), \method{} achieves
$69.39\%$ prompt-level and $89.57\%$ atom-level accuracy,
compared with $54.12\%$ and $86.85\%$ for RAISE.
Protocols and further results appear in
Appendix~\ref{appendix:generalization_robustness}.

\subsection{Parser and verifier reliability}
\label{sec:reliability}

An independent GPT-5.6 Sol audit examines whether each
compiled visual program preserves all requirements of
its original prompt.
Complete coverage is found for $781/800$ GenEval2 prompts
($97.63\%$) and $147/200$ DrawBench prompts ($73.50\%$).
Incomplete programs include omitted requirements and
incorrect predicate bindings, reflecting the dependence
of \method{} on the initial prompt parsing.

We also audit 1,000 saved candidate--predicate decisions,
stratified from 32,510 records across GenEval2 and DrawBench.
GPT-5.6 Sol independently judges each image and predicate
without access to the verifier's decisions.
Treating \texttt{satisfied} as the positive prediction and
\texttt{uncertain} or \texttt{violated} as negative,
the population-weighted precision and recall are
$91.7\%$ and $61.8\%$, respectively, after excluding
ambiguous reference judgments.
Recall is particularly low for action and semantic-role
predicates ($36.7\%$).
This pattern indicates that the verifier can leave
satisfied requirements unresolved, potentially leading
to unnecessary refinement or budget exhaustion.
Despite this limitation, \method{} uses fewer image-model
executions and MLLM calls than RAISE in the main GenEval2
comparison (13.6 vs. 19.8 and 2.53 vs. 9.12 per prompt).
Audit protocols, sampling weights, and results by
predicate family are provided in
Appendix~\ref{appendix:reliability_audits}.

\subsection{Visual quality and human evaluation}
\label{sec:quality}
\begin{table}[t]
\centering
\caption{Final-image Q-Align quality (1--5; higher is better).
GenEval2 and DrawBench contain 800 and 200 images per method,
respectively. The evaluator receives no prompt.}
\label{tab:qalign_main}
\small
\begin{tabular*}{0.82\linewidth}{@{\extracolsep{\fill}}lrr@{}}
\toprule
Method & GenEval2 & DrawBench \\
\midrule
Direct & 4.8265 & 4.7522 \\
BoN+NVILA & 4.8569 & 4.7703 \\
T2I-Copilot & 4.8427 & 4.7211 \\
RAISE & 4.7605 & 4.7935 \\
\method{} & 4.7934 & 4.6997 \\
\bottomrule
\end{tabular*}
\end{table}

We assess whether improvements in prompt alignment
are accompanied by changes in visual quality.
Q-Align~\citep{wu2023qalign} measures image quality
without conditioning on the prompt
(Table~\ref{tab:qalign_main}).
On GenEval2, \method{} scores below direct generation,
BoN+NVILA, and T2I-Copilot, but above RAISE;
all four differences are significant after Holm correction.
On DrawBench, it scores significantly below
BoN+NVILA and RAISE, while differences from direct
generation and T2I-Copilot are not significant.

To examine changes during refinement, we compare the
initial candidate with the selected output on GenEval2.
VQAScore increases by $0.0549$, whereas Q-Align decreases
by $0.0313$ on average.
These changes suggest a modest trade-off between
prompt alignment and visual quality.

We also conduct a blinded DrawBench study to assess
human judgments of visual quality and full-prompt alignment.
A total of 800 distinct image pairs receive three ratings
each on both criteria, yielding 2,400 paired evaluations
and 4,800 individual judgments.
Tie-adjusted preference for \method{} ranges from
$46.9\%$ to $52.0\%$ for quality and from $46.7\%$ to
$54.1\%$ for alignment across the four baselines.
No comparison shows a statistically significant difference
in either criterion after Holm correction.
Study procedures and detailed results appear in
Appendix~\ref{appendix:quality_evaluation}.

\section{Limitations and Broader Impacts}
\label{sec:limitations}

\method{} relies on accurate visual-program construction
and predicate verification.
Errors in the initial program can misdirect refinement,
while verifier false negatives can lead to unnecessary
computation and false positives can cause premature acceptance.
The supported predicate vocabulary and rule-based refinement
policy also limit the requirements that \method{} can handle,
and its effectiveness depends on the capabilities of the
underlying generator and editor.
Comparisons with learned refinement policies remain
an open direction.

Although \method{} achieves strong results on GenEval2,
its gains are less consistent on natural-language prompts.
Automated visual-quality evaluations also indicate
modest quality degradation in some settings, while
the human evaluation finds no statistically significant
differences in either visual quality or prompt alignment.
Appendix~\ref{appendix:additional_limitations}
discusses these limitations and potential safeguards.

Improved compositional controllability can benefit
creative, educational, and accessibility applications,
but may also facilitate the generation of misleading
or harmful synthetic content.
The underlying models may inherit biases from their
training data, motivating appropriate safety checks
and content provenance measures in deployment.

\section{Conclusion}
\label{sec:conclusion}

We introduced \method{}, a training-free inference-time
refinement framework that compiles a prompt into a fixed
visual program and uses predicate verification to guide
editing, resampling, and stopping.
On GenEval2, \method{} substantially improves prompt-level
accuracy over single-pass generation and outperforms
the strongest compared refinement baseline while using
fewer image-model executions, MLLM calls, and tokens.
These findings suggest that fixed prompt requirements
and predicate-level feedback provide a useful basis
for controlling inference-time refinement in
compositional text-to-image generation.

\clearpage
\begin{ack}
This work was supported by grants from the Institute of
Information \& Communications Technology Planning \&
Evaluation (IITP), funded by the Ministry of Science
and ICT (MSIT), Republic of Korea, through the
Artificial Intelligence Graduate School Program
(POSTECH; No. RS-2019-II191906),
the AI Research Hub Project (No. RS-2024-00457882),
the Information Technology Research Center (ITRC)
program (No. IITP-2026-RS-2024-00437866),
and the Development of Personalized Video Automatic
Generation and Insertion Technology project
(No. RS-2026-25511821).
This research was also supported by the Fundamental
Research Program (PNKB300) of the Korea Institute
of Materials Science, funded by MSIT.
The authors declare no competing interests.
\end{ack}

\bibliographystyle{plainnat}
\bibliography{references}

\newpage
\appendix
\section{Method Details}
\label{appendix:method_details}

This section describes the default rules for constructing the visual
program, verifying its predicates, and selecting refinement actions.
Model checkpoints, numerical thresholds, and other implementation
settings are provided in Appendix~\ref{appendix:implementation_details}.

\subsection{Visual-Program Construction}
\label{appendix:method_visual_program}

The parser converts the original prompt into object declarations and
typed predicates for counts, attributes, spatial relations, exclusions,
scene requirements, and visible text. The resulting program serves as
the fixed specification for all subsequent candidates. Because an error
in this specification could affect the entire refinement process,
\method{} validates the program before generation begins.

Compilation comprises MLLM parsing, deterministic normalization and
validation, and conditional MLLM review. Normalization first removes
representation artifacts that do not change the interpreted visual
requirements. Examples include removing prompt framing phrases
mistakenly parsed as objects, canonicalizing descriptive fields without
changing their predicate types, and resolving duplicate predicate
identifiers. These cleanup operations do not themselves trigger review.

Other corrections may affect the predicate structure and are therefore
recorded as fixes. They include removing duplicate object declarations,
normalizing relation or attribute names, removing invalid self-relations,
and adding declarations or existence requirements for referenced objects.
Normalization also resolves redundant count constraints and conflicts
between positive requirements and exclusions, including exclusion
constraints whose object labels duplicate those of required objects.
When an unsupported relation or attribute name cannot be resolved
deterministically, the validator records a warning instead of silently
changing the requirement.

An MLLM reviewer is invoked only if the normalization and validation
report contains a fix or warning. It examines the normalized program
against the source prompt and either approves it or returns a corrected
program. The reviewed program is normalized once more before being
finalized. Thus, review is conditional on a detected issue rather than
performed at every refinement round.

The prompt rewriter separately requests eight alternative phrasings of
the original prompt. Empty, duplicate, and source-identical rewrites
are discarded.
If none remain, the source prompt is used as the sole pool entry.
Thus, the final pool $\mathcal{R}=\{r_j\}_{j=1}^{M}$ contains
between one and eight prompts. An initial prompt is selected
deterministically using the program identifier and seed. Subsequent
resampling prefers unused rewrites and reuses the pool when necessary.
A single scene instruction is appended unless the prompt explicitly
requests a grid or multiple panels.

\subsection{Predicate-State Computation}
\label{appendix:method_predicate_states}

For a candidate image, each predicate is evaluated using the visual
evidence appropriate to its type. The verifier returns a state in
$\{\mathrm{satisfied},\mathrm{uncertain},\mathrm{violated}\}$ and a
verification score where applicable. Scores support state assignment,
action selection, retry decisions, and fallback ranking. The verifier
may reuse detections, masks, image features, and depth estimates across
predicates, but evaluates each requirement separately.

\paragraph{Object selection and spatial relations.}
For an object with a positive exact count of $N$, the verifier selects
up to $N$ detections with the highest confidence. Otherwise, it uses the
highest-confidence detection. A spatial predicate is evaluated from the
selected detections of its subject and reference objects. Their masks
are combined by union; a bounding box is rasterized when a valid mask
is unavailable. The resulting object footprint has mask $M_o$, tight
bounding box $B_o=(x_o^-,y_o^-,x_o^+,y_o^+)$, and centroid
$c_o=(c_o^x,c_o^y)$. For depth relations, $d_o$ is the arithmetic mean
of the regionwise mean depths, not the pixel mean over the union mask.
Depth preprocessing is specified in
Appendix~\ref{appendix:implementation_verifier}.

Define $\operatorname{clip}(x)=\min(1,\max(0,x))$. The normalized overlap
of one-dimensional intervals is
\[
\begin{aligned}
    \Omega([a,b],[c,d]) &= \\
    &\operatorname{clip}\!\left(
    \frac{\max(0,\min(b,d)-\max(a,c))}
    {\max(1,\min(b-a,d-c))}\right).
\end{aligned}
\]
For subject and reference masks, $J$ denotes mask intersection over
union, $I$ denotes overlap relative to the smaller mask, and $C$ denotes
containment conflict:
\[
\begin{aligned}
    J &= \frac{|M_s\cap M_r|}{|M_s\cup M_r|}, \\
    I &= \frac{|M_s\cap M_r|}{\min(|M_s|,|M_r|)}, \\
    C &= \max\!\left(
      \frac{|M_s\cap M_r|}{|M_s|},
      \frac{|M_s\cap M_r|}{|M_r|}\right).
\end{aligned}
\]
Each relation is assigned a score $q_r\in[0,1]$ using the rules in
Table~\ref{tab:spatial_relation_scoring}. The default state thresholds
are $0.60$ for satisfaction and $0.35$ for uncertainty: a score below
$0.35$ is treated as violated.

\FloatBarrier
\begin{table*}[!htbp]
\centering
\caption{Spatial relation scores used by the default verifier.
All scores are clipped to $[0,1]$. The constants are those used
in the main experiments.}
\label{tab:spatial_relation_scoring}
\footnotesize
\begin{tabularx}{\textwidth}{p{0.18\textwidth}X}
\toprule
Relation & Verification score \\
\midrule
Left/right/\newline above/below &
Let $a$ be the requested axis and $b$ the orthogonal axis, with
image lengths $L_a,L_b$. Signed centroid and edge gaps
$\Delta_c,\Delta_e$ increase when the required order holds.
For left of, $\Delta_c=c_r^x-c_s^x$ and
$\Delta_e=x_r^- - x_s^+$.
Define $u_c=\operatorname{clip}(0.5+\Delta_c/(0.50L_a))$,
$u_e=\operatorname{clip}((\Delta_e+0.02L_a)/(0.14L_a))$, and
$u_b=\operatorname{clip}(0.75\Omega_b+
0.25\operatorname{clip}(1-|\Delta_b|/(0.55L_b)))$,
where $\Omega_b$ is orthogonal interval overlap and
$\Delta_b=c_r^b-c_s^b$.
The base score is
$q_{\mathrm{base}}=\operatorname{clip}(0.45u_c+0.35u_e+0.20u_b)$.
A weighted penalty $p$ combines overlap along the requested axis,
mask overlap $I$, and containment conflict $C$, using weights
$0.05$, $0.45$, and $0.30$, respectively. For left/right relations,
it additionally incorporates support contact $u_{\mathrm{on}}$
with weight $0.35$. The score is
$q_r=\operatorname{clip}(q_{\mathrm{base}}(1-p))$. \\
\addlinespace
Near &
Let $D$ denote the image diagonal, and let
$\delta_{\mathrm{edge}}$ be the Euclidean distance between boxes
after overlapping horizontal and vertical gaps are set to zero.
Then $q_r=\operatorname{clip}(
0.45\operatorname{clip}(1-\|c_s-c_r\|/(0.18D))+
0.55\operatorname{clip}(1-\delta_{\mathrm{edge}}/(0.18D)))$. \\
\addlinespace
Inside/in &
Both predicates use bounding box containment:
$q_r=\operatorname{clip}(
0.80|B_s\cap B_r|/|B_s|+
0.20\mathbf{1}[c_s\in B_r])$. \\
\addlinespace
Overlapping/\newline intersecting &
The score combines mask intersection over union and overlap
relative to the smaller mask:
$q_r=\operatorname{clip}(0.65J+0.35I)$. \\
\addlinespace
On &
For each column $x$ where the subject and reference boxes overlap,
let $b_s(x)$ be the lowest subject mask pixel, and set
$\tau=\max(1,0.02H)$ for image height $H$.
Among reference pixels with $y\geq b_s(x)-\tau$, let
$y_r^*(x)$ be the topmost one.
A column is valid when both $b_s(x)$ and $y_r^*(x)$ exist.
For valid columns, define
$\ell_x=\operatorname{clip}(1-|y_r^*(x)-b_s(x)|/\tau)$.
Let $\overline\ell$ be the mean of $\ell_x$ over valid columns,
and let $v$ be the fraction of horizontally overlapping box columns
that are valid.
Then $u_{\mathrm{on}}=\operatorname{clip}(0.70\overline\ell+0.30v)$,
$q_{\mathrm{base}}=\operatorname{clip}(
0.55u_{\mathrm{on}}+0.25\Omega_x+0.20q_{\mathrm{above}})$,
and $q_r=\operatorname{clip}(
q_{\mathrm{base}}-0.25|M_s\cap M_r|/|M_s|)$. \\
\addlinespace
In front of/\newline behind &
If either object lacks depth evidence, the score is zero.
Otherwise, let $\Delta_d$ be positive when the requested depth
order holds, $A=\max(\Omega_x,\Omega_y)$, and $q_{\mathrm{near}}$
be the near score.
With $u_s=\operatorname{clip}(0.60A+0.40q_{\mathrm{near}})$,
the score is
$q_r=\operatorname{clip}(
\operatorname{clip}(\Delta_d/28.0)(0.80+0.20u_s))$. \\
\bottomrule
\end{tabularx}
\end{table*}

\paragraph{Counts and attributes.}
Object counts are obtained from detected instances after removing
duplicate proposals. Two confidence levels yield strong and weak
counts. An exact count is satisfied when both agree with the required
number, uncertain when the target lies between them, and violated
otherwise. Existence and lower-bound requirements are evaluated from
the same detections. Further detection and filtering settings are
given in Appendix~\ref{appendix:implementation_verifier}.

Attribute predicates compare selected object crops against text queries
for appearance, pattern, material, pose, state, and actions. The lowest
score across the selected object instances determines the aggregate
attribute score. Action predicates initially follow this region and
text scoring procedure. Once all other predicate families cease to
block acceptance, a crop-based MLLM verifier evaluates the action
requirements directly. Its categorical judgments replace the initial
scores and states according to the rules in
Appendix~\ref{appendix:implementation_verifier}.

\paragraph{Scene, exclusion, and visible text.}
Scene predicates are first evaluated from detections of scene or
background regions. When suitable detections are unavailable, the
verifier uses image and residual background scores. Exclusion checks
search for forbidden objects and discount detections that can be
explained by a required object's attribute. A residual detection still
contributes evidence of a violation. Visible text is evaluated by an
MLLM using the requested string and candidate image rather than by
exact OCR matching. The score aggregation and state mapping for these
checks are specified in
Appendix~\ref{appendix:implementation_verifier}.

\subsection{Refinement Policy}
\label{appendix:method_refinement_policy}

When several predicates block acceptance, the controller orders them
by their verification states and predicate types. Violated predicates
precede uncertain predicates. Within each state, the type order is
count, spatial relation, attribute, exclusion, global scene, and
visible text. Lower verification scores break remaining ties. The
controller considers predicates in this order and applies the first
eligible refinement rule; a predicate can be skipped when the
corresponding edit requires conditions that are not yet met.

The ordering reflects dependencies between corrections. Missing or
incorrect object counts can prevent meaningful evaluation of other
properties. Layout changes are considered before attribute changes
because resampling may replace or move existing objects. Exclusions
follow attributes within the same severity level, but removal of a
forbidden object has no separate requirement that other objects be
stable. Scene changes are attempted only after the required foreground
composition is established. Visible text is placed last because it
generally requires a different correction mechanism.

Editing is reserved for failures for which a localized modification is
supported by the current image. Specialized addition and removal rules
apply to exact count predicates only, and require plausible detections
for every required object category. The controller can add missing
instances of an already detected category or remove a single excess
instance. It resamples when a required category is absent, when more
than one excess instance would need removal, or when a lower-bound
count fails. Exclusion failures can independently trigger removal of a
forbidden object. Attribute edits require satisfied count predicates
for the subject and, for directed actions, the target object. Spatial
relation errors normally trigger resampling instead of repositioning
objects. A scene edit is considered only when object, count, attribute,
spatial, and exclusion requirements no longer block acceptance.

The policy also restricts repeated edits on the same failure for a
given rewrite. A removal operation is attempted at most once for a
matching target and operation. For an unsuccessful attribute or scene
edit, its verified output serves as the reference for a possible retry.
A retry requires exactly one matching failed edit and an improvement
in the target state or satisfaction, or an increase of at least $0.12$
in the mean target verification score relative to that output.
Object addition follows a different rule: another attempt is permitted
only if the remaining count deficit has decreased from that recorded
at the previous failed attempt. Once a predicate is satisfied, it no
longer blocks acceptance. Otherwise, the controller resamples when
further editing is not eligible.

Edit instructions refer to the selected requirement and ask the editor
to limit unrelated changes. Addition instructions specify the missing
instances and any relevant spatial relations. Removal instructions
target excess or forbidden objects while retaining required instances.
Attribute edits request a change to the target property, whereas scene
edits ask for a change to the background or environment while keeping
the subjects intact. Representative templates are given in
Appendix~\ref{appendix:implementation_editing_mllm}.

\subsection{Limited Auditor Override and Fallback}
\label{appendix:method_uncertainty_fallback}

A limited MLLM auditor handles cases in which the modular verifier
continues to reject a candidate despite potentially adequate visual
evidence. The auditor is eligible only when every remaining blocking
predicate belongs to an allowed family and has remained unsatisfied
for at least two consecutive rounds within the current trajectory.
A fresh resampling action resets this persistence history. Repeated
calls for an unchanged set of eligible blocking predicate identifiers
are suppressed.

Eligible requirements include color, material, shape, pattern, pose,
state, actions, and contrastive size attributes; global scene
requirements; and the spatial relations \emph{in}, \emph{inside}, and
\emph{on}. Contrastive size is eligible only when it compares at least
two distinct object identifiers with at least two distinct size values.
Counts, exclusions, visible text, and other spatial relations,
including directional and depth relations, are not eligible.
The auditor receives the candidate image, original prompt, and
remaining failed requirements. It can override either uncertain or
violated predicate states, but accepts the candidate only if it judges
all remaining requirements to be satisfied.

If no candidate satisfies the program gate or receives an auditor
override within the execution budget, \method{} returns a fallback
from the verified candidate history. Candidates are compared
lexicographically by the numbers of satisfied count, spatial, visible
text, exclusion, attribute, and global scene predicates, in that order.
Remaining ties are broken by the total number of satisfied predicates,
the mean verification score of unsatisfied predicates, and the earlier
refinement round. An unavailable verification score contributes zero
to this mean. Fallback outputs are recorded as budget exhausted rather
than accepted candidates.

\subsection{Tool Extensions and Controller Design}
\label{appendix:controller_extensions}

New detectors, segmenters, depth estimators, text recognizers, or
region--text scorers can replace verifier backends if they preserve
the predicate-state interface. A new predicate family additionally
needs typed arguments, an evidence-to-state mapping, dependencies,
and routing rules.

The deterministic policy allows the ablations in
Table~\ref{tab:geneval2_ablation_full} to isolate modular feedback
and structured routing without controller training. These ablations
do not establish optimality or replace a comparison with a learned
policy. We have not compared against an RL-trained controller or
generator under matched conditions.

Learning action selection and stopping from recorded trajectories
is a direction for future work.

\section{Implementation Details}
\label{appendix:implementation_details}

We describe the default configuration used in the main experiments.
Unless stated otherwise, ablations change only the component under
study while preserving the remaining setup. The provenance of the
archived reliability audit is discussed separately in
Appendix~\ref{appendix:verifier_audit}.

\subsection{Model and Runtime Setup}
\label{appendix:implementation_runtime}

\paragraph{Image models and execution budget.}
In the main GenEval2 comparison, each independent run is limited to
$B=32$ image-model
executions. Both initial generation and subsequent generation or editing
calls count toward this budget. We use
\texttt{black-forest-labs/}\allowbreak\texttt{FLUX.2-klein-9B} for both generation and
image-conditioned editing at $1024\times1024$ resolution. Both modes
use four inference steps with guidance scale $1.0$.
FLUX.2-dev~\citep{flux2dev} is used only as a separate single-pass reference.

\paragraph{Baseline configuration.}
For T2I-Copilot and RAISE, we use their official implementations and
replace the interfaces to the image model and MLLM with the same
FLUX.2-Klein-9B and Qwen3-VL-32B backends used by \method{}, where
applicable. Their agent loops, stopping rules, refinement policies,
and candidate selection procedures follow the respective baseline
implementations. This comparison holds the principal model backbones
fixed while retaining the methods' distinct control strategies.

\paragraph{Sampling and prompt variation.}
The default base seed is $s_{\mathrm{base}}=42$ for the first independent
sample of each prompt. For independent sample index $i$ beginning at
zero, the base seed is $42+i$. An image-model execution at round $t$,
indexed from one, uses
\[
    z_t = \operatorname{RandInt}_{s_{\mathrm{base}}+t-1}
    (0,2^{31}-1),
\]
where $\operatorname{RandInt}_s$ denotes a pseudorandom integer drawn
from a generator initialized with seed $s$. This schedule is shared by
generation and editing, rather than assigning distinct offsets to the
two operations. The random generator is instantiated on the CPU for
generation and on CUDA for editing when available.

At the start of each run, an MLLM rewriter requests eight alternative
prompts intended to preserve the source requirements. Empty, duplicate,
and source-identical outputs are discarded, so the final number of
usable rewrites can be smaller than eight. Initial generation and
resampling select from this pool as described in
Appendix~\ref{appendix:method_visual_program}. If no usable rewrite
remains, the original prompt is used. Each resampling call generates a
new image with its own seed; no intermediate denoising trajectory is
reused.

\paragraph{Ablation configurations.}
The default system combines modular predicate verification with a
rule-based refinement policy. Our ablations replace the modular verifier
with an MLLM verifier, randomize the policy, disable resampling or
editing, or remove prompt rewriting. In the \emph{Random policy} variant,
the target predicate and action are sampled at random. Disabling
resampling retains the initial generation but restricts subsequent
image-model calls to editing. Other variants are defined by the
corresponding component changes described in Sec.~\ref{sec:ablation}.

\paragraph{Hardware and runtime measurement.}
Development and evaluation used NVIDIA RTX A6000 48GB and RTX 6000 Ada
Generation 48GB GPUs. Each image generation pipeline ran on one 48GB
GPU, while the shared Qwen3-VL-32B vLLM service ran separately on two
48GB GPUs. The runtime values in
Table~\ref{tab:geneval2_main_full_cost} correspond to the main GenEval2
$T=0$ full cost log collected with the image pipeline on an RTX 6000
Ada Generation GPU. We omit timings for experiments conducted across
heterogeneous machines.

The recorded runtime includes image generation and editing, predicate
verification, auditor requests, client-observed MLLM latency including
queueing, and local initialization or loading performed during the run.
The timer ends after saving the selected candidate and preceding
internal artifacts, but before the final summary and result logs are
serialized. Starting the separately hosted vLLM service is not
included in the measurement.

\subsection{Editing and MLLM Components}
\label{appendix:implementation_editing_mllm}

\paragraph{Image-conditioned editing.}
Editing uses the native instructional editing mode of the FLUX.2-Klein
checkpoint. Given the current image $I_t$, selected predicate, and
instruction $u_t$, the editor returns
\[
    I_{t+1}=E_{\theta}(I_t,u_t;z_t),
\]
where $z_t$ is the round seed. The input is converted to RGB, and the
output retains the original resolution. We do not supply a separate
inpainting mask, bounding box, or denoising strength. Instead, the
instruction identifies the intended change and requests that the
remaining image content be retained.

Table~\ref{tab:edit_instruction_templates} illustrates the principal
instruction patterns. The exact wording depends on the predicate and
its object relations. For example, addition instructions can include
a relation involving either the added subject or the reference object;
removal instructions omit count wording for an exclusion predicate.
All variants append instructions to preserve the existing subjects,
framing, and other requirements not targeted by the edit.

\begin{table}[t]
\centering
\caption{Representative instructions for the four editing operations.
The full instruction depends on the selected predicate and includes
clauses that discourage changes outside the target requirement.}
\label{tab:edit_instruction_templates}
\footnotesize
\begin{tabularx}{\linewidth}{p{0.20\linewidth}X}
\toprule
Operation & Representative instruction \\
\midrule
Add object &
``Add $\langle n\rangle$ more $\langle\mathrm{object}\rangle$ so that
$\langle\mathrm{count\ requirement}\rangle$.''
If relevant, specify the required relation to another object.
\\
\addlinespace
Remove object &
``Remove only the extra $\langle\mathrm{object}\rangle$, preferably
a secondary or background instance.''
For exact count correction, also specify the remaining required count.
For an exclusion, remove the forbidden object without requiring a
count target; distinguish duplicates from the primary subject.
\\
\addlinespace
Change attribute &
``Change the $\langle\mathrm{object}\rangle$ so that it is
$\langle\mathrm{value}\rangle$.''
The requested property can be a color, material, shape, pattern, size,
pose, visible state, or action involving another object.
\\
\addlinespace
Change scene &
``Change only the background and surrounding environment so the
scene clearly reads as $\langle\mathrm{scene\ value}\rangle$.''
Preserve the existing subjects, their arrangement, and other
required properties.
\\
\bottomrule
\end{tabularx}
\end{table}

\paragraph{MLLM components.}
All language and multimodal components use
\texttt{Qwen/}\allowbreak\texttt{Qwen3-VL-32B-Instruct} through an OpenAI-compatible vLLM
endpoint. The main GenEval2 configuration uses temperature $T=0$
and seed $42$. The model is used for initial parsing, conditional
program review, prompt rewriting, visible-text verification, late
verification of action predicates, and the limited auditor override.
Parser, reviewer, rewriter, and auditor requests use structured
function calling. If a structured request is retried, its seed is
incremented by the retry index.

The visible-text and action verifiers use temperature zero, a timeout
of $180$ seconds, and input images resized to a maximum side length of
$1024$ pixels. Their output limits are $96$ and $128$ tokens,
respectively.

\paragraph{MLLM prompt templates.}
Table~\ref{tab:mllm_prompt_templates} summarizes the default prompt
templates and structured outputs for the MLLM components used for
visual-program construction, prompt rewriting, and the limited auditor. The parser system
prompt additionally includes few-shot examples instantiating the same
schema.

\begin{table}[t]
\centering
\caption{Default MLLM prompt templates used by \method{} for visual
program construction, prompt rewriting, and the limited auditor.}
\label{tab:mllm_prompt_templates}
\footnotesize
\begin{tabularx}{\linewidth}{p{0.18\linewidth}X}
\toprule
Component & Prompt template and output schema \\
\midrule
Visual-program parser &
\textbf{System instruction:} ``You are a visual-program compiler for
text-to-image prompts.'' Compile the user prompt into a fixed visual
program for deterministic visual checks. Preserve explicit object
identity, counts, attributes, including action-related attributes and
their targets, spatial relations, global scene constraints, visible text,
and relation direction. Use supported predicate names only; declare
canonical objects once; represent multiplicity through count constraints
rather than per-instance objects unless the prompt explicitly distinguishes
instances. \textbf{User template:} ``Compile the following user prompt
into the structured visual program schema. user\_prompt:
\(\langle\text{prompt}\rangle\).'' \textbf{Output schema:}
\texttt{parser\_reasoning}, \texttt{source\_prompt},
\texttt{objects}, \texttt{at\_least\_count\_constraints},
\texttt{exclusion\_constraints}, \texttt{exact\_count\_constraints},
\texttt{relation\_constraints}, \texttt{attribute\_constraints},
\texttt{global\_scene\_constraints}, and \texttt{text\_constraints};
empty buckets are returned as empty lists. Object records include
\texttt{object\_id}, \texttt{label}, optional
\texttt{proposal\_text}, \texttt{aliases}, and \texttt{description};
constraint records contain the referenced object IDs plus the relevant
count, relation, attribute, scene value, or visible-text fields. \\
\addlinespace
Visual-program review and repair &
\textbf{System instruction:} ``You are a reviewer and repairer for a
structured visual program that was compiled from a text-to-image
prompt.'' Treat the normalized candidate as a starting point, not as
ground truth. Check object declarations, object-ID consistency, count
attachment, action-related attribute targets, relation direction,
supported predicate names, self-relations, self-targeted action-related
attributes, unsupported type/size
attributes, and exclusions that are not explicit absence requests.
\textbf{User template:} provide \texttt{original\_prompt},
\texttt{normalization\_report}, and
\texttt{candidate\_visual\_program}. \textbf{Output schema:}
\texttt{approved\_candidate},
\texttt{review\_reasoning}, \texttt{detected\_issues}, and
\texttt{reviewed\_program}. If the candidate is already correct, return
an equivalent reviewed program; otherwise return a repaired program. \\
\addlinespace
Prompt rewriter &
\textbf{System instruction:} ``You are a prompt rewriter for a
text-to-image model.'' Given one original prompt, return exactly
\(N\) rewritten prompts that are more descriptive and visually concrete
while preserving the exact meaning. Preserve every explicit detail
including object identity, count, attributes, including action roles,
spatial relations, background, visible text, medium, and style. Do not add
new salient objects or object-level attributes that are not stated or
directly implied. Prefer one coherent scene and vary style, framing,
mood, or scene detail across rewrites. \textbf{User template:}
``Original prompt: \(\langle\text{prompt}\rangle\). Return exactly
\(N\) rewritten prompts as JSON with one key named
\texttt{rewritten\_prompts}.'' \textbf{Output schema:}
\texttt{rewritten\_prompts}: list of strings. \\
\addlinespace
Limited in-loop auditor &
\textbf{System instruction:} ``You are reviewing one candidate image
against a short list of visual checks.'' Decide whether every listed
check is visually satisfied while the image still matches the original
prompt. Treat the listed checks as the focus, but fail if broader prompt
mismatches make the image clearly wrong. Prefer failure when a listed
color, material, pattern, shape, size, action-related attribute, relation,
or visible object is wrong or too unclear to verify confidently.
\textbf{User template:} provide the candidate image,
\texttt{user\_prompt}, and \texttt{checks\_to\_verify}.
\textbf{Output schema:}
\texttt{all\_checks\_passed}, \texttt{short\_reason}, and
\texttt{check\_reasoning}. \\
\bottomrule
\end{tabularx}
\end{table}

\subsection{Verifier Backends and Thresholds}
\label{appendix:implementation_verifier}

The default verifier combines detector outputs, region and text
similarity, geometric measurements, and a limited set of MLLM checks.
Shared visual features and detections are cached across predicates for
each candidate image.

\paragraph{Object proposals and counts.}
We use SAM3~\citep{carion2025sam3} with text prompting to detect and segment object instances,
with a minimum proposal confidence of $0.30$. Overlapping proposals for
the same object are deduplicated when their intersection divided by
the smaller mask area is at least $0.85$, using box overlap when masks
are unavailable. Connected duplicates are merged by the union of their
masks or boxes, and their confidence is set to the maximum of the merged
proposals. Nearly identical proposals assigned to different object
identifiers are further resolved in favor of the higher-confidence
proposal when intersection over the smaller region is at least $0.95$,
intersection over union is at least $0.90$, and the ratio of the smaller
to larger area is at least $0.85$. No additional mask area threshold
is applied to counting proposals.

Count states are based on the strong and weak detection counts
described in Appendix~\ref{appendix:method_predicate_states}.
Spatial and attribute verification use selected instances from these
proposals. For an object with a positive exact count of $N$, up to the
$N$ highest-confidence instances are selected; otherwise, the
highest-confidence instance is used.

\paragraph{Region and text scoring.}
We use \texttt{google/siglip2-base-patch16-224} to evaluate object
attributes and provide image and scene similarity scores. For each
selected instance, the bounding rectangle of its mask is expanded by
$20$ pixels and cropped from the original RGB image. The background
within the crop is retained, and black padding is applied to obtain a
square input. If no valid mask is available, the detection box defines
the crop.

For each attribute query $q$, we evaluate three text formulations:
``a photo of $q$,'' ``a close-up photo of $q$,'' and ``$q$.''
The sigmoid of each model logit is used as a compatibility score.
We take the maximum across formulations and alternative queries for
the same attribute. When the requirement applies to multiple selected
instances, the minimum score across instances is used. These scores
are mapped to predicate states using the family-specific thresholds
listed below.

\paragraph{Depth evidence.}
Depth-sensitive relations use MiDaS DPT Large~\citep{ranftl2022midas,ranftl2021dpt}. The predicted depth map
is interpolated to the image resolution, normalized to its own minimum
and maximum, inverted, and converted to an eight-bit range of
$[0,255]$, where smaller values indicate nearer regions. This is a
relative depth representation, not metric depth. The depth assigned to
an instance is the mean over its mask or, when no mask is available,
its bounding box. For an object represented by multiple instances,
these region means are averaged equally. Spatial relation scores use
this representation as specified in
Table~\ref{tab:spatial_relation_scoring}.

\paragraph{Action verification.}
Action requirements initially use the same SigLIP2 scores as other
attributes. Once all non-action requirements are satisfied, an MLLM
verifies each remaining action directly from a crop. For a directed
action, the crop covers the union of the subject and target regions,
with padding equal to $10\%$ of the union box dimensions on each side,
clipped to the image boundaries. Actions without a target use the
subject box. The MLLM maps \texttt{true}, \texttt{false}, and
\texttt{null} to satisfied, violated, and uncertain, with corresponding
scores $1$, $0$, and $0.5$. For multiple subject instances, any
violation makes the aggregate predicate violated; otherwise, an
uncertain instance makes it uncertain.

Visible text is also evaluated by an MLLM returning a categorical
judgment, rather than by exact OCR matching. Its \texttt{null}
judgment corresponds to uncertain, but has no continuous score.

\paragraph{Scene verification.}
Scene predicates first consider SAM3 proposals corresponding to the
requested scene or background. Let $W,H$ denote image dimensions. The
union of foreground masks is dilated by
\[
    r=\max(4,\operatorname{round}(0.01\min(W,H)))
\]
pixels. A scene proposal is retained if its region outside the
dilated foreground occupies at least $10\%$ of the image.
When no qualifying proposal is found, the verifier evaluates the full
image using SigLIP2. It also evaluates a residual background image
when that region covers at least $10\%$ of the image. The residual
image is formed by replacing foreground pixels with the mean RGB
color of the background. Full-image and residual scores are combined
with weights $0.45$ and $0.55$, renormalized over available terms.
Proposal evidence uses satisfaction and uncertainty thresholds
$0.65/0.35$; the SigLIP2 fallback uses $0.60/0.35$.

\paragraph{Exclusion verification.}
A forbidden category is checked using its corresponding proposals.
Some apparent forbidden detections may instead describe attributes of
required objects. To avoid treating them as independent instances, the
verifier identifies positive objects whose attribute values share
tokens with the forbidden label or its aliases. A forbidden proposal
is compared with these positive regions and treated as overlapping
attribute evidence only when intersection over the smaller region is
at least $0.90$, area similarity is at least $0.70$, and intersection
over union is at least $0.75$.

Let $\rho$ denote the fraction of the forbidden proposal not explained
by the overlapping positive region, and let $c$ be its confidence.
Its effective confidence is zero if $\rho\leq0.10$, and $c\rho$
otherwise. The maximum remaining effective confidence determines the
state: satisfied below $0.35$, uncertain from $0.35$ up to $0.65$,
and violated at or above $0.65$.

\paragraph{Decision thresholds.}
The other default satisfaction and uncertainty thresholds are
$0.65/0.35$ for object evidence, $0.60/0.35$ for attributes,
$0.55/0.35$ for initial action scores, and $0.60/0.35$ for
spatial relations. The initial action thresholds are not used for the
late MLLM judgments, which are mapped directly to discrete states.
The color attribute verifier uses SigLIP2 scores without a separate
color classifier.

\FloatBarrier
\section{Additional GenEval2 Results}
\label{appendix:geneval2_results}

\subsection{Main Full Cost Log}
\label{appendix:geneval2_main_full_cost}
\begin{table}[!htbp]
\centering
\caption{Full GenEval2 cost log at $T=0$.
FLUX.2-dev~\citep{flux2dev} is a separate single-pass reference.
All wall-clock seconds in this table were measured under the same
runtime setup on an RTX 6000 Ada Generation 48GB GPU.
Cost is reported per prompt; \emph{Exec.} counts image-model executions,
kTok in/out reports MLLM input/output tokens in thousands, and Img.
counts MLLM image inputs.
For cost columns, best and second-best values among inference-time
methods are shown in \textbf{bold} and \underline{underlined},
respectively; missing cost entries are not ranked.}
\label{tab:geneval2_main_full_cost}
\scriptsize
\setlength{\tabcolsep}{2.4pt}
\renewcommand{\arraystretch}{1.12}
\begin{tabular*}{\linewidth}{@{\extracolsep{\fill}}lcccccccccccc@{}}
\toprule
\multirow{2}{*}{Method}
& \multicolumn{5}{c}{Cost / prompt}
& \multicolumn{7}{c}{Accuracy (\%)} \\
\cmidrule(lr){2-6}
\cmidrule(lr){7-13}
& Exec. $\downarrow$
& Call $\downarrow$
& \makecell{kTok\\in/out $\downarrow$}
& Img. $\downarrow$
& Sec. $\downarrow$
& GM $\uparrow$
& AM $\uparrow$
& Obj. $\uparrow$
& Attr. $\uparrow$
& Cnt. $\uparrow$
& Pos. $\uparrow$
& Verb $\uparrow$ \\
\midrule
FLUX.2-dev
& 1.00 & -- & -- & -- & 66.1
& 42.09 & 83.34 & 96.04 & 95.12 & 65.52 & 73.81 & 51.06 \\

FLUX.2-Klein-9B
& 1.00 & -- & -- & -- & 4.31
& 34.16 & 79.06 & 95.82 & 88.86 & 57.55 & 75.13 & 37.55 \\

\midrule
\multicolumn{13}{l}{\emph{Inference-time scaling with FLUX.2-Klein-9B}} \\

BoN+NVILA
& 32.0 & -- & -- & -- & \second{153.1}
& 41.60 & 82.38 & 97.10 & 91.10 & 61.55 & 80.39 & 44.83 \\

T2I-Copilot
& \best{1.24} & \second{4.48}
& \makecell{\second{12.61}/\\\second{2.59}}
& \second{1.24} & \best{127.3}
& 36.56 & 81.19 & 95.91 & 93.29 & 59.25 & 77.84 & 58.49 \\

RAISE
& 19.8 & 9.12 & \makecell{38.60/\\10.78} & 7.12 & 718.4
& 52.23 & 86.28 & 97.18 & 95.59 & 68.47 & 86.57 & 46.42 \\

\method{}
& \second{13.6} & \best{2.53}
& \makecell{\best{8.08}/\\\best{1.12}}
& \best{0.36} & 175.6
& 71.46 & 90.25 & 96.70 & 92.15 & 85.03 & 81.13 & 41.58 \\
\bottomrule
\end{tabular*}
\end{table}

Table~\ref{tab:geneval2_main_full_cost} expands the compact main
GenEval2 table with the full cost log for the $T=0$ setting. Unlike the
additional experiments below, all wall-clock seconds in this table were
measured under the same runtime setup, so this is the only appendix
table that reports runtime.

\subsection{Baseline Behavior in the GenEval2 Budget Sweep}
\label{appendix:geneval2_budget_behavior}

The budget sweeps in Figure~\ref{fig:pareto_plot} were conducted
independently of the main comparison in Table~\ref{tab:geneval2_main}.
Minor run-to-run variation was observed even with the same seeds
and MLLM decoding temperature ($T=0$).

Figure~\ref{fig:pareto_plot} reports realized rather than maximum
image-model executions because adaptive methods can terminate
before exhausting the budget. This distinction is important for
interpreting T2I-Copilot. In our automatic setting, its internal
quality evaluator often marks the current image as complete after
the initial generation or after only a small number of refinement
steps. As a result, increasing the maximum budget rarely increases
the realized number of image-model executions, and the method
remains in a low-cost but low-GM region. This behavior
suggests that adaptive termination is useful only when the internal
completion signal is well aligned with compositional prompt
satisfaction.

BoN+NVILA improves prompt-level accuracy by sampling and
reranking multiple candidates, but it provides no targeted feedback
about which prompt requirements remain unsatisfied. RAISE
improves alignment more substantially through multi-round
requirement-driven refinement, but uses more Qwen3-VL calls and
tokens than \method{} in our shared setting. These trends support
the use of explicit predicate states for deciding not only whether
to continue refinement, but also which refinement action to take.

\subsection{Ablation Details}
\label{appendix:additional_ablation}
\label{appendix:ablation_atomicity}
\begin{table}[t]
\centering
\caption{Detailed GenEval2 ablation results at $T=0$.
This table expands Table~\ref{tab:geneval2_ablation} with MLLM
input/output tokens, image inputs, and accuracy by atom type.
Cost is reported per prompt; \emph{Exec.} counts image-model executions,
kTok in/out reports MLLM input/output tokens in thousands, and Img.
counts MLLM image inputs.}
\label{tab:geneval2_ablation_full}
\scriptsize
\setlength{\tabcolsep}{2.4pt}
\renewcommand{\arraystretch}{1.12}
\begin{tabular*}{\linewidth}{@{\extracolsep{\fill}}lccccccccccc@{}}
\toprule
\multirow{2}{*}{Method}
& \multicolumn{4}{c}{Cost / prompt}
& \multicolumn{7}{c}{Accuracy (\%)} \\
\cmidrule(lr){2-5}
\cmidrule(lr){6-12}
& Exec. $\downarrow$
& Call $\downarrow$
& \makecell{kTok\\in/out $\downarrow$}
& Img. $\downarrow$
& GM $\uparrow$
& AM $\uparrow$
& Obj. $\uparrow$
& Attr. $\uparrow$
& Cnt. $\uparrow$
& Pos. $\uparrow$
& Verb $\uparrow$ \\
\midrule
FLUX.2-Klein-9B
& 1.00 & -- & -- & --
& 34.16 & 79.06 & 95.82 & 88.86 & 57.55 & 75.13 & 37.55 \\

\midrule
\method{}
& 13.6 & 2.53 & \makecell{8.08/\\1.12} & 0.36
& 71.46 & 90.25 & 96.70 & 92.15 & 85.03
& 81.13 & 41.58 \\

\makecell[l]{w/o modular\\verifier}
& 5.75 & 7.94 & \makecell{18.18/\\2.80} & 5.94
& 63.73 & 87.74 & 95.73 & 93.74 & 77.97 & 79.76 & 39.94 \\

Random policy
& 16.5 & 3.76 & \makecell{10.79/\\1.44} & 1.33
& 65.89 & 89.16 & 96.95 & 91.85 & 81.65 & 81.51 & 43.24 \\

w/o resampling
& 17.0 & 4.12 & \makecell{11.83/\\1.56} & 1.82
& 63.98 & 87.35 & 96.04 & 90.27 & 79.27 & 77.35 & 36.21 \\

w/o editing
& 18.9 & 2.23 & \makecell{7.31/\\1.09} & 0.00
& 64.03 & 89.15 & 96.81 & 90.44 & 82.52 & 81.04 & 39.04 \\

w/o rewrites
& 17.8 & 1.47 & \makecell{6.71/\\0.50} & 0.23
& 59.71 & 86.66 & 96.08 & 89.64 & 78.39 & 77.91 & 38.11 \\
\bottomrule
\end{tabular*}
\end{table}

\begin{table}[t]
\centering
\caption{GenEval2 prompt-level accuracy (\%) by atomicity at $T=0$.
All variants follow the shared setup in Sec.~\ref{sec:exp_setup}
except for the ablated component.
The \emph{w/o modular verifier} variant keeps visual-program parsing
but obtains all predicate states from an MLLM given the current image
and visual program.}
\label{tab:geneval2_ablation_atomicity}
\small
\setlength{\tabcolsep}{5pt}
\begin{tabularx}{\linewidth}{lYYYYYYYY}
\toprule
Method & 3 & 4 & 5 & 6 & 7 & 8 & 9 & 10 \\
\midrule
FLUX.2-Klein-9B
& 63.17 & 55.11 & 38.78 & 34.27
& 23.22 & 21.66 & 19.90 & 17.19 \\
\midrule

\method{}
& 89.24 & 89.69 & 78.59 & 74.60
& 71.19 & 62.04 & 57.96 & 48.39 \\

\makecell[l]{w/o modular\\verifier}
& 85.02 & 78.75 & 71.26 & 70.57
& 57.14 & 53.86 & 53.25 & 40.01 \\

Random policy
& 85.76 & 83.81 & 81.45 & 76.24
& 60.00 & 53.82 & 49.85 & 36.20 \\

w/o resampling
& 84.02 & 82.23 & 71.52 & 69.86
& 58.42 & 54.16 & 51.27 & 40.40 \\

w/o editing
& 89.64 & 80.46 & 78.68 & 66.74
& 55.43 & 55.01 & 48.79 & 37.47 \\

w/o rewrites
& 76.44 & 79.26 & 69.47 & 58.55
& 52.52 & 55.85 & 49.69 & 35.88 \\
\bottomrule
\end{tabularx}
\end{table}

Table~\ref{tab:geneval2_ablation_full} reports the full ablation
breakdown omitted from the compact main-text table, including
MLLM image inputs, token usage, and atom-type scores.
Table~\ref{tab:geneval2_ablation_atomicity} reports the same
ablations by prompt atomicity.

The \emph{w/o modular verifier} variant keeps the visual-program
parser fixed, but replaces the modular predicate verifier with one
MLLM verifier call per verification step. The MLLM receives the
current image and visual program, scores all predicates, and returns
the same discrete predicate states used by the action selector. This
variant reduces prompt-level accuracy and substantially increases
token usage, indicating that the visual-program structure alone is
not sufficient: modular perception evidence provides a more
effective predicate-state signal with lower token cost than asking an MLLM
to judge all predicates from the full image at each refinement round.

The policy and action ablations further show that predicate states
must be paired with an appropriate controller. The \emph{Random policy}
variant selects both the blocking target and next
action type at random, reducing GM while increasing image-model executions.
Removing resampling forces all post-initial refinement to use
editing, which is less reliable for global layout, count, or object-set
failures. Removing editing disables localized correction, which is
useful for attributes and localized count or exclusion failures.
These results show that the
controller should route different predicate failures to different
computation types rather than simply continue refinement.

Finally, removing prompt rewrites causes the largest ablation drop
despite increasing realized image-model executions. This suggests
that repeatedly sampling from a single prompt wording gives less
useful exploration than drawing candidates from a pool of
prompt rewrites while keeping the original visual
program fixed as the acceptance criterion. The atomicity-wise
breakdown shows that the full controller has the highest GM at
atomicities $4$ and $7$--$10$, with its clearest advantages on
the more complex prompts.

\subsection{Stochastic MLLM Decoding}
\label{appendix:geneval2_t015}
\begin{table}[t]
\centering
\caption{GenEval2 results at MLLM decoding temperature $T=0.15$.
$\Delta P$ is the gain in prompt-level accuracy over single-pass
FLUX.2-Klein-9B, in percentage points.
We report per-prompt inference cost and accuracy gain per unit of cost.
\emph{Exec.} counts image-model executions, and
\emph{kTok} is the Qwen3-VL-32B input/output token total in thousands.
Best and second-best values among inference-time methods are shown in
\textbf{bold} and \underline{underlined}, respectively; missing cost
entries are not ranked.}
\label{tab:geneval2_t0.15}
\scriptsize
\setlength{\tabcolsep}{2.9pt}
\renewcommand{\arraystretch}{1.10}
\begin{tabular*}{\linewidth}{@{\extracolsep{\fill}}lccccccccc@{}}
\toprule
\multirow{2}{*}{Method}
& \multicolumn{3}{c}{Accuracy (\%)}
& \multicolumn{3}{c}{Cost}
& \multicolumn{3}{c}{Gain / cost} \\
\cmidrule(lr){2-4}
\cmidrule(lr){5-7}
\cmidrule(lr){8-10}
& GM $\uparrow$
& $\Delta P$ $\uparrow$
& AM $\uparrow$
& Exec. $\downarrow$
& Call $\downarrow$
& kTok $\downarrow$
& $\Delta P$/Exec. $\uparrow$
& $\Delta P$/Call $\uparrow$
& $\Delta P$/kTok $\uparrow$ \\
\midrule
FLUX.2-Klein-9B
& 34.16 & -- & 79.06
& 1.0 & -- & -- & -- & -- & -- \\

\midrule
\multicolumn{10}{l}{\emph{Inference-time scaling with FLUX.2-Klein-9B}} \\

BoN+NVILA
& 41.60 & +7.44 & 82.38
& 32.0 & -- & -- & 0.23 & -- & -- \\

T2I-Copilot
& 35.51 & +1.35 & 80.57
& \best{1.22} & \second{4.43} & \second{15.0}
& \second{1.11} & 0.30 & 0.09 \\

RAISE
& \second{52.81} & \second{+18.65} & \second{86.42}
& 19.9 & 9.13 & 49.9
& 0.94 & \second{2.04} & \second{0.37} \\

\method{}
& \best{68.79} & \best{+34.63} & \best{89.38}
& \second{13.9} & \best{2.48} & \best{9.1}
& \best{2.49} & \best{13.96} & \best{3.79} \\
\bottomrule
\end{tabular*}
\end{table}

\begin{table}[t]
\centering
\caption{GenEval2 prompt-level accuracy (\%) by atomicity at MLLM decoding temperature
$T=0.15$.
FLUX.2-dev~\citep{flux2dev} is included as a larger single-pass generator reference.
Best and second-best scores are shown in \textbf{bold} and
\underline{underlined}, respectively.}
\label{tab:geneval2_atomicity_t0.15}
\small
\setlength{\tabcolsep}{5pt}
\begin{tabularx}{\linewidth}{lYYYYYYYY}
\toprule
Method & 3 & 4 & 5 & 6 & 7 & 8 & 9 & 10 \\
\midrule
FLUX.2-dev
& 73.24 & 65.58 & 48.95 & 47.56
& 31.58 & 30.70 & 21.41 & 17.74 \\

FLUX.2-Klein-9B
& 63.17 & 55.11 & 38.78 & 34.27 & 23.22 & 21.66 & 19.90 & 17.19 \\

\midrule
\multicolumn{9}{l}{\emph{Inference-time scaling with FLUX.2-Klein-9B}} \\

BoN+NVILA
& 75.72 & 64.70 & 52.05 & 44.12 & 28.06 & 28.03 & 23.64 & 16.49 \\

T2I-Copilot
& 57.97 & 56.69 & 37.19 & 36.01 & 31.91 & 26.01 & 19.87 & 18.41 \\

RAISE
& \best{90.59} & \second{75.67} & \second{56.02} & \second{60.79}
& \second{34.43} & \second{43.74} & \second{31.59} & \second{29.68} \\

\method{}
& \second{83.54} & \best{84.12} & \best{83.40} & \best{72.80}
& \best{63.70} & \best{61.34} & \best{55.77} & \best{45.69} \\
\bottomrule
\end{tabularx}
\end{table}

Tables~\ref{tab:geneval2_t0.15} and
\ref{tab:geneval2_atomicity_t0.15} report additional GenEval2 results
with stochastic MLLM decoding at $T=0.15$. \method{} remains the best
inference-time method, reaching a prompt-level accuracy of $68.79$ and a
$+34.63$ point improvement over single-pass FLUX.2-Klein-9B. It also
retains the highest GM gain per unit of cost, using fewer
image-model executions, MLLM calls, and tokens than RAISE. By atomicity,
\method{} is second-best at atomicity $3$ and best from atomicity $4$ to
$10$, again showing that predicate-state refinement is most useful when
many prompt requirements must be jointly satisfied.

\subsection{FLUX.2-Klein-4B Generator}
\label{appendix:flux2_klein_4b}
\begin{table}[t]
\centering
\caption{GenEval2 results with FLUX.2-Klein-4B at $T=0$.
FLUX.2-Klein-4B is used as both generator and editor, and
Qwen3-VL-32B is used as the MLLM.
This table also includes the single-pass FLUX.2-Klein-9B result as a
reference.
Cost is reported per prompt; \emph{Exec.} counts image-model executions,
kTok in/out reports MLLM input/output tokens in thousands, and Img.
counts MLLM image inputs.
Best and second-best values among FLUX.2-Klein-4B-based inference-time
methods are shown in \textbf{bold} and \underline{underlined},
respectively; for cost columns, lower is better and missing entries are
not ranked.}
\label{tab:geneval2_klein4b}
\scriptsize
\setlength{\tabcolsep}{2.4pt}
\renewcommand{\arraystretch}{1.12}
\begin{tabular*}{\linewidth}{@{\extracolsep{\fill}}lccccccccccc@{}}
\toprule
\multirow{2}{*}{Method}
& \multicolumn{4}{c}{Cost / prompt}
& \multicolumn{7}{c}{Accuracy (\%)} \\
\cmidrule(lr){2-5}
\cmidrule(lr){6-12}
& Exec. $\downarrow$
& Call $\downarrow$
& \makecell{kTok\\in/out $\downarrow$}
& Img. $\downarrow$
& GM $\uparrow$
& AM $\uparrow$
& Obj. $\uparrow$
& Attr. $\uparrow$
& Cnt. $\uparrow$
& Pos. $\uparrow$
& Verb $\uparrow$ \\
\midrule
FLUX.2-Klein-9B
& 1.00 & -- & -- & --
& 34.16 & 79.06 & 95.82 & 88.86 & 57.55 & 75.13 & 37.55 \\

FLUX.2-Klein-4B
& 1.00 & -- & -- & --
& 25.11 & 73.07 & 90.92 & 81.73 & 51.91 & 68.22 & 28.01 \\

\midrule
\multicolumn{12}{l}{\emph{Inference-time scaling with FLUX.2-Klein-4B}} \\

BoN+NVILA
& 32.0 & -- & -- & --
& 32.75 & 77.09 & 93.25 & 84.93 & 55.46 & 75.15 & 33.59 \\

T2I-Copilot
& \best{1.19} & \second{4.37}
& \makecell{\second{12.16}/\\\second{2.54}} & \second{1.19}
& 30.38 & 77.94 & 93.76 & \second{90.05} & 54.62
& \second{78.23} & \best{49.03} \\

RAISE
& 22.3 & 10.3 & \makecell{44.41/\\12.30} & 8.33
& \second{46.32} & \second{83.49} & \best{95.20}
& \best{92.53} & \second{64.98} & \best{83.02}
& \second{46.01} \\

\method{}
& \second{17.0} & \best{2.75}
& \makecell{\best{9.06}/\\\best{1.24}} & \best{0.79}
& \best{58.84} & \best{85.52} & \second{93.83}
& 86.61 & \best{77.82} & 78.20 & 38.63 \\
\bottomrule
\end{tabular*}
\end{table}

\begin{table}[t]
\centering
\caption{GenEval2 prompt-level accuracy (\%) by atomicity with FLUX.2-Klein-4B at
$T=0$.
FLUX.2-Klein-4B is used as both generator and editor, and
Qwen3-VL-32B is used as the MLLM.
The single-pass FLUX.2-Klein-9B result is included as a reference.
Best and second-best scores among FLUX.2-Klein-4B-based inference-time
methods are shown in \textbf{bold} and \underline{underlined},
respectively.}
\label{tab:geneval2_klein4b_atomicity}
\small
\setlength{\tabcolsep}{5pt}
\begin{tabularx}{\linewidth}{lYYYYYYYY}
\toprule
Method & 3 & 4 & 5 & 6 & 7 & 8 & 9 & 10 \\
\midrule
FLUX.2-Klein-9B
& 63.17 & 55.11 & 38.78 & 34.27
& 23.22 & 21.66 & 19.90 & 17.19 \\

FLUX.2-Klein-4B
& 44.21 & 46.09 & 31.63 & 27.03
& 16.02 & 15.63 & 11.52 & 8.80 \\

\midrule
\multicolumn{9}{l}{\emph{Inference-time scaling with FLUX.2-Klein-4B}} \\

BoN+NVILA
& 62.85 & 57.33 & 39.46 & 36.92
& 22.32 & 17.72 & 18.27 & 7.10 \\

T2I-Copilot
& 54.99 & 56.65 & 33.28 & 34.76
& 22.93 & 16.88 & 14.13 & 9.43 \\

RAISE
& \best{80.69} & \second{72.22} & \second{49.69}
& \second{50.14} & \second{33.22} & \second{32.53}
& \second{28.74} & \second{23.32} \\

\method{}
& \second{78.29} & \best{80.53} & \best{65.67}
& \best{66.91} & \best{51.56} & \best{51.40}
& \best{39.47} & \best{36.92} \\
\bottomrule
\end{tabularx}
\end{table}

Tables~\ref{tab:geneval2_klein4b} and
\ref{tab:geneval2_klein4b_atomicity} report additional GenEval2
results using FLUX.2-Klein-4B~\citep{flux2klein4b} as both generator and editor with
Qwen3-VL-32B at $T=0$. The smaller single-pass generator starts below
FLUX.2-Klein-9B, but \method{} raises prompt-level accuracy from
$25.11$ to $58.84$ and AM score from $73.07$ to
$85.52$. Compared with RAISE, \method{} improves prompt-level accuracy by $12.52$ points while using fewer image-model executions,
MLLM calls, tokens, and image inputs. By atomicity, \method{} is
best from atomicity $4$ to $10$ and second-best at atomicity $3$.

\section{Additional Benchmarks and Audits}
\label{appendix:additional_benchmarks}

\subsection{DrawBench}
\label{appendix:drawbench}
\begin{table}[t]
\centering
\caption{
DrawBench results at $T=0$.
All methods follow the shared setup in Sec.~\ref{sec:exp_setup}.
Cost is reported per prompt, where Exec., Call, kTok, and Img.
denote image-model executions, MLLM calls, MLLM input/output tokens in
thousands, and MLLM image inputs, respectively.
For cost columns, lower is better; for alignment and preference metrics,
higher is better. Best and second-best values among non-single-pass
methods are shown in \textbf{bold} and \underline{underlined},
respectively; missing cost entries are not ranked.
}
\label{tab:drawbench_main}
\scriptsize
\setlength{\tabcolsep}{3.0pt}
\renewcommand{\arraystretch}{1.12}
\begin{tabular*}{\linewidth}{@{\extracolsep{\fill}}lccccccc@{}}
\toprule
\multirow{2}{*}{Method}
& \multicolumn{4}{c}{Cost}
& \multicolumn{3}{c}{Alignment and preference metrics} \\
\cmidrule(lr){2-5}
\cmidrule(lr){6-8}
& Exec. $\downarrow$
& Call $\downarrow$
& \makecell{kTok\\in/out $\downarrow$}
& Img. $\downarrow$
& VQAScore $\uparrow$
& ImageReward $\uparrow$
& HPS-v3 $\uparrow$ \\
\midrule
FLUX.2-Klein-9B
& 1.00 & -- & -- & --
& 0.856 & 1.206 & 11.01 \\

BoN+NVILA
& 32.0 & -- & -- & --
& 0.887 & \best{1.271} & \best{11.12} \\

T2I-Copilot
& \best{1.02} & \second{4.04}
& \makecell{\second{10.19}/\\\second{2.21}} & \second{1.02}
& 0.855 & \second{1.196} & 8.14 \\

RAISE
& 14.4 & 6.49 & \makecell{26.16/\\7.95} & 4.47
& \best{0.898} & 1.194 & \second{11.01} \\

\method{}
& \second{7.07} & \best{3.57}
& \makecell{\best{8.35}/\\\best{1.10}} & \best{0.37}
& \second{0.896} & 1.038 & 10.61 \\
\bottomrule
\end{tabular*}
\end{table}

Table~\ref{tab:drawbench_main} reports results on
DrawBench~\citep{saharia2022drawbench}, using VQAScore~\citep{li2024vqascore},
ImageReward~\citep{xu2023imagereward}, and HPS-v3~\citep{ma2025hpsv3}.
On DrawBench, \method{} improves VQAScore from $0.856$ for single-pass
FLUX.2-Klein-9B to $0.896$, close to the best compared inference-time
method, RAISE, at $0.898$, while using fewer image-model executions,
MLLM calls, tokens, and image inputs. This
suggests that predicate-driven refinement can improve broad semantic
prompt-image alignment beyond GenEval2. However, ImageReward and HPS-v3
do not improve: BoN+NVILA achieves the highest scores on both
preference-oriented metrics, and \method{} remains below the single-pass
generator. This pattern is consistent with the design of \method{},
which targets semantic controllability and prompt requirement
satisfaction rather than generic preference-reward optimization.
The independent quality and human analyses in
Appendices~\ref{appendix:quality_analysis} and~\ref{appendix:human_evaluation}
qualify this interpretation: some Q-Align deficits are significant,
although overall human comparisons are not.

\begin{figure}[p]
\centering
\scriptsize
\setlength{\tabcolsep}{1.2pt}
\renewcommand{\arraystretch}{1.04}
\newcommand{\qmoreimg}[1]{\includegraphics[width=0.185\linewidth]{#1}}
\newcommand{\qmoreprompt}[1]{\multicolumn{5}{p{0.96\linewidth}}{\centering \textit{Prompt:} #1} \\[0.20em]}
\begin{tabular}{ccccc}
\textbf{FLUX.2-Klein-9B}
& \textbf{BoN+NVILA}
& \textbf{T2I-Copilot}
& \textbf{RAISE}
& \textbf{\method{} (ours)} \\

\qmoreimg{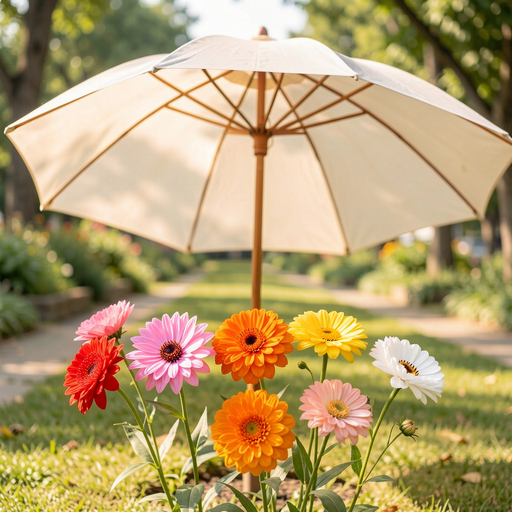}
& \qmoreimg{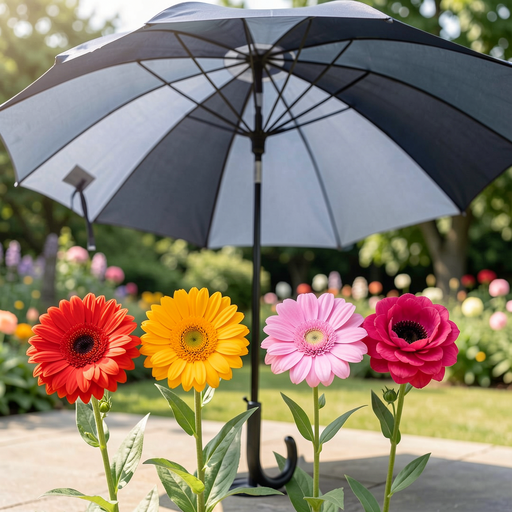}
& \qmoreimg{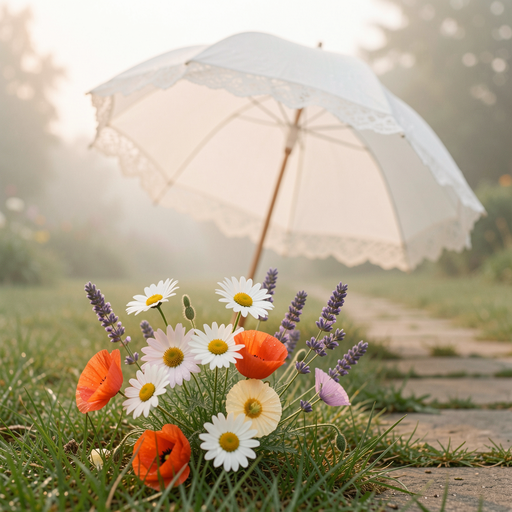}
& \qmoreimg{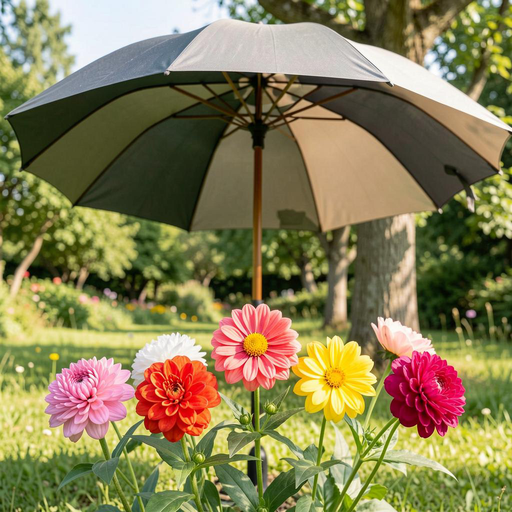}
& \qmoreimg{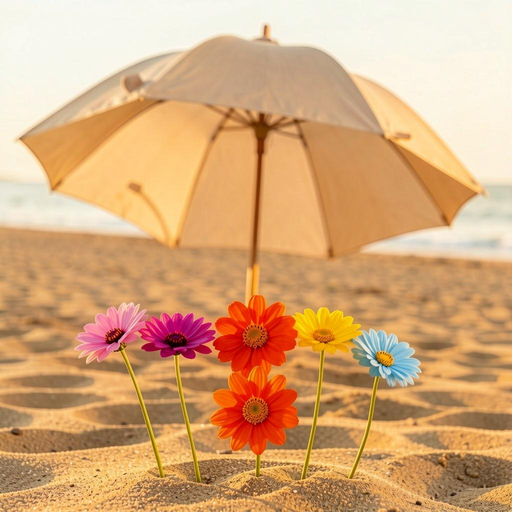} \\
\qmoreprompt{six flowers in front of a umbrella}

\qmoreimg{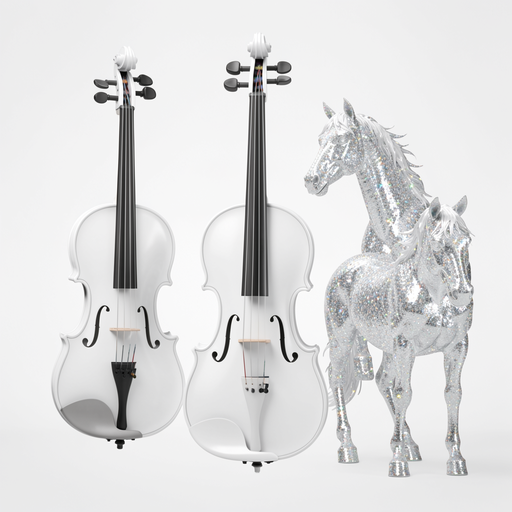}
& \qmoreimg{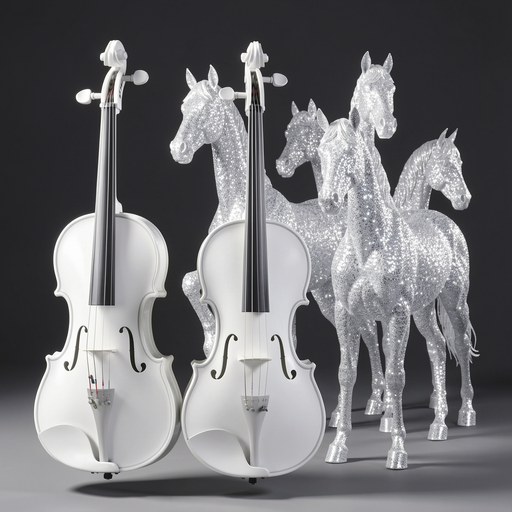}
& \qmoreimg{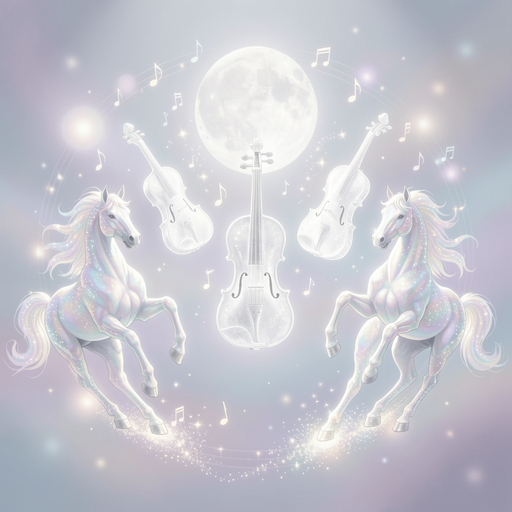}
& \qmoreimg{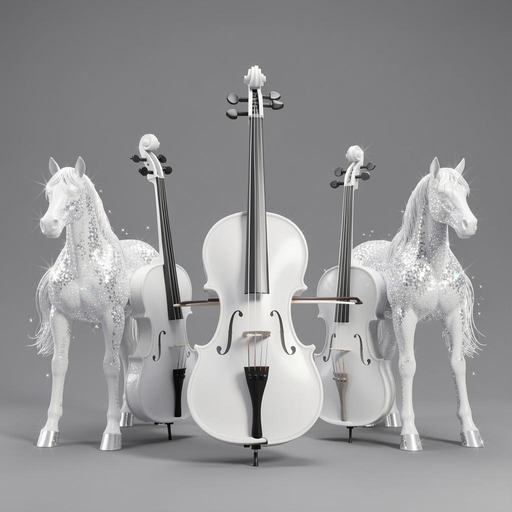}
& \qmoreimg{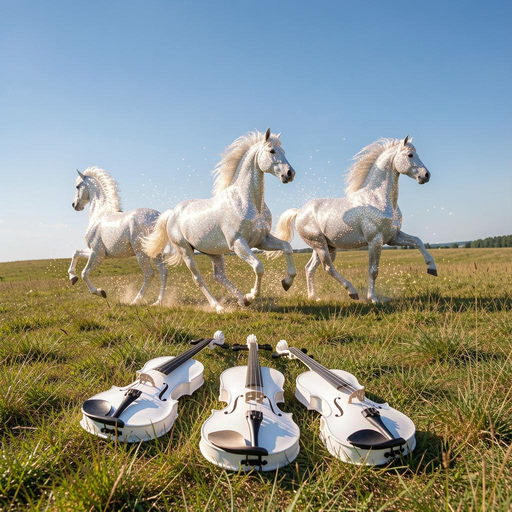} \\
\qmoreprompt{three white violins and three sparkling horses}

\qmoreimg{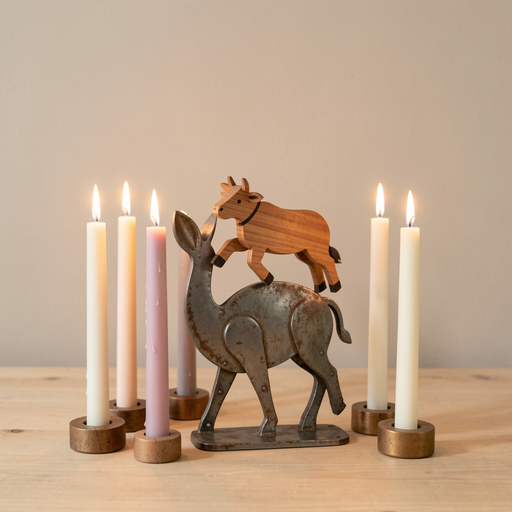}
& \qmoreimg{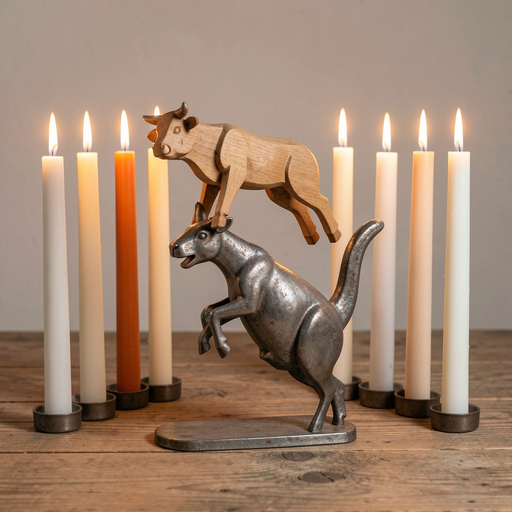}
& \qmoreimg{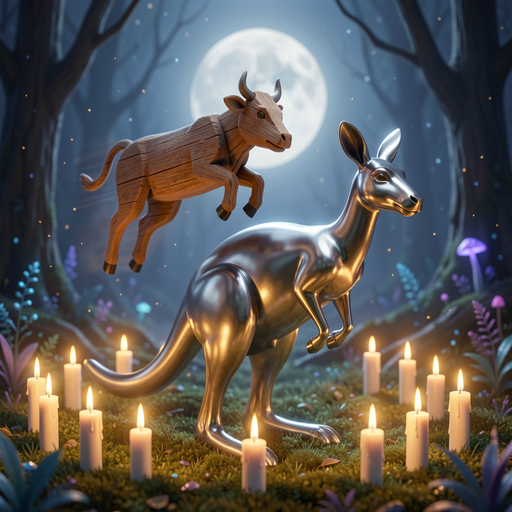}
& \qmoreimg{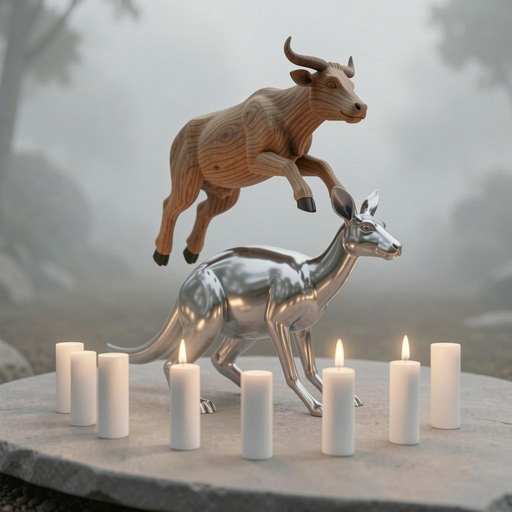}
& \qmoreimg{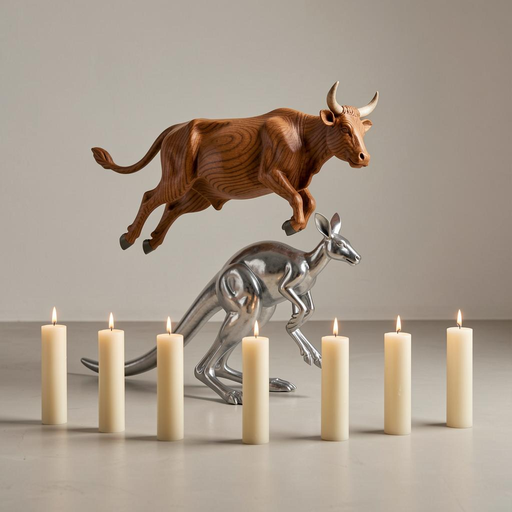} \\
\qmoreprompt{seven candles, and a wooden cow jumping over a metal kangaroo}

\qmoreimg{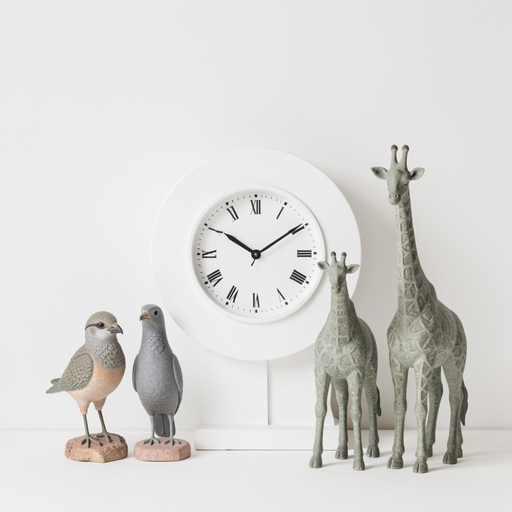}
& \qmoreimg{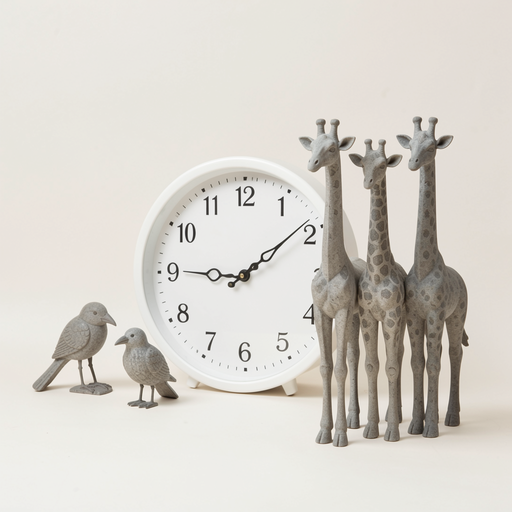}
& \qmoreimg{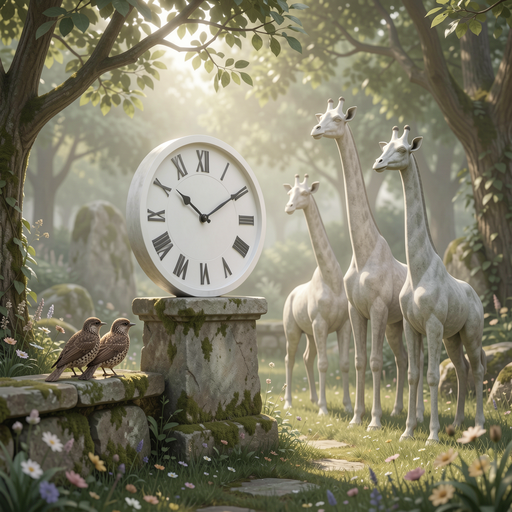}
& \qmoreimg{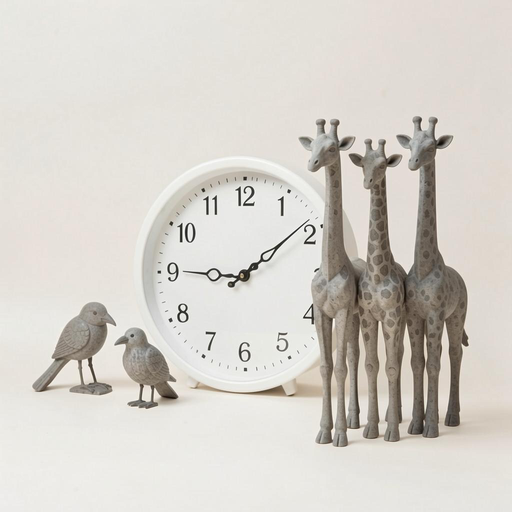}
& \qmoreimg{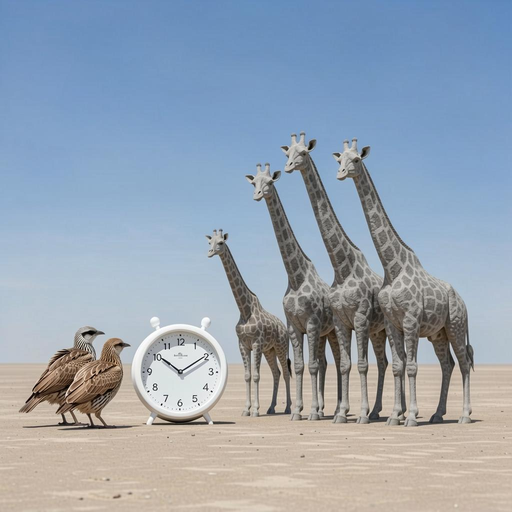} \\
\qmoreprompt{two birds to the left of a white clock, and four stone giraffes}

\qmoreimg{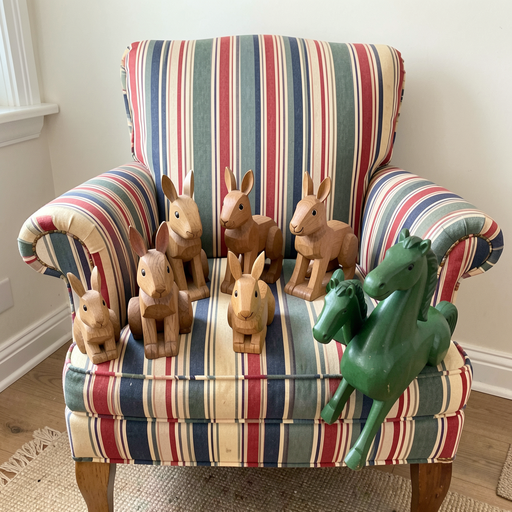}
& \qmoreimg{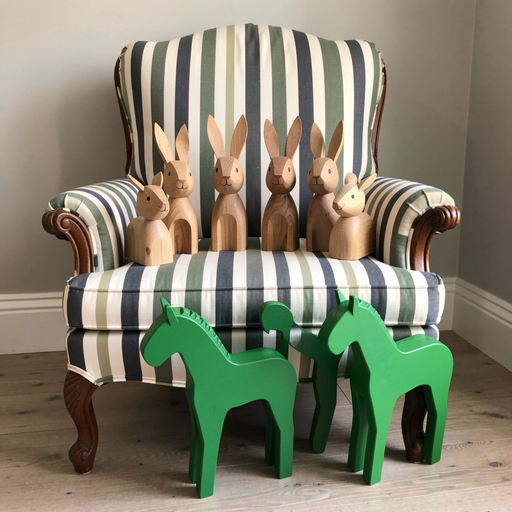}
& \qmoreimg{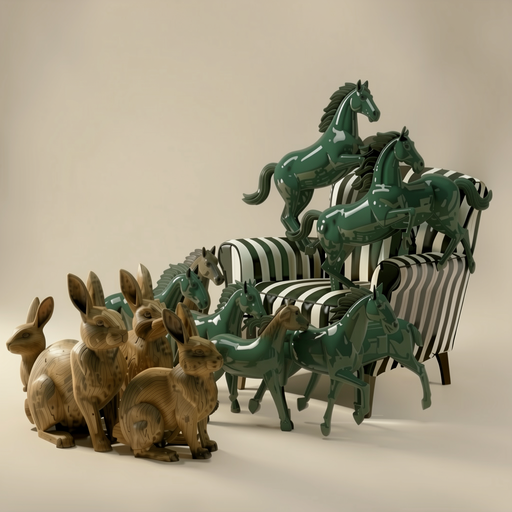}
& \qmoreimg{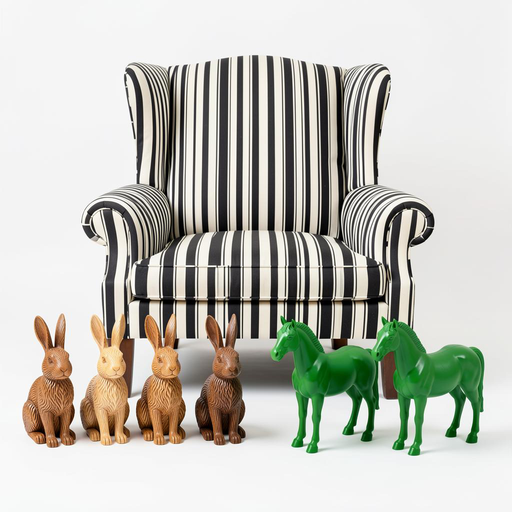}
& \qmoreimg{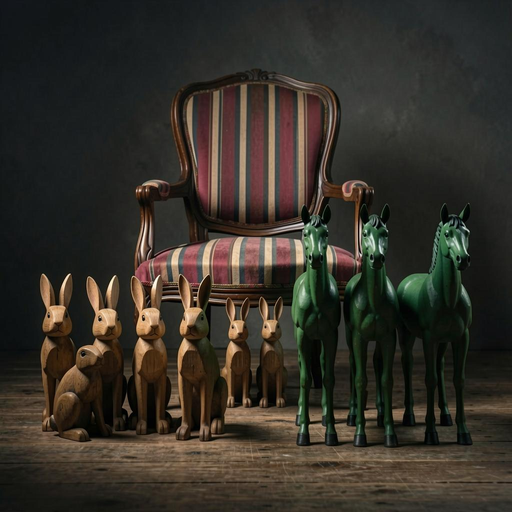} \\
\qmoreprompt{seven wooden rabbits, and three green horses in front of a striped chair}

\qmoreimg{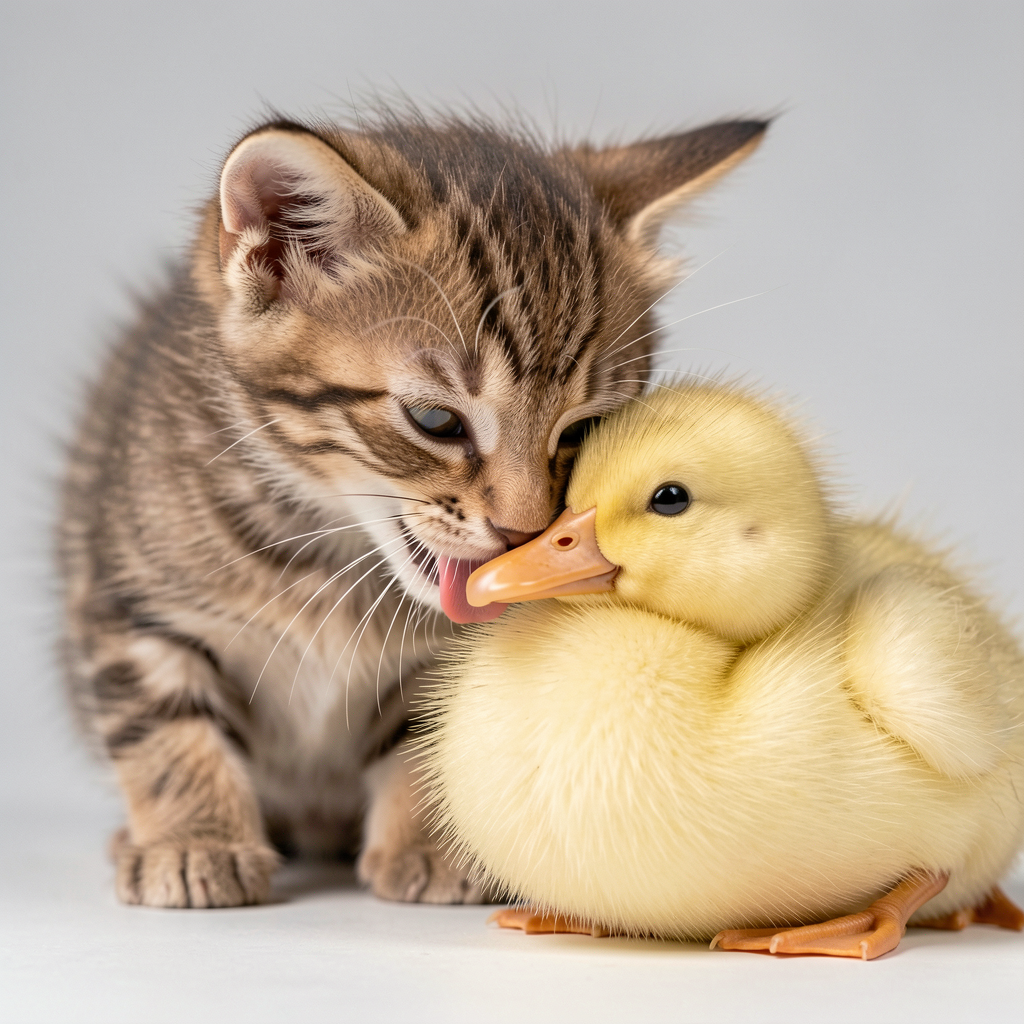}
& \qmoreimg{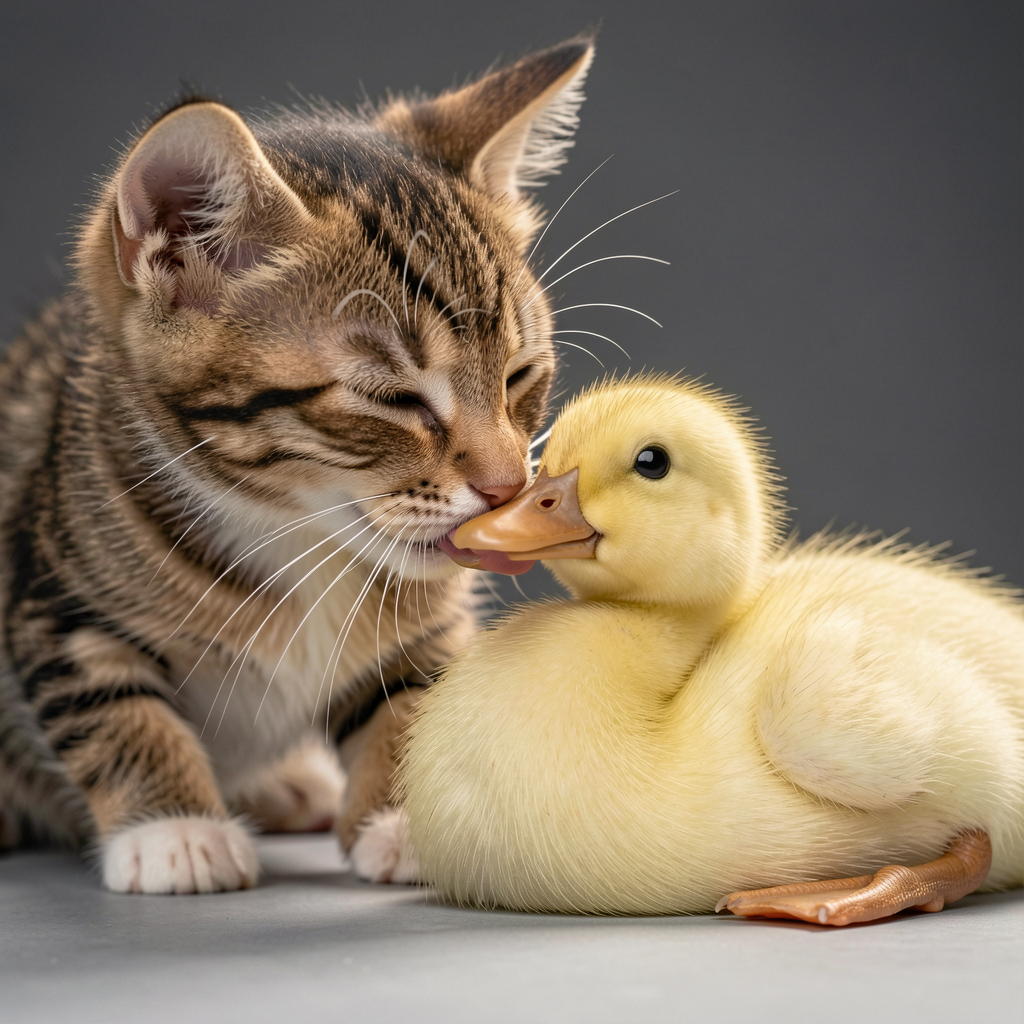}
& \qmoreimg{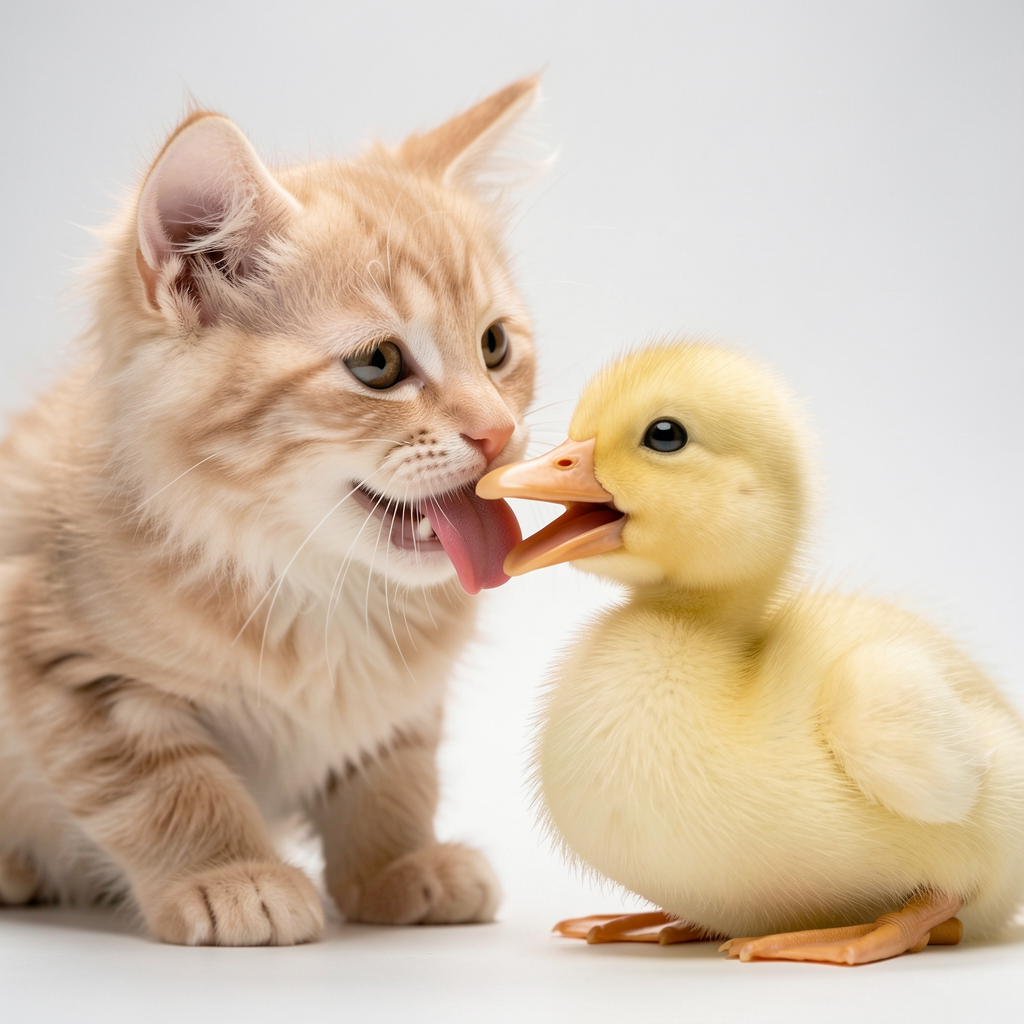}
& \qmoreimg{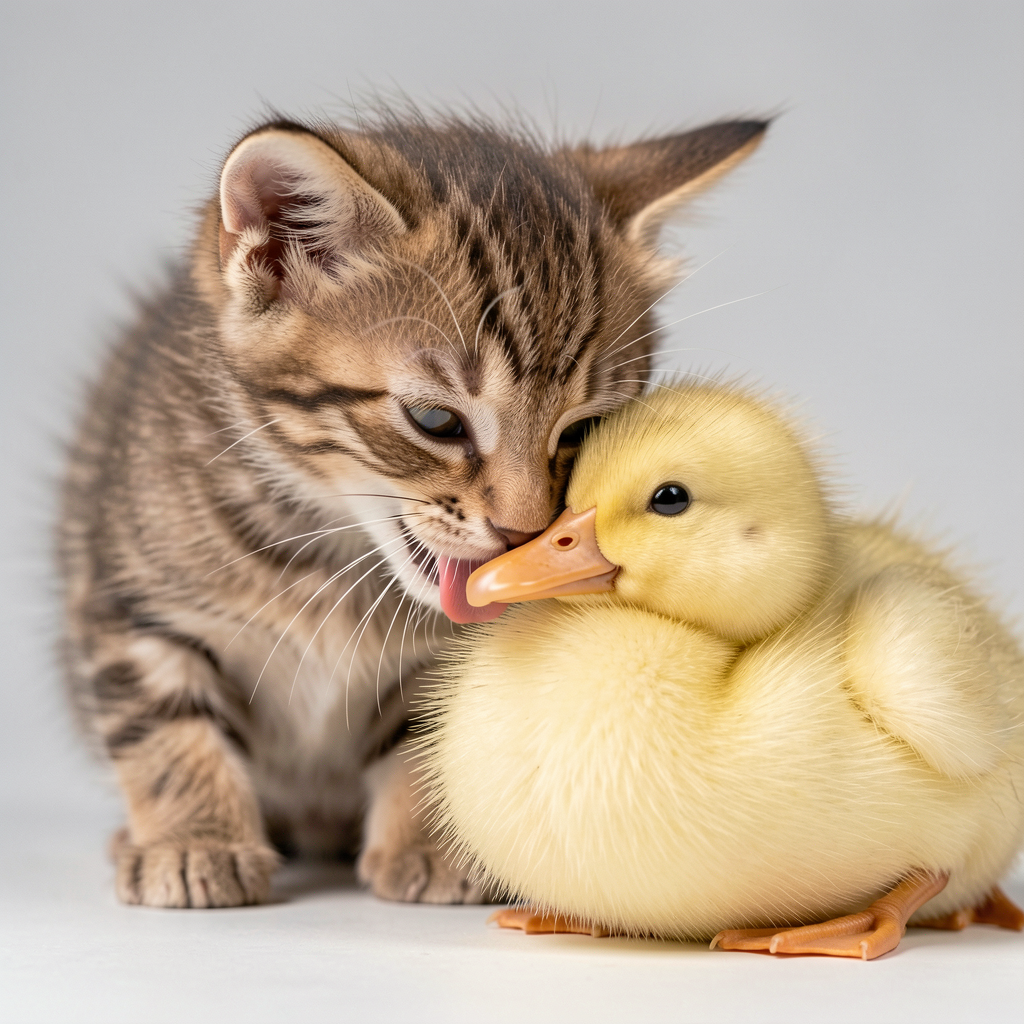}
& \qmoreimg{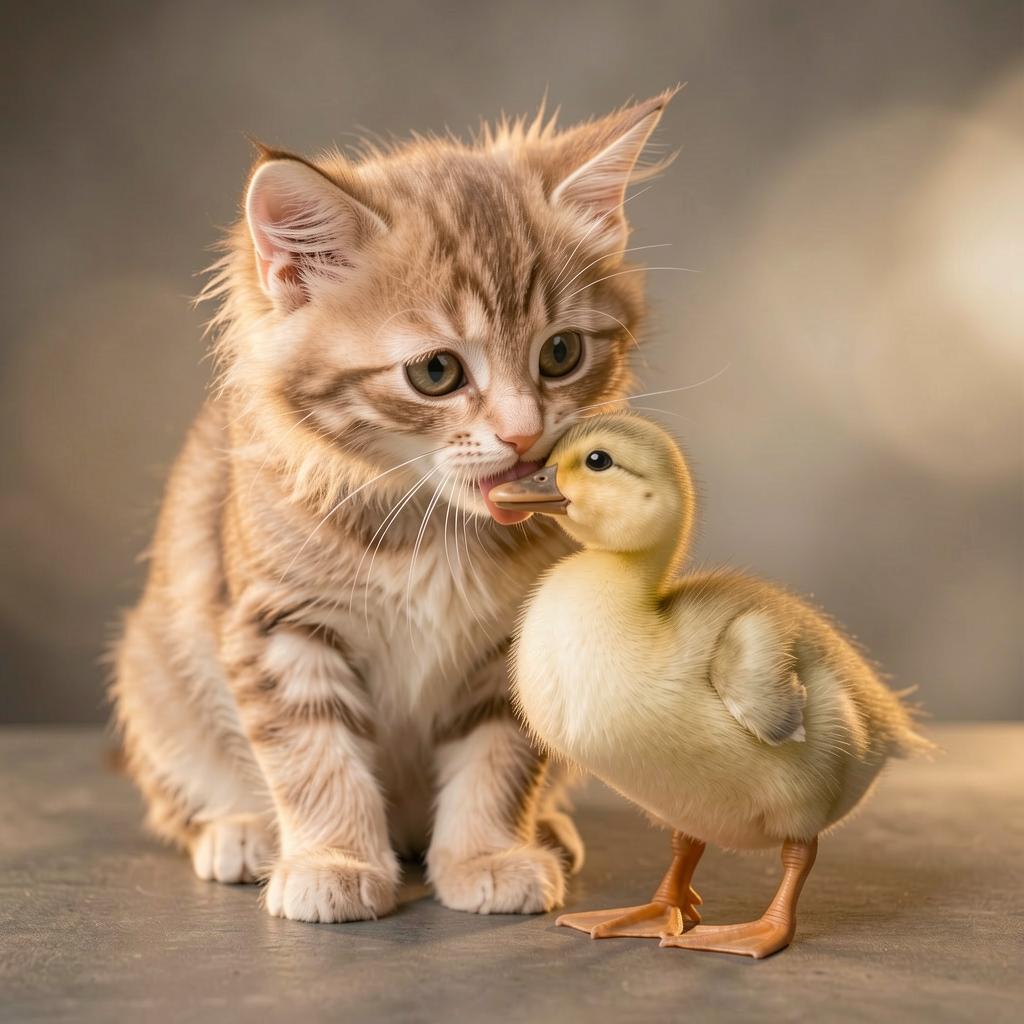} \\
\qmoreprompt{35mm macro shot a kitten licking a baby duck, studio lighting}

\qmoreimg{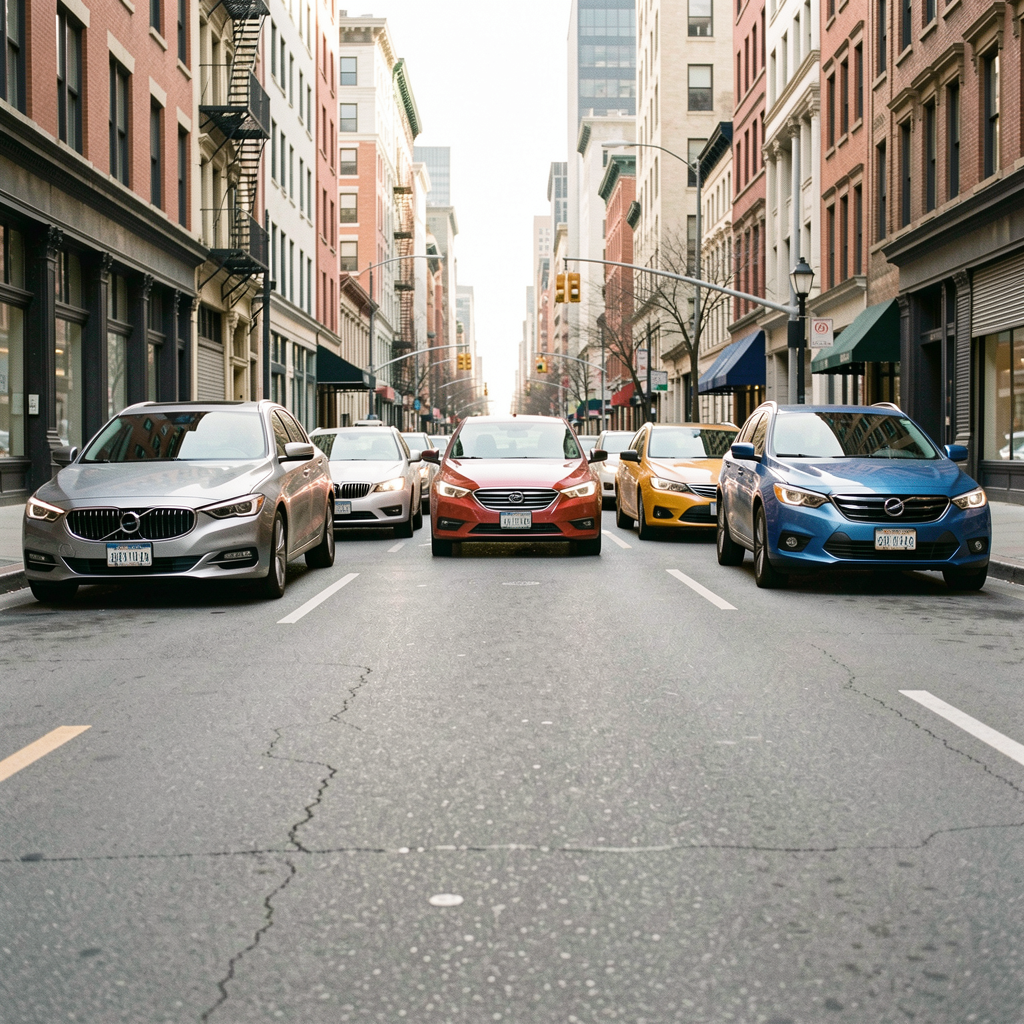}
& \qmoreimg{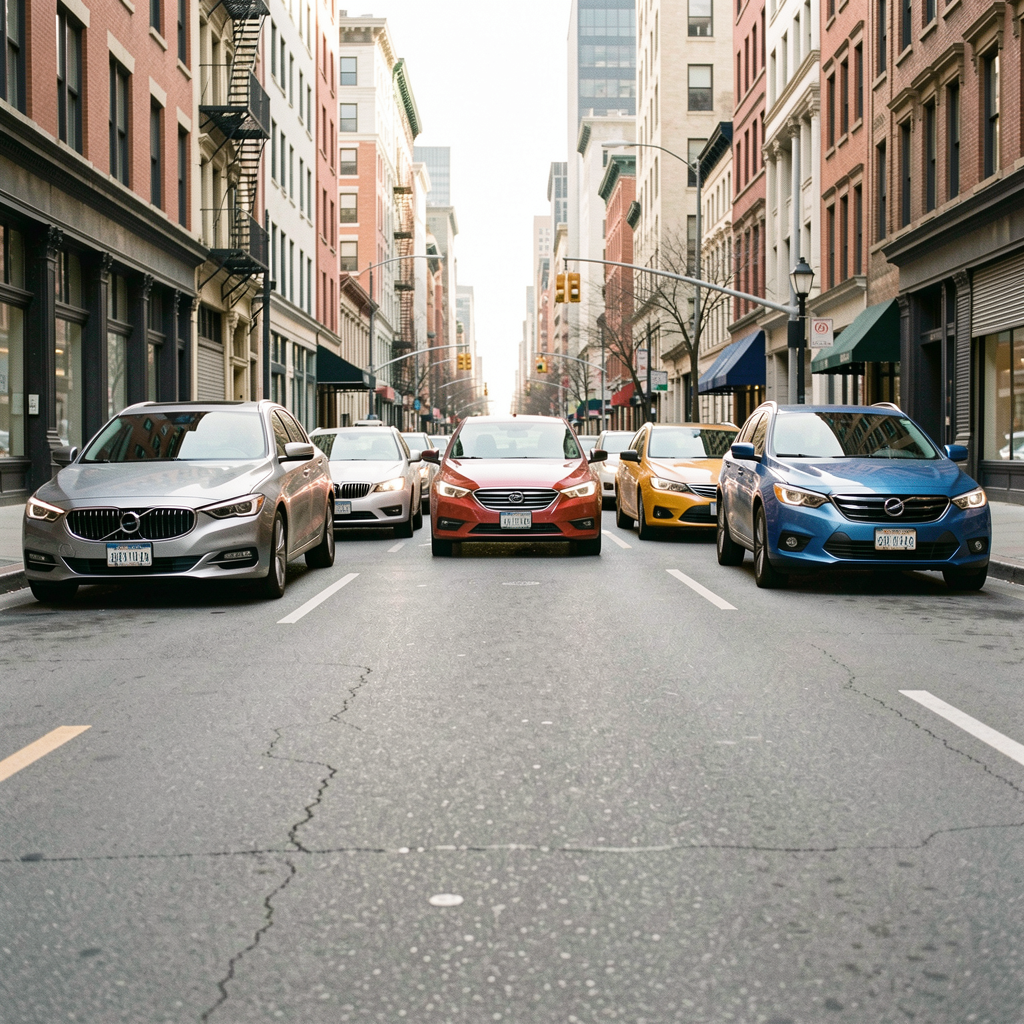}
& \qmoreimg{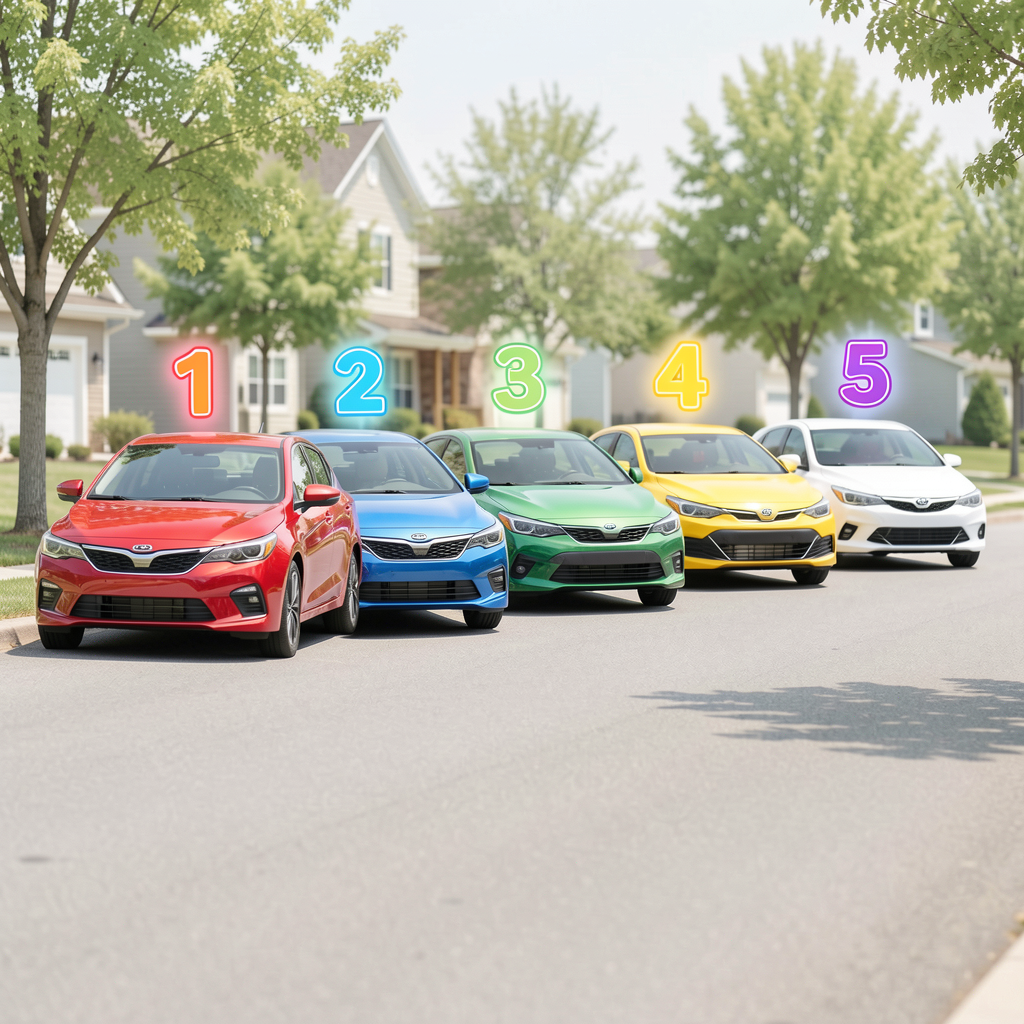}
& \qmoreimg{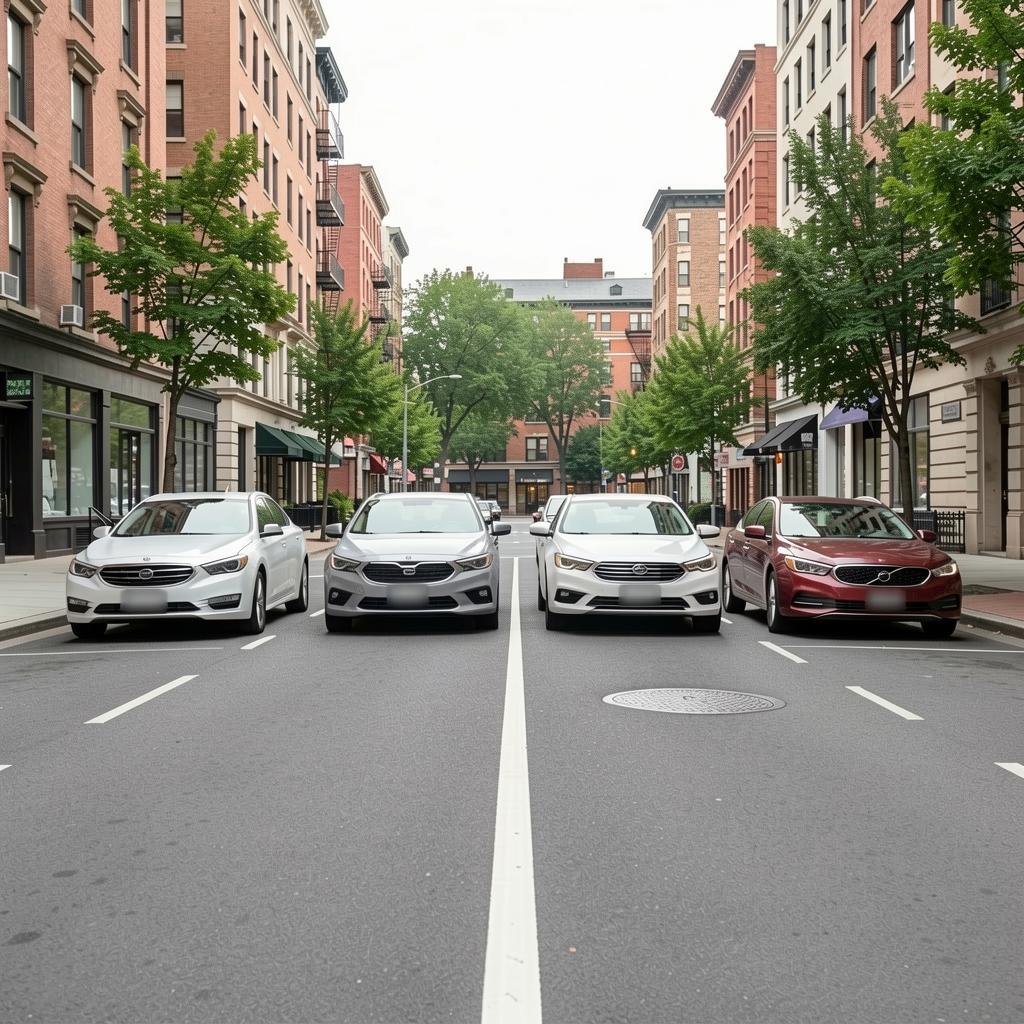}
& \qmoreimg{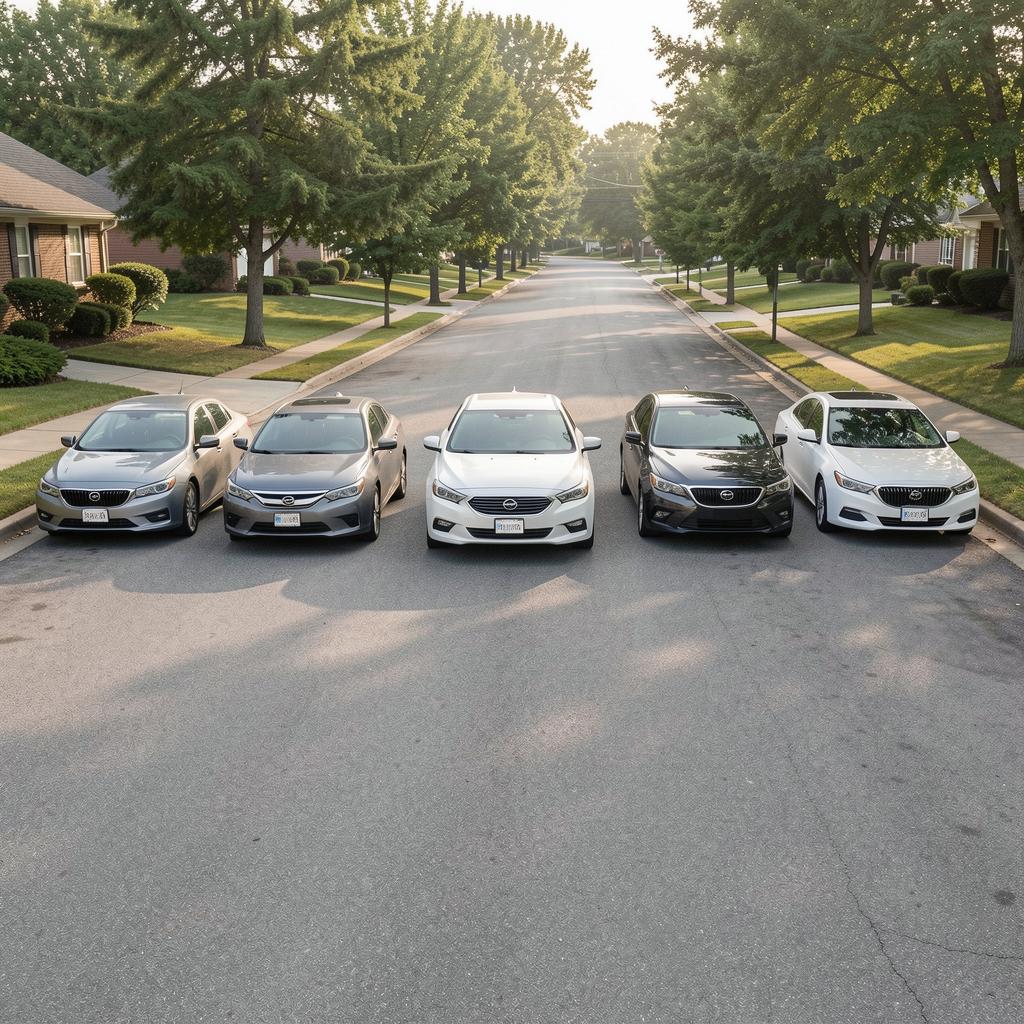} \\
\qmoreprompt{Five cars on the street}
\end{tabular}
\caption{
Additional qualitative examples across GenEval2 and DrawBench prompts.
The first five rows are GenEval2 compositional prompts, and the last two
rows are DrawBench prompts. Each row uses the same prompt across five
generation or refinement methods.
}
\label{fig:qualitative_more_examples}
\end{figure}

Figure~\ref{fig:qualitative_more_examples} provides additional
qualitative examples, including five GenEval2 compositional prompts and
two DrawBench prompts.

\subsection{Original GenEval}
\label{appendix:geneval}
\begin{table}[t]
\centering
\caption{
Original GenEval (GenEval1) accuracy (\%) at $T=0$.
Cost is reported per prompt under the shared accounting in
Sec.~\ref{sec:exp_setup}. Lower is better for cost columns, and higher
is better for accuracy. Best and second-best values among
non-single-pass methods are shown in \textbf{bold} and
\underline{underlined}, respectively; missing cost entries are not
ranked.
}
\label{tab:geneval_main}
\scriptsize
\setlength{\tabcolsep}{2.2pt}
\renewcommand{\arraystretch}{1.12}
\begin{tabular*}{\linewidth}{@{\extracolsep{\fill}}lccccccccccc@{}}
\toprule
\multirow{2}{*}{Method}
& \multicolumn{4}{c}{Efficiency}
& \multicolumn{7}{c}{Accuracy (\%)} \\
\cmidrule(lr){2-5} \cmidrule(lr){6-12}
& Exec. $\downarrow$
& Call $\downarrow$
& \makecell{kTok\\in/out $\downarrow$}
& Img. $\downarrow$
& Overall $\uparrow$
& Single $\uparrow$
& Two $\uparrow$
& Cnt. $\uparrow$
& Color $\uparrow$
& Pos. $\uparrow$
& Attr. $\uparrow$ \\
\midrule
FLUX.2-Klein-9B
& 1.00 & -- & -- & --
& 85.0 & 100.0 & 96.0 & 80.0 & 91.0 & 72.0 & 71.0 \\

BoN+NVILA
& 32.0 & -- & -- & --
& \best{96.4} & \best{100.0} & \best{100.0}
& \best{98.1} & \best{97.3} & \second{92.0} & \best{91.0} \\

T2I-Copilot
& \best{1.04} & \second{4.08}
& \makecell{\second{10.00}/\\\second{2.09}} & \second{1.04}
& 86.0 & 99.4 & 95.2 & 70.0 & 90.4 & 87.0 & 74.0 \\

RAISE
& 11.2 & 4.87 & \makecell{18.56/\\5.51} & 2.87
& \second{96.3} & \best{100.0} & \second{99.5}
& \second{94.4} & \second{97.1} & \best{96.5}
& \second{90.5} \\

\method{}
& \second{1.65} & \best{2.06}
& \makecell{\best{7.42}/\\\best{0.74}} & \best{0.06}
& 90.0 & 99.1 & 96.0 & 87.2 & 89.1 & 91.0 & 77.5 \\
\bottomrule
\end{tabular*}
\end{table}

Table~\ref{tab:geneval_main} reports results on the original GenEval
benchmark. \method{} reaches $90.0\%$ overall accuracy while using only
$1.65$ image-model executions, $2.06$ MLLM calls, and $7.42/0.74$k
MLLM input/output tokens per prompt. Its original GenEval score is lower
than BoN+NVILA and RAISE, but we treat this benchmark as secondary
because it is now substantially saturated and affected by benchmark
drift. In particular, the original GenEval evaluator uses a fixed
pipeline based on Mask2Former~\citep{cheng2022mask2former} and CLIP scores~\citep{ghosh2023geneval},
which can under-recognize outputs from current high-fidelity generators.
We therefore use these results as an additional reference point rather
than the main evidence for compositional alignment, and rely on GenEval2
for the primary evaluation because it was designed to address such
drift~\citep{kamath2025geneval2}.

\subsection{Original GenEval Post-hoc Audit}
\label{appendix:false_negative}
\begin{table}[t]
\centering
\caption{Post-hoc audit of \method{} outputs marked negative by the original
GenEval evaluator.
The audit was conducted with GPT-5.5 and was not used to change the
reported GenEval scores in Table~\ref{tab:geneval_main}.
The last column is the fraction judged satisfied among audited
evaluator-negative judgments.}
\label{tab:geneval_false_negative_audit}
\small
\setlength{\tabcolsep}{8pt}
\begin{tabular*}{0.78\linewidth}{@{\extracolsep{\fill}}lrrrr@{}}
\toprule
Audit subset & \shortstack{Judged\\satisfied} & \shortstack{Judged\\unsatisfied} & Total & Satisfied (\%) \\
\midrule
All audited negatives & 190 & 37 & 227 & 83.7 \\
\midrule
Color attribution & 75 & 15 & 90 & 83.3 \\
Colors & 29 & 12 & 41 & 70.7 \\
Counting & 35 & 6 & 41 & 85.4 \\
Position & 32 & 4 & 36 & 88.9 \\
Two objects & 16 & 0 & 16 & 100.0 \\
Single object & 3 & 0 & 3 & 100.0 \\
\bottomrule
\end{tabular*}
\end{table}

\begin{figure}[t]
\centering
\graphicspath{{figures/assets/geneval1_fn/}{./figures/assets/geneval1_fn/}{../figures/assets/geneval1_fn/}{assets/geneval1_fn/}}
\newcommand{\genevalfnexample}[4]{%
\begin{subfigure}[t]{0.235\linewidth}
    \centering
    \includegraphics[width=\linewidth]{#1}
    \vspace{0.15em}
    \parbox[t]{\linewidth}{\scriptsize\raggedright
    \textbf{#2}\\
    \emph{Prompt:} #3\\
    \emph{Reason:} #4}
\end{subfigure}%
}
\genevalfnexample
{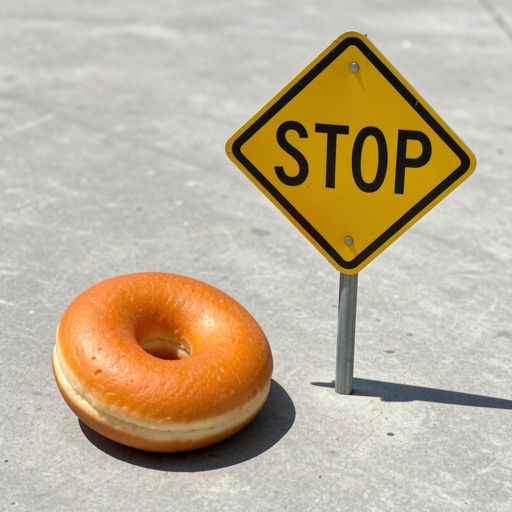}
{Color attribution}
{a photo of an orange donut and a yellow stop sign}
{Stop sign not found.}
\hfill
\genevalfnexample
{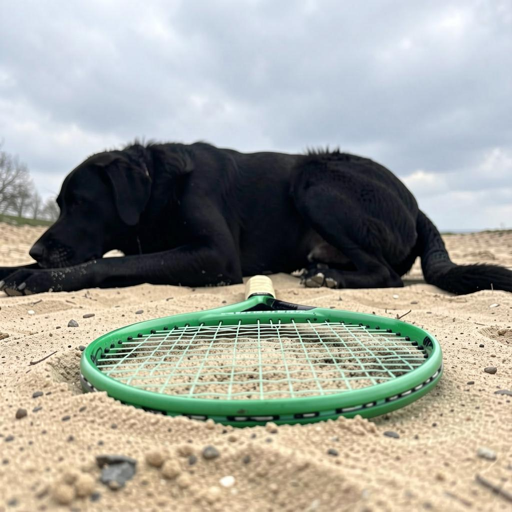}
{Color attribution}
{a photo of a green tennis racket and a black dog}
{Not green; brown racket detected.}
\hfill
\genevalfnexample
{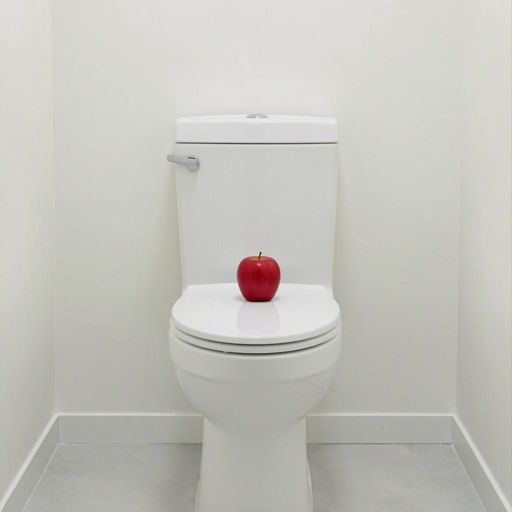}
{Color attribution}
{a photo of a white toilet and a red apple}
{Not red; pink apple detected.}
\hfill
\genevalfnexample
{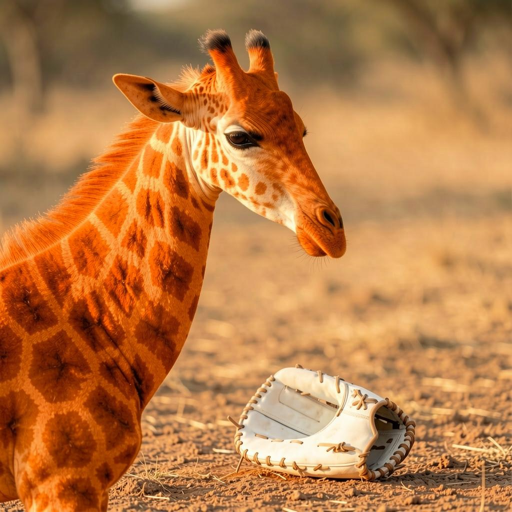}
{Color attribution}
{a photo of an orange giraffe and a white baseball glove}
{Not white; brown glove detected.}

\vspace{0.75em}

\genevalfnexample
{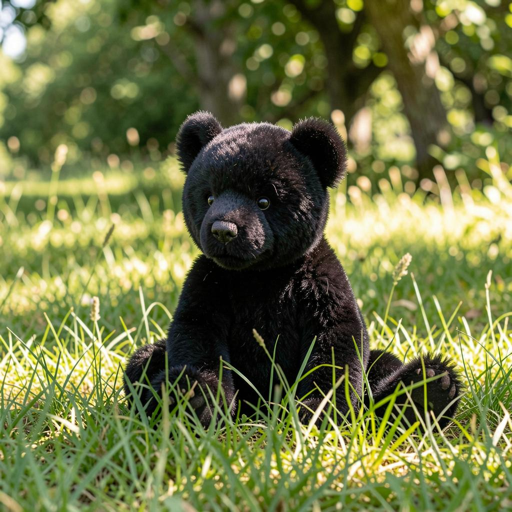}
{Colors}
{a photo of a black teddy bear}
{Teddy bear not found.}
\hfill
\genevalfnexample
{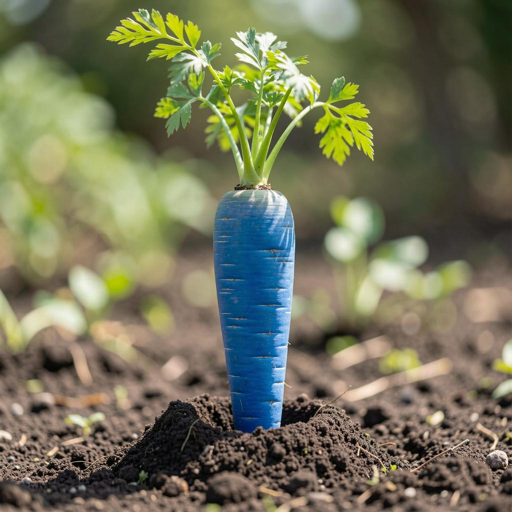}
{Colors}
{a photo of a blue carrot}
{Carrot not found.}
\hfill
\genevalfnexample
{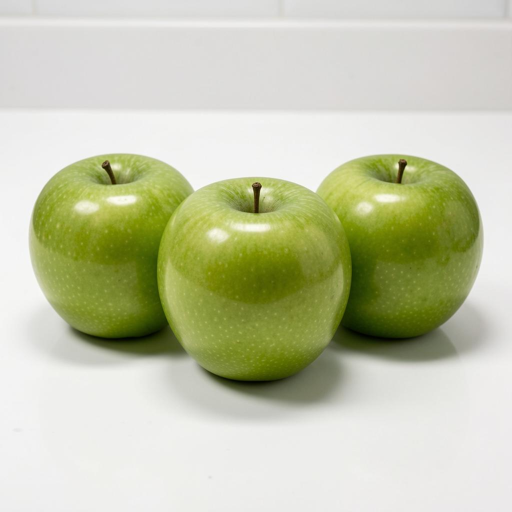}
{Counting}
{a photo of three apples}
{Only two apples counted.}
\hfill
\genevalfnexample
{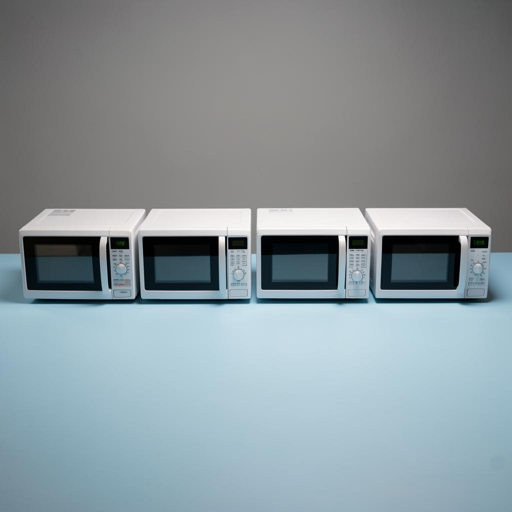}
{Counting}
{a photo of four microwaves}
{Only three microwaves counted.}

\vspace{0.75em}

\genevalfnexample
{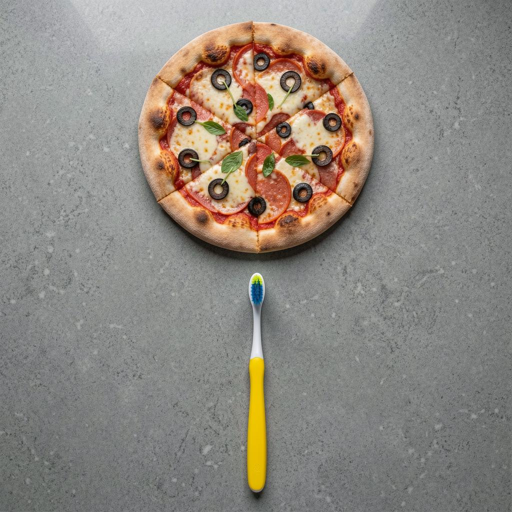}
{Position}
{a photo of a toothbrush below a pizza}
{Toothbrush not found.}
\hfill
\genevalfnexample
{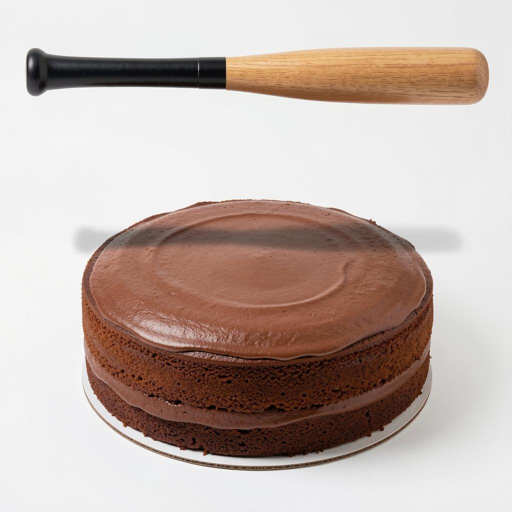}
{Position}
{a photo of a cake below a baseball bat}
{Baseball bat not found.}
\hfill
\genevalfnexample
{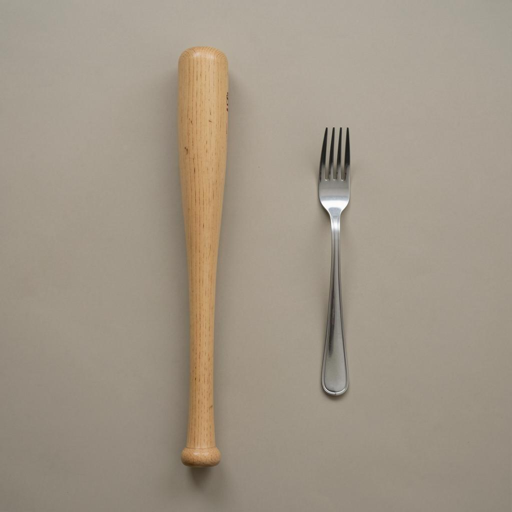}
{Two objects}
{a photo of a baseball bat and a fork}
{Baseball bat not found.}
\hfill
\genevalfnexample
{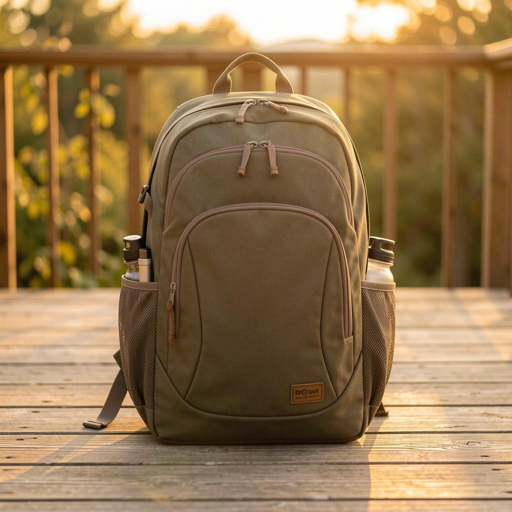}
{Single object}
{a photo of a backpack}
{Backpack not found.}
\caption{Representative cases judged satisfied by GPT-5.5
after being marked negative by the original GenEval evaluator.
Each panel shows the generated image, prompt, and original evaluator's
failure reason.}
\label{fig:geneval_false_negative_examples}
\end{figure}

Among $227$ \method{} outputs marked negative by the original
GenEval evaluator, an independent GPT-5.5 audit judges $190$ ($83.7\%$)
as satisfying the evaluated requirements and $37$ as unsatisfied
(Table~\ref{tab:geneval_false_negative_audit}).
Reported GenEval scores are unchanged.
This judged-satisfied fraction is conditional on audited evaluator-negative
judgments, not a false-negative rate over all reference-positive images.
Disagreements across all GenEval categories, including color attribution,
counting, and position (Figure~\ref{fig:geneval_false_negative_examples}),
are consistent with evaluator drift and support GenEval2 as the primary
evaluation.

\clearpage
\section{Generalization and Robustness}
\label{appendix:generalization_robustness}

This section reports natural-language evaluation, generation-seed
stability, and transfer to another generator/editor family.
Unless specified otherwise, the image backbone is FLUX.2-Klein-9B.

\subsection{Natural-language Generalization on UniGenBench++}
\label{appendix:rebuttal_generalization}

We evaluate one final image per method on all 600 English Short prompts
in UniGenBench++~\citep{wang2025unigenbench}, using the benchmark's
official Qwen3-VL-32B evaluation model.
The benchmark has 1,813 binary testpoints across 10 primary dimensions.
All methods use the same prompts and testpoints, with one generation
run and one evaluator run per method.

Table~\ref{tab:unigenbench_aggregates} reports three aggregates:
testpoint accuracy weights testpoints equally; mean per-prompt
accuracy averages within each prompt before averaging across prompts;
prompt exact pass requires every testpoint in a prompt to pass.

\begin{table}[htbp]
\centering
\caption{UniGenBench++ results (\%) over 1,813 testpoints and
600 prompts.}
\label{tab:unigenbench_aggregates}
\small
\begin{tabular}{lrrr}
\toprule
Method & Testpoint accuracy & Mean per-prompt & Prompt exact pass \\
\midrule
Direct generation & 82.13 & 81.07 & 61.17 \\
BoN+NVILA & 82.40 & 81.71 & 63.00 \\
RAISE & 85.88 & 85.15 & 69.50 \\
T2I-Copilot & \textbf{86.71} & \textbf{85.80} & \textbf{70.33} \\
\method{} & 83.89 & 83.34 & 64.50 \\
\bottomrule
\end{tabular}
\end{table}

Two-sided paired randomization tests on per-prompt accuracy use Holm
correction over all 10 method pairs. None of \method{}'s four
comparisons is significant at $0.05$: its point estimate is above
direct generation and BoN+NVILA but below the agentic baselines.

Table~\ref{tab:unigenbench_dimensions} gives the primary-dimension
breakdown. No comparison involving \method{} is significant after
Holm correction over the 100 pair-by-dimension tests, so its Text
Generation and Action leads are descriptive.

\begin{table}[htbp]
\centering
\caption{UniGenBench++ primary-dimension testpoint accuracy
(\%). $N$ is the number of testpoints; bold marks the highest estimate.}
\label{tab:unigenbench_dimensions}
\small
\setlength{\tabcolsep}{3pt}
\begin{tabular*}{\linewidth}{@{\extracolsep{\fill}}lrrrrrr@{}}
\toprule
Dimension & $N$ & Direct & BoN+NVILA & RAISE & T2I-Copilot & \method{} \\
\midrule
Style & 250 & 98.00 & \textbf{98.80} & 98.40 & \textbf{98.80} & 98.40 \\
World Knowledge & 158 & 94.30 & 92.41 & 95.57 & \textbf{96.84} & 91.14 \\
Attribute & 234 & 82.05 & 86.32 & \textbf{88.89} & 88.46 & 84.19 \\
Action & 263 & 82.13 & 82.13 & 80.61 & 82.51 & \textbf{83.27} \\
Relationship & 197 & 85.79 & 87.82 & 89.85 & \textbf{90.86} & 84.77 \\
Compound & 194 & 81.96 & 82.47 & 84.54 & \textbf{86.60} & 78.87 \\
Grammar & 187 & 75.40 & 71.66 & 83.42 & \textbf{83.96} & 81.82 \\
Logical Reasoning & 109 & 57.80 & 53.21 & 66.06 & \textbf{73.39} & 64.22 \\
Entity Layout & 134 & 88.81 & 87.31 & \textbf{94.03} & 88.06 & 88.81 \\
Text Generation & 87 & 41.38 & 47.13 & 51.72 & 52.87 & \textbf{60.92} \\
\bottomrule
\end{tabular*}
\end{table}

\subsection{Independent DrawBench Requirement Diagnostic}
\label{appendix:drawbench_requirements}

We complement the full-prompt DrawBench results in
Table~\ref{tab:drawbench_main} with an independently constructed
Soft-TIFA-style diagnostic~\citep{hu2023tifa}.
GPT-5.6 Sol decomposes the original prompts without seeing
\method{}'s visual programs. The diagnostic scores 703 image-checkable
requirements from 199/200 prompts using
Qwen3-VL-32B-Instruct~\citep{bai2025qwen3}, at
temperature $0$ with \texttt{top\_logprobs=20}. Each score is the
first-token probability of \emph{Yes} or the correct count.
We aggregate these probabilities with the same GM/AM rules as in
Sec.~\ref{sec:exp_setup}.

\begin{table}[htbp]
\centering
\caption{DrawBench requirement-level GM and AM scores (0--100),
computed from 703 independently derived requirements across 199 prompts.}
\label{tab:drawbench_independent_softtifa}
\small
\begin{tabular}{lrr}
\toprule
Method & GM & AM \\
\midrule
Direct generation & 65.32 & 81.44 \\
BoN+NVILA & 70.05 & 84.13 \\
T2I-Copilot & 73.79 & 87.23 \\
RAISE & \textbf{78.77} & \textbf{89.53} \\
\method{} & 77.62 & 88.78 \\
\bottomrule
\end{tabular}
\end{table}

\method{} improves GM over direct generation and BoN+NVILA under
two-sided paired Wilcoxon signed-rank tests with Holm correction
over four baseline comparisons ($p<0.05$); differences from the
agentic baselines are not significant. The diagnostic excludes
subjective, abstract, or holistic requirements and therefore
complements the full-prompt VQAScore and human evaluation.

\subsection{Five-seed Stability on a Fixed GenEval2 Subset}
\label{appendix:seed_stability}

We use 240 fixed GenEval2 prompts, sampled with seed 42 as 30 prompts
at each atomicity from 3 to 10. Generation seeds are
$\{0,1,2,3,4\}$, paired across methods. All methods use
FLUX.2-Klein-9B, with agentic MLLM temperature $0$, at most 32
image-model executions, and Qwen3-VL-32B-Instruct
Soft-TIFA evaluation.

\begin{table}[htbp]
\centering
\caption{Five-seed GenEval2 results on a fixed subset of 240 prompts.
Prompt-level and atom-level accuracy (\%) are reported as
mean $\pm$ sample SD across five paired generation seeds.}
\label{tab:five_seed_summary}
\small
\begin{tabular}{lrr}
\toprule
Method & Prompt acc. & Atom acc. \\
\midrule
Direct generation & $35.69\pm1.89$ & $79.53\pm0.64$ \\
T2I-Copilot & $34.41\pm0.73$ & $81.01\pm0.37$ \\
RAISE & $49.39\pm2.64$ & $85.10\pm0.79$ \\
\method{} & $\mathbf{65.88}\pm2.32$ & $\mathbf{88.85}\pm0.66$ \\
\bottomrule
\end{tabular}
\end{table}

\method{} leads every compared method in all five seeds on this
fixed subset. RAISE's scores include 24/1,200 runs (2.00\%) that
fall back to the original prompt after an LLM timeout.

\subsection{Transfer to Qwen-Image Generation and Editing}
\label{appendix:qwen_image_transfer}

We evaluate 240 GenEval2 prompts, 30 at each atomicity from 3 to 10,
using Qwen-Image-2512~\citep{qwenimage2512} as the generator and
Qwen-Image-Edit-2511~\citep{qwenimageedit2511} as the editor, from
the Qwen-Image family~\citep{qwenimage2025}. Methods share the
same subset and image backbones; \method{} uses a Qwen3-VL-32B
controller. This extends the FLUX.2-Klein-4B comparison in
Appendix~\ref{appendix:flux2_klein_4b} to another generator family.

Table~\ref{tab:qwen_image_transfer} in the main text reports the results.
These point estimates are reported without a paired significance test.

\FloatBarrier
\section{Visual Quality and Human Evaluation}
\label{appendix:quality_evaluation}

\subsection{Blinded Human Quality and Alignment Evaluation}
\label{appendix:human_evaluation}

\paragraph{Study design and sessions.}
We compare \method{} with direct FLUX.2-Klein-9B generation,
BoN+NVILA, T2I-Copilot, and RAISE on all 200 DrawBench prompts.
Each comparison pairs the final \method{} image with the final image
from one baseline for the same prompt, giving $200\times4=800$
unique image pairs. Each pair is assigned to three distinct evaluation
sessions, giving 2,400 pair evaluations and 4,800 judgments across
the two criteria.

A total of 25 volunteers participated across 40 evaluation sessions
of 60 pairs each. Each session contains
15 pairs per baseline and 60 distinct prompts. Method identities are
hidden, the images are labeled \emph{Image A} and \emph{Image B}, and
left/right placement is randomized. Participants received no compensation.

\begin{figure}[p]
\centering
\includegraphics[width=\linewidth,height=0.85\textheight,keepaspectratio]{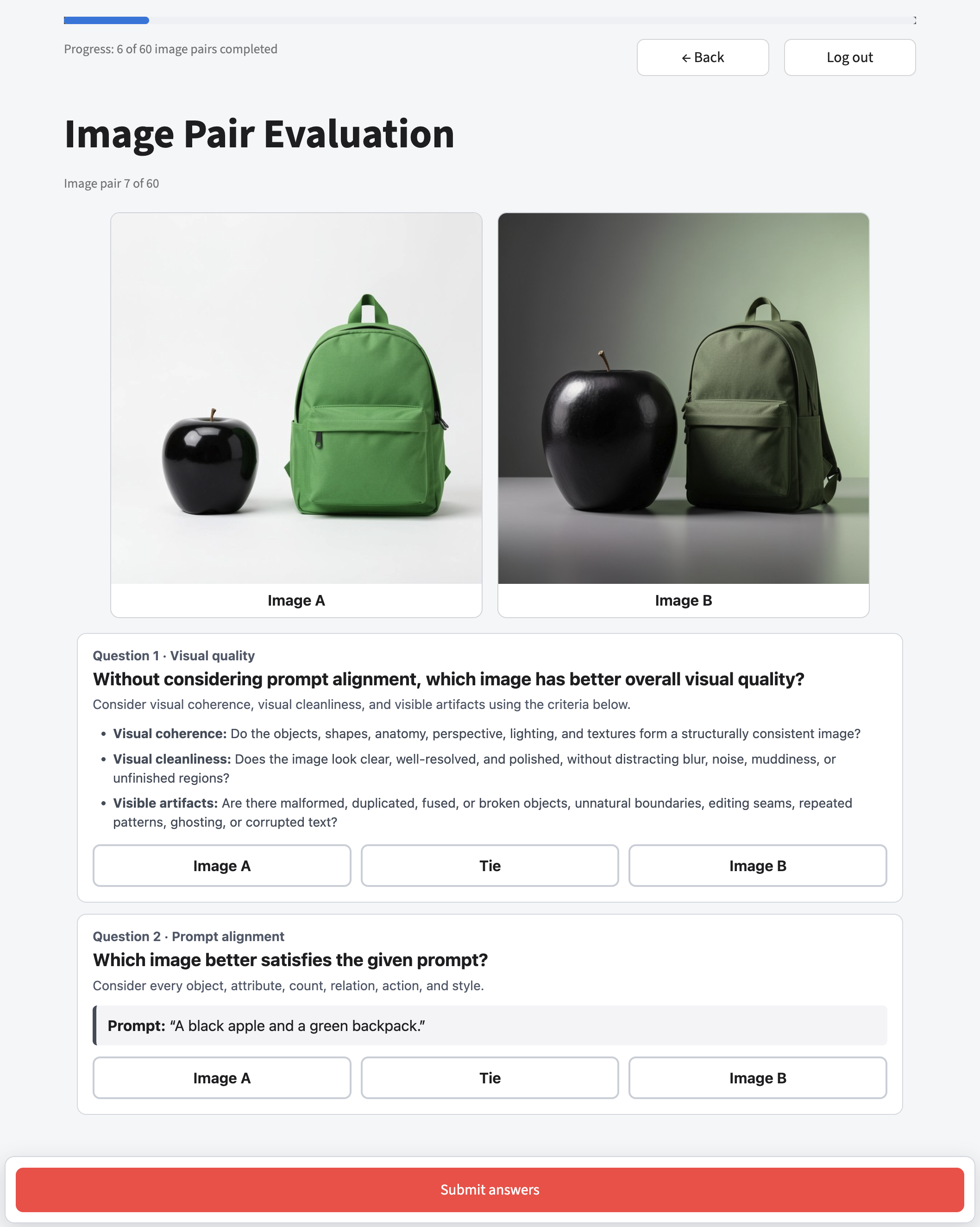}
\caption{Human evaluation interface for the blinded DrawBench
comparison. Method names are hidden behind Image A and Image B.
Participants answer separate visual-quality and prompt-alignment questions,
each with Image A, Tie, and Image B options.}
\label{fig:human_study_interface}
\end{figure}

\paragraph{Participant-facing instructions.}
Figure~\ref{fig:human_study_interface} shows the evaluation interface.
Visual quality is presented as Question 1 and prompt alignment as
Question 2, with the original prompt shown beneath the alignment question.
Each question offers \emph{Image A}, \emph{Tie}, and \emph{Image B}.
The on-screen evaluation instructions are reproduced below.

\begin{quote}
\small
\textbf{Question 1: Visual quality.}
Without considering prompt alignment, which image has better overall
visual quality? Consider visual coherence, visual cleanliness, and visible
artifacts using the criteria below.
\begin{itemize}
\setlength{\itemsep}{1pt}
\setlength{\parskip}{0pt}
\item \textbf{Visual coherence:} Do the objects, shapes, anatomy, perspective,
lighting, and textures form a structurally consistent image?
\item \textbf{Visual cleanliness:} Does the image look clear, well-resolved,
and polished, without distracting blur, noise, muddiness, or unfinished regions?
\item \textbf{Visible artifacts:} Are there malformed, duplicated, fused, or
broken objects, unnatural boundaries, editing seams, repeated patterns,
ghosting, or corrupted text?
\end{itemize}
\textbf{Question 2: Prompt alignment.}
Which image better satisfies the given prompt? Consider every object,
attribute, count, relation, action, and style.
\end{quote}

\paragraph{Preference analysis.}

Table~\ref{tab:human_drawbench_full} reports tie-adjusted preference,
$(W+0.5T)/(W+T+L)$, where $W$, $T$, and $L$ are \method{} wins,
ties, and baseline wins. Individual 95\% confidence intervals use
10,000 prompt-cluster bootstrap resamples. Significance is assessed
separately using two-sided exact sign tests on prompt-level net wins,
excluding zero net outcomes, with Holm correction over four
comparisons per criterion. Thus the three ratings for one prompt
are not treated as independent samples.

\begin{table}[htbp]
\centering
\caption{Human preference for \method{} on DrawBench (\%;
600 ratings per baseline and criterion). Ties receive half credit.
Brackets show individual 95\% prompt-cluster bootstrap intervals.}
\label{tab:human_drawbench_full}
\small
\begin{tabular}{lrr}
\toprule
Baseline & Quality [95\% CI] & Alignment [95\% CI] \\
\midrule
Direct & 51.3 [47.7, 55.0] & 54.1 [50.1, 58.0] \\
BoN+NVILA & 46.9 [43.1, 50.8] & 49.0 [45.3, 52.7] \\
T2I-Copilot & 52.0 [48.3, 55.7] & 47.7 [43.9, 51.5] \\
RAISE & 48.2 [44.8, 51.7] & 46.7 [42.9, 50.4] \\
\bottomrule
\end{tabular}
\end{table}

The bootstrap intervals describe tie-adjusted preferences, whereas
significance is assessed using Holm-corrected prompt-level sign tests.
No overall comparison is significant after Holm correction.
These results do not establish equivalence between methods.

\FloatBarrier
\subsection{No-reference Final-image Quality}
\label{appendix:quality_analysis}

We evaluate all 800 GenEval2 and 200 DrawBench final images per
method with Q-Align's \texttt{quality} task
(\texttt{q-future/one-align})~\citep{wu2023qalign}.
The evaluator receives no prompt and returns a score from 1 to 5.
Table~\ref{tab:qalign_main} reports mean scores.

Two-sided paired Wilcoxon tests use Holm correction over all
10 method pairs within each benchmark. At $p<0.05$, \method{}
scores lower than direct generation, BoN+NVILA, and T2I-Copilot
and higher than RAISE on GenEval2. On DrawBench, it scores
significantly lower than BoN+NVILA and RAISE; its differences
from direct generation and T2I-Copilot are not significant.

\subsection{Quality and Alignment Across Refinement Checkpoints}
\label{appendix:quality_trajectory}

The \emph{initial} checkpoint is the first generated candidate,
\emph{middle} is the candidate at the trajectory's middle round,
and \emph{final} is the selected output. Selection can return an
earlier candidate, so these checkpoints need not be chronological.
We recover paired checkpoints for 800 GenEval2 and 191/200
DrawBench prompts.

Table~\ref{tab:trajectory_checkpoints} reports Q-Align quality and
full-prompt VQAScore~\citep{li2024vqascore} alignment. All checkpoints
for a prompt are scored in the same VQAScore batch.

\begin{table}[htbp]
\centering
\caption{Checkpoint means and selected-final minus initial
differences. Q-Align measures quality; VQAScore measures alignment.}
\label{tab:trajectory_checkpoints}
\small
\begin{tabular}{llrrrr}
\toprule
Benchmark ($n$) & Metric & Initial & Middle & Final & $\Delta$ \\
\midrule
GenEval2 (800) & Q-Align & 4.8247 & 4.8054 & 4.7934 & $-0.0313$ \\
 & VQAScore & 0.8254 & 0.8182 & 0.8803 & +0.0549 \\
DrawBench (191) & Q-Align & 4.6921 & 4.7135 & 4.7092 & +0.0171 \\
 & VQAScore & 0.8813 & 0.8764 & 0.8964 & +0.0151 \\
\bottomrule
\end{tabular}
\end{table}

Two-sided paired Wilcoxon tests with Pratt handling of zero
differences find both GenEval2 changes significant at nominal
$p<0.01$, and neither DrawBench change significant at $0.05$.
The DrawBench final mean differs from Table~\ref{tab:qalign_main}
because only 191 trajectories are recoverable. These comparisons
describe observed trajectories and do not isolate a causal effect of
editing.

\FloatBarrier
\section{Parser and Verifier Reliability}
\label{appendix:reliability_audits}

\subsection{Source-requirement Coverage and Parser Ablation}
\label{appendix:parser_audit}

GPT-5.6 Sol compares each original prompt with the final compiled
visual program. A program is complete only if it preserves every
source requirement, allowing equivalent or compositional
representations. Missing or contradicted requirements, parser
failures, and unrelated salient foreground objects make it
incomplete. The rubric includes subjective and holistic requirements,
even when they are excluded from the image-checkable diagnostic
in Appendix~\ref{appendix:drawbench_requirements}.

\begin{table}[htbp]
\centering
\caption{Source-requirement coverage of compiled programs.
Requirement errors include omitted requirements and incorrect predicate
bindings. Explicit parser failures are reported separately.}
\label{tab:parser_source_coverage}
\small
\begin{tabular}{lrrrrr}
\toprule
Benchmark & $N$ & Complete & \makecell{Requirement\\errors} & Parser failures & Coverage \\
\midrule
GenEval2 & 800 & 781 & 19 & 0 & 97.63\% \\
DrawBench & 200 & 147 & 44 & 9 & 73.50\% \\
Pooled & 1,000 & 928 & 63 & 9 & 92.80\% \\
\bottomrule
\end{tabular}
\end{table}

GenEval2's 19 incomplete programs comprise 15 missing actions and
four incorrect relation bindings. DrawBench omissions concentrate
on pose, style, roles, and holistic composition.

\paragraph{Parser-backbone ablation.}
Replacing Qwen3-VL-32B-Instruct with GPT-5.6 Sol on the same
200 DrawBench prompts increases completeness from 73.50\% to
82.00\% and removes the nine explicit parser
failures. The parser prompt, schema, normalization, conditional-review
policy, and coverage criterion are fixed. The difference is significant
under a two-sided exact McNemar test ($p<0.01$); conditional review
adds no further coverage gain.
This is a single-run comparison, and the replacement parser and audit
reference use the same model, so shared-model bias remains possible.

\subsection{Stratified Predicate-verifier Reliability Audit}
\label{appendix:verifier_audit}

We audit a stratified sample of 1,000 candidate--predicate decisions
from 32,510 saved verifier records across GenEval2 and DrawBench.
Strata combine benchmark, detailed predicate family, and verifier state,
covering initial, edited, and resampled candidates.

GPT-5.6 Sol receives only the image and predicate, with verifier
outputs and method metadata hidden. Its judgments serve as reference
labels; the 10 uncertain judgments are excluded from precision/recall.
A positive verifier decision is
\emph{satisfied}; \emph{violated} and \emph{uncertain} are non-positive.

\begin{table}[htbp]
\centering
\caption{Population-weighted predicate-verifier audit (\%).
Audit $n$ includes reference-uncertain records; precision and recall
exclude them and use inverse-stratum weights.}
\label{tab:verifier_weighted_full}
\small
\begin{tabular}{lrrr}
\toprule
Predicate family & Audit $n$ & Precision & Recall \\
\midrule
Object existence & 30 & 100.0 & 55.4 \\
Exact count & 60 & 87.5 & 76.1 \\
Attribute & 324 & 98.5 & 51.0 \\
Action / semantic role & 60 & 80.9 & 36.7 \\
Spatial relation & 433 & 94.3 & 57.9 \\
Global scene / style & 61 & 100.0 & 61.2 \\
Visible text & 30 & 100.0 & 85.1 \\
Exclusion & 2 & 100.0 & 100.0 \\
\midrule
Pooled & 1,000 & 91.7 & 61.8 \\
\bottomrule
\end{tabular}
\end{table}

Weights are $N_h/n_h$ for population and sample sizes in stratum $h$.
For the pooled estimates, 10,000 stratified bootstrap resamples give
95\% intervals of [88.6, 94.8] for precision and [58.3, 64.9] for recall.
The intervals account for record strata but do not additionally
cluster predicates from the same image.
Recall is lowest for action/role predicates. The exclusion estimate
is especially uncertain because it is based on only two records.

\paragraph{Provenance and scope.}
The saved decisions were produced with the same SigLIP2
scorer~\citep{tschannen2025siglip2} used in the main experiments
(Appendix~\ref{appendix:implementation_verifier}).
The GenEval2 records come from retained shorter trajectories;
the estimates characterize these saved decisions.

\FloatBarrier

\clearpage
\section{Additional Limitations and Broader Impacts}
\label{appendix:additional_limitations}

\paragraph{Visual-program and verifier dependence.}
\method{} relies on a fixed visual program throughout inference. If the
parser produces a semantically wrong but well-formed program, later
refinement rounds may optimize the wrong requirements. Conditional
review is triggered by validation issues and may miss these semantic
errors. Verification can fail on small or occluded objects, unusual
viewpoints, rare attributes, or geometric relations. False negatives
can cause unnecessary refinement or budget exhaustion; false positives
can cause premature acceptance.

\paragraph{Override, scope, and evaluation.}
The limited auditor can resolve some persistent verification
failures, but its errors can also cause premature acceptance. Its
eligibility rules are specified in
Appendix~\ref{appendix:method_uncertainty_fallback}.
\method{} is limited by its predicate vocabulary, rule-based policy,
and generator/editor capabilities, and requires additional compute
over single-pass generation. Its gains are less consistent on
natural-language prompts than on GenEval2. Additional languages and
interactive settings remain to be evaluated.

\paragraph{Quality and controller scope.}
The alignment gains coexist with Q-Align deficits against some baselines
and a small measured GenEval2 trajectory trade-off
(Appendices~\ref{appendix:quality_analysis} and~\ref{appendix:quality_trajectory}).
Training-free here means that inference uses pretrained components
without training the controller, generator, or editor for this task.

\paragraph{Broader impacts.}
As discussed in Sec.~\ref{sec:limitations}, improved controllability
can support creative, educational, and accessibility applications,
but also make misleading or harmful content easier to tailor.
Inherited model biases and misuse risks motivate safety filters,
content provenance, and deployment policies.


\end{document}